\pdfoutput=1

\documentclass[11pt]{article}

\usepackage[final]{acl}

\usepackage{times}
\usepackage{latexsym}
\usepackage{booktabs}
\usepackage{hyperref} 
\usepackage{tcolorbox}
\usepackage{subcaption}
\usepackage{array}
\usepackage{xcolor}
\usepackage{soul}
\usepackage{multicol,multirow}
\usepackage{todonotes}
\usepackage{xspace}
\definecolor{mygold}{HTML}{eeba0a}
\definecolor{mygrey}{HTML}{bac8ca}
\definecolor{mylightgreen}{HTML}{D1F694}
\definecolor{myblue}{HTML}{30C0F0}
\definecolor{myorange}{HTML}{F5AF22}
\definecolor{myred}{HTML}{FF576A}
\definecolor{darkred}{HTML}{91270F}
\definecolor{darkgreen}{HTML}{5E893E}
\definecolor{lightblue}{HTML}{b6ecfc}
\sethlcolor{mylightgreen}
\usepackage{makecell}

\usepackage{subcaption}
\usepackage{cleveref}

\usepackage[table]{xcolor}
\usepackage{longtable}
\usepackage{amssymb}
\usepackage{pifont}
\newcommand{\cmark}{\textcolor{mygold}{\ding{51}}}
\newcommand{\cmarksilver}{\textcolor{mygrey}{\ding{51}}}
\definecolor{lightpink}{rgb}{1.0, 0.8, 0.9}

\newcommand{\digitbf}[1]{%
  \colorbox{lightblue}{\textbf{#1}}%
}

\newcommand{\claim}[1]{claim normalization\xspace}

\usepackage[T1]{fontenc}

\usepackage[utf8]{inputenc}

\usepackage{microtype}

\usepackage{inconsolata}

\usepackage{graphicx}

\title{M4FC: a Multimodal, Multilingual, Multicultural,
Multitask \\ Real-World Fact-Checking Dataset}

\author{
Jiahui Geng\thanks{These authors contributed equally to this work.}$^{1,2}$, Jonathan Tonglet*$^{3,4,5}$, Iryna Gurevych$^{1,3}$
\\ 
\textsuperscript{1}Mohamed bin Zayed University of Artificial Intelligence (MBZUAI), UAE\\
\textsuperscript{2} Linköping University, Sweden\\
        \textsuperscript{3}Ubiquitous Knowledge Processing Lab (UKP Lab), Department of Computer Science, \\ TU Darmstadt and National Research Center for Applied Cybersecurity ATHENE, Germany\\ 
\textsuperscript{4}Department of Electrical Engineering, KU Leuven, Belgium\\
\textsuperscript{5}Department of Computer Science, KU Leuven, Belgium\\
\href{https://www.ukp.tu-darmstadt.de}{www.ukp.tu-darmstadt.de}
}

\begin{document}
\maketitle
\begin{abstract}
Existing real-world datasets for multimodal fact-checking have multiple limitations: they contain few instances, cover only one or two languages, focus on a single task, or rely on external news article sets to source true claims.
To address these shortcomings, we introduce M4FC, a new real-world dataset comprising 4,982 images paired with 6,980 claims. The images, verified by professional fact-checkers from 22 organizations, represent a diverse range of cultural and geographic contexts. Each claim is available in one or two out of ten languages. M4FC spans six multimodal fact-checking tasks: visual claim extraction, claimant intent prediction, fake image detection, image contextualization, location verification, and verdict prediction.  We provide baseline results for all tasks and analyze how combining intermediate tasks affects verdict prediction performance. We make our dataset and code publicly available.\footnote{\href{https://github.com/UKPLab/M4FC}{github.com/UKPLab/M4FC}}
\end{abstract}

\section{Introduction}

Since 2020, approximately 80\% of online misinformation verified by fact-checkers has been multimodal, combining text with images, videos, or audio  \citep{dufour2024ammeba}. Debunking multimodal misinformation is time-consuming for human fact-checkers \citep{silverman2013verification,khan2024debunking}.  It typically involves performing several subtasks and using specialized tools, such as satellite and street-view imagery \citep{bellingcat2021,10017287,khan2024debunking}. Additionally, it requires expert knowledge of specific languages and cultures \citep{silverman2013verification}. This work targets the two most common categories of image-based misinformation \citep{dufour2024ammeba}: (1) out-of-context misinformation, where false captions misrepresent authentic images; and (2) misinformation using manipulated or fake images.

\begin{figure*}
    \centering
    \includegraphics[width=\linewidth]{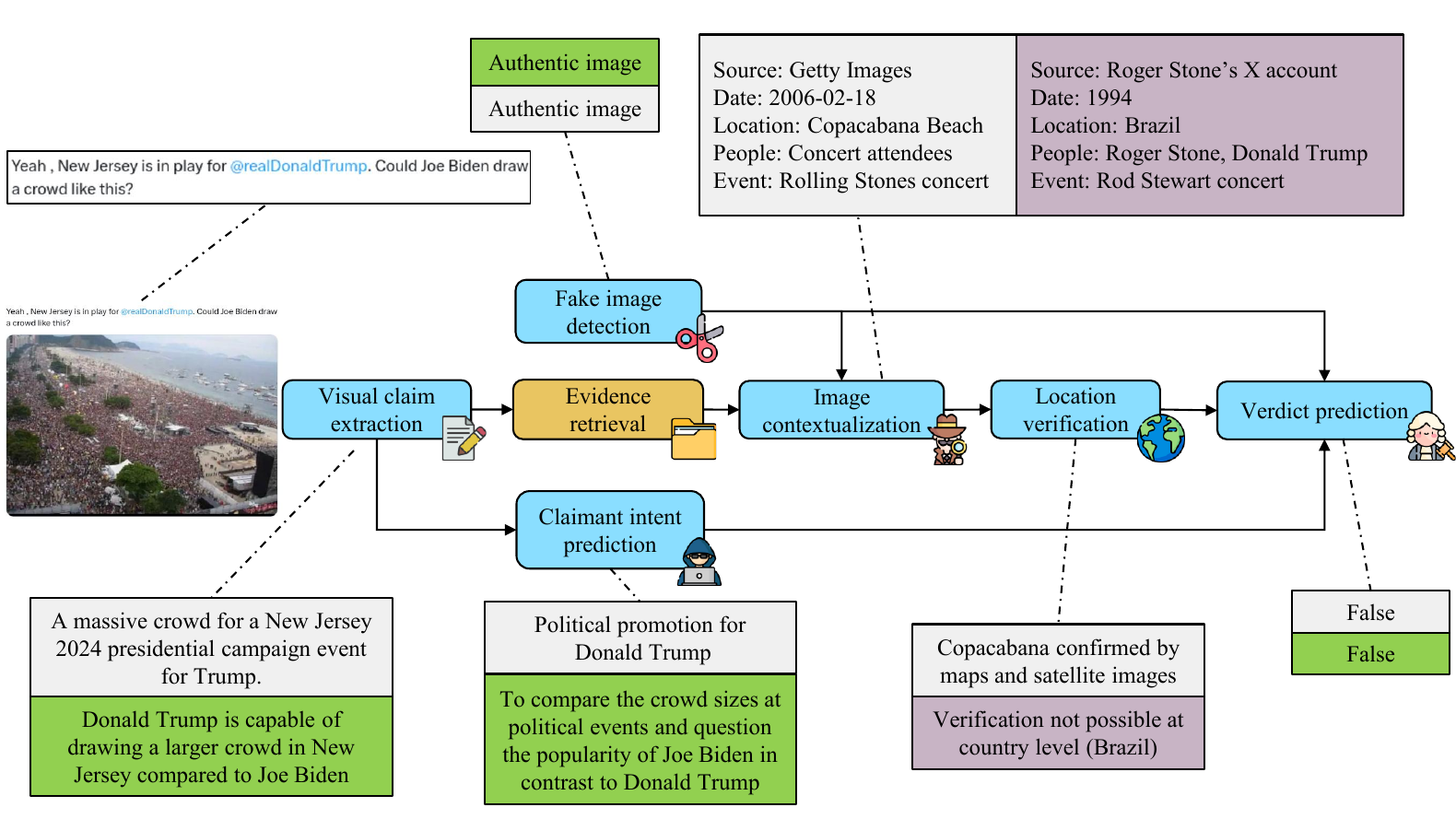}
    \caption{A false multimodal claim from M4FC. The six AFC tasks are shown in blue and their ground truth outputs in grey. Purple boxes indicate (partially) incorrect LLM answers, while green boxes indicate correct ones.
}
    \label{fig:pipeline}
\end{figure*}

Multimodal automated fact-checking (AFC) assists human fact-checkers in verifying multimodal misinformation \citep{akhtar-etal-2023-multimodal}.   Several datasets have been proposed. Synthetic datasets, on the one hand, are large and scalable \citep{luo-etal-2021-newsclippings,10440475} but cannot fully mimic real-world misinformation. On the other hand,  datasets consisting of real-world multimodal claims tend to be much smaller \citep{cheema-etal-2022-mm,10.1145/3539618.3591896,papadopoulos2024verite}. While multimodal misinformation is a global problem, these datasets cover only one or two languages and lack geographic diversity. Furthermore, most datasets cover only the verdict prediction task, whereas the human fact-checking workflow involves several intermediate tasks.

In this work, we introduce \textbf{M4FC}, a \textbf{m}ultimodal, \textbf{m}ultilingual, \textbf{m}ulticultural, and \textbf{m}ultitask real-world  \textbf{f}act-\textbf{c}hecking dataset, with 4,982 images and 6,980 claims. M4FC addresses several limitations of existing resources, as shown in Table \ref{tab:literature-review}. It contains labels for six multimodal AFC tasks organized into a pipeline, as illustrated in Figure \ref{fig:pipeline}. This pipeline automates a larger part of the fact-checking workflow than prior datasets. Images, claims, task labels, and metadata are drawn from articles produced by 22 fact-checking organizations across 17 countries. As a result, M4FC offers broad cultural and geographic coverage and spans 10 claim languages, substantially more than prior  real-world datasets: Arabic, Dutch, English, French, German, Portuguese, Spanish, Tamil, Telugu, and Turkish.
We introduce two new multimodal AFC  tasks. \textit{Visual claim extraction} formulates verifiable claims based on a screenshot of a social media post. \textit{Location verification} validates a candidate location for an image using two types of evidence previously overlooked in multimodal AFC research: maps and satellite images. We provide baseline results for all tasks. Furthermore, we evaluate how different combinations of intermediate tasks affect verdict prediction performance. 

In summary, our contributions are as follows: (1) we introduce M4FC, a large-scale real-world multimodal, multilingual, multicultural, and multitask fact-checking dataset; (2) we introduce two new challenging multimodal AFC tasks; and (3) we conduct comprehensive experiments, including analyses of how combining intermediate tasks impacts verdict prediction performance.

\section{Related work}

\begin{table*}
  
  \centering
  \resizebox{0.9\textwidth}{!}{ %
  \begin{tabular}{lllllllllll}
    \toprule
    &  & & & \multicolumn{4}{c}{Claims} &  & \multicolumn{2}{c}{Evidence}        
    \\
    \cline{5-8}\cline{10-11}
    Dataset     & Type & \# tasks & \# images & \# claims  &     \# languages  & Data split & Source  &  & Leakage? & Type \\
    \midrule
    MEIR \citep{10.1145/3240508.3240707} & Synthetic & 1 & 57,940 & 57,940 & 1 (En) & Unknown & Flickr &   & -  &  - \\
    NewsCLIPpings \citep{luo-etal-2021-newsclippings}  & Synthetic & 1 & 85,360 & 42,680  & 1 (En) & Claim & News  &  & -  & - \\    
    DGM4 \citep{10440475} & Synthetic & 1 & 200,559  & 139,560 & 1 (En) & Unknown & News &   & -  & - \\
    MiRAGeNews \citep{huang-etal-2024-miragenews} & Synthetic &  1 & 12,500 &  12,500 &  1 (En) & Unknown & News &  & - &  - \\
    MMFakeBench \citep{liu2025mmfakebenchmixedsourcemultimodalmisinformation}  & Synthetic & 1 & 11,000 & 11,000  & 1 (En) & Unknown & Diverse  &  & -  & - \\   
    Weibo \citep{10.1145/3123266.3123454} &  Real-world & 1 & 9,528 & 9,528  & 1 (Zh) & Event & FC \& News  &  & -  & - \\
    Fauxtography \citep{zlatkova-etal-2019-fact} & Real-world & 1 & 1,233 & 1,233 & 1 (En) & Random & FC \& News &  & No &    Text  \\
    MM-Claims \citep{cheema-etal-2022-mm} &Real-world & 1  & 3,400 & 3,400 & 1 (En) &  Unknown & Twitter (now X) &  & -  & - \\
    COSMOS (test) \citep{aneja2023cosmos}  & Real-world & 1 & 1,700 & 3,400   & 1 (En) & No split  & FC \& News &  & - & -    \\
    MR2 \citep{10.1145/3539618.3591896} & Real-world & 1  & 14,700 & 14,700 & 2 (En, Zh)  & Event & FC \& News &  & Yes &   Text, Image   \\
    Ookpik \citep{pham2024ookpik}  & Real-world & 1  & 545 & 1,090   & 1 (En)  & No split & FC \& News  & & - &  -   \\
    Post-4V \citep{geng2024multimodallargelanguagemodels}  & Real-world & 1  & 186 & 186 & 1 (En)  & No split & FC & & - &  -     \\
    VERITE \citep{papadopoulos2024verite} & Real-world & 1 & 662 & 1,000  & 1 (En) & No split & FC & & Yes &  Text, Image   \\
    5Pils \citep{tonglet-etal-2024-image}   & Real-world & 1 & 1,676 & 1,676 & 1 (En) & Temporal & FC  &   & No &  Text  \\
    5Pils-OOC \citep{tonglet-etal-2025-cove}  & Real-world & 2 &  624 & 1,248  & 1 (En) & No split & FC  & &   No &  Text, Image    \\
    AVerImaTeC \citep{cao2025averimatec}  & Real-world  & 1  &  1,297  & 1,297  & 1 (En) & Temporal  & FC & &  No &  Text, Image \\
    XFacta \citep{xiao2025xfactacontemporaryrealworlddataset} & Real-world &  1 & 2,400  & 2,400  & 1 (En) & Random & CN \& News &  &  Unknown  &  Text, Image \\
    \midrule
    \textbf{M4FC (ours)} & \textbf{Real-world} & \textbf{6} & \textbf{4,982} & \textbf{6,980}  & \textbf{10} & \textbf{Temporal} & \textbf{FC} & & \textbf{No} & \textbf{Text, Map, Satellite}    \\
    \bottomrule
  \end{tabular}}
  \caption{Comparison with relevant multimodal AFC datasets. En and Zh denote claims in English and Chinese, while FC, News, and CN refer to fact-checking articles, news articles, and X Community Notes, respectively.}
  \label{tab:literature-review}
\end{table*}

\subsection{Synthetic multimodal AFC datasets}

Several synthetic datasets have been proposed for multimodal AFC \citep{10.1145/3240508.3240707,10.1145/3372278.3390670,luo-etal-2021-newsclippings,10440475,liu2025mmfakebenchmixedsourcemultimodalmisinformation,he2025explainablebilingualmultimodalmisinformation}.
 While large and scalable, they often suffer from biases. \citet{10944123} demonstrated that simple models can exploit these biases to achieve SOTA performance, raising concerns about the reliability of synthetic data for multimodal AFC.

\subsection{Real-world multimodal AFC datasets}
Real-world datasets contain real claims published on the web and fact-checked by humans. They are summarized in Table \ref{tab:literature-review}. Because human fact-checking is time-intensive, real-world datasets are smaller than synthetic ones. They also often lack true claims \citep{10.1145/3477495.3531744}, as fact-checkers prioritize writing articles about false claims. To compensate, many datasets sample image-caption pairs from reputable news outlets and treat them as true claims \citep{10.1145/3123266.3123454, zlatkova-etal-2019-fact,aneja2023cosmos,10.1145/3539618.3591896}. However, this introduces a distribution mismatch \citep{10.1145/3477495.3531744}, as the content and resolution of news images, and the writing style of their captions, may differ too much from real-world misinformation. Most existing datasets cover one or two claim languages and provide labels for only one task: verdict prediction. Many datasets gather evidence via reverse image search (RIS) \citep{zlatkova-etal-2019-fact,tonglet-etal-2024-image} or by querying a search engine \citep{cao2025averimatec}. To ensure a realistic setting and prevent the ground truth from leaking in the evidence set, i.e., evidence leakage, several datasets exclude evidence written by fact-checking organizations or published after the claim date \citep{zlatkova-etal-2019-fact,tonglet-etal-2024-image,cao2025averimatec}.

M4FC addresses several limitations of existing datasets. By gathering data from 22 fact-checking organizations from around the world, it covers 10 languages and is geographically diverse. Furthermore, M4FC includes six tasks, aligning multimodal AFC more closely with human fact-checking practices.

\section{M4FC Dataset}

\subsection{Six multimodal AFC tasks}
\label{sec:tasks}

M4FC includes six tasks, two of which are new: visual claim extraction and location verification.

\begin{figure}[!t]
    \centering
    \includegraphics[width=\columnwidth]{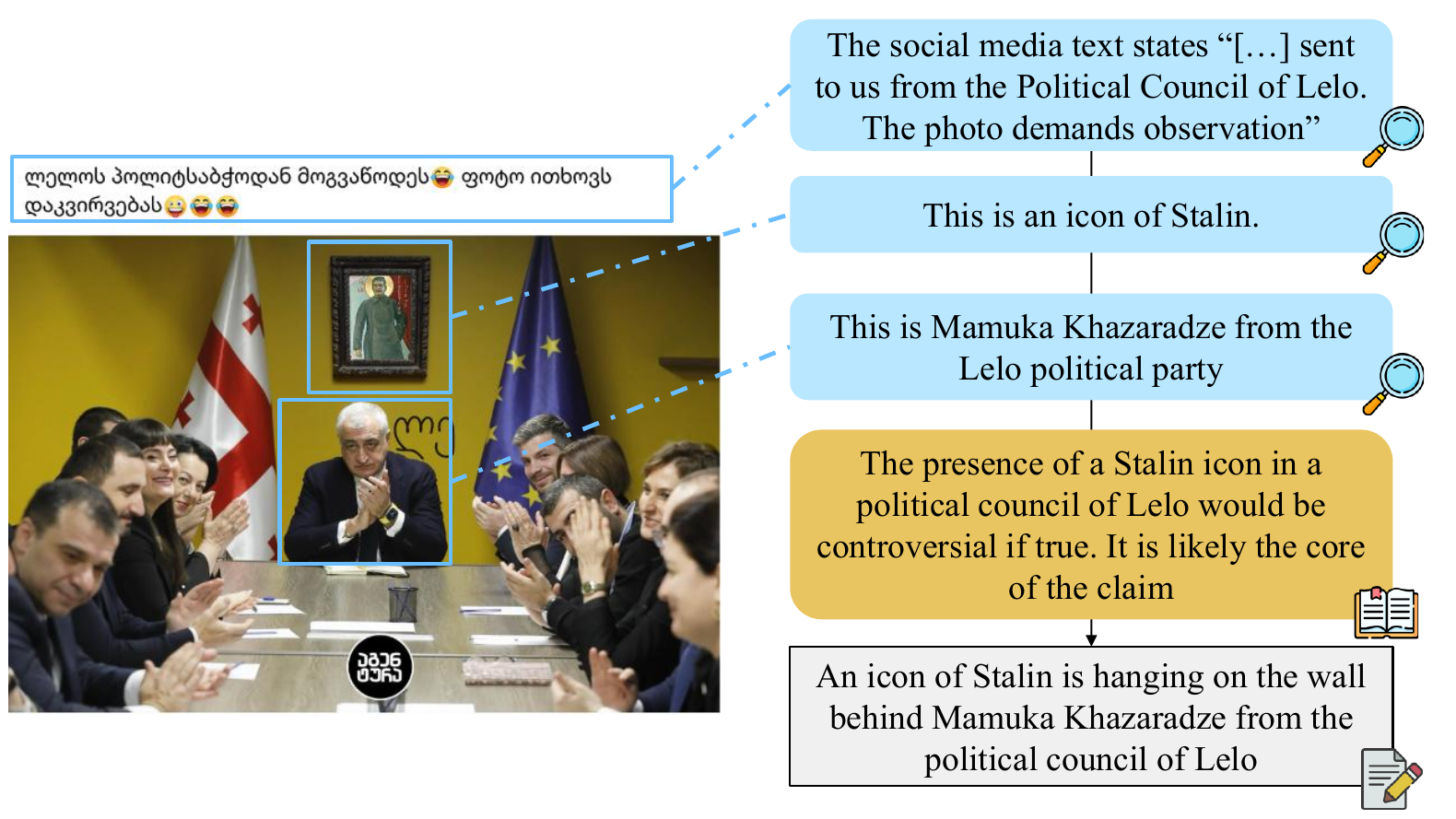}
    \caption{Illustration of a human reasoning for visual claim extraction. Blue, yellow, and gray boxes indicate visual cues, external knowledge, and the extracted claim, respectively.}
    \label{fig:claim-extraction}
\end{figure}

Claims are not always expressed online in an explicit and verifiable format. 
\textbf{Visual claim extraction (VCE)} generates a verifiable English-language claim from the screenshot of a social media post.  This task involves two challenges: (1) extracting visual cues and text from a wide range of languages, and (2) leveraging external knowledge. As shown in Figure \ref{fig:claim-extraction}, parts of the claim appear in the text, others from the wall, and further clarification requires external knowledge.

\begin{figure}
    \centering
    \includegraphics[width=\columnwidth]{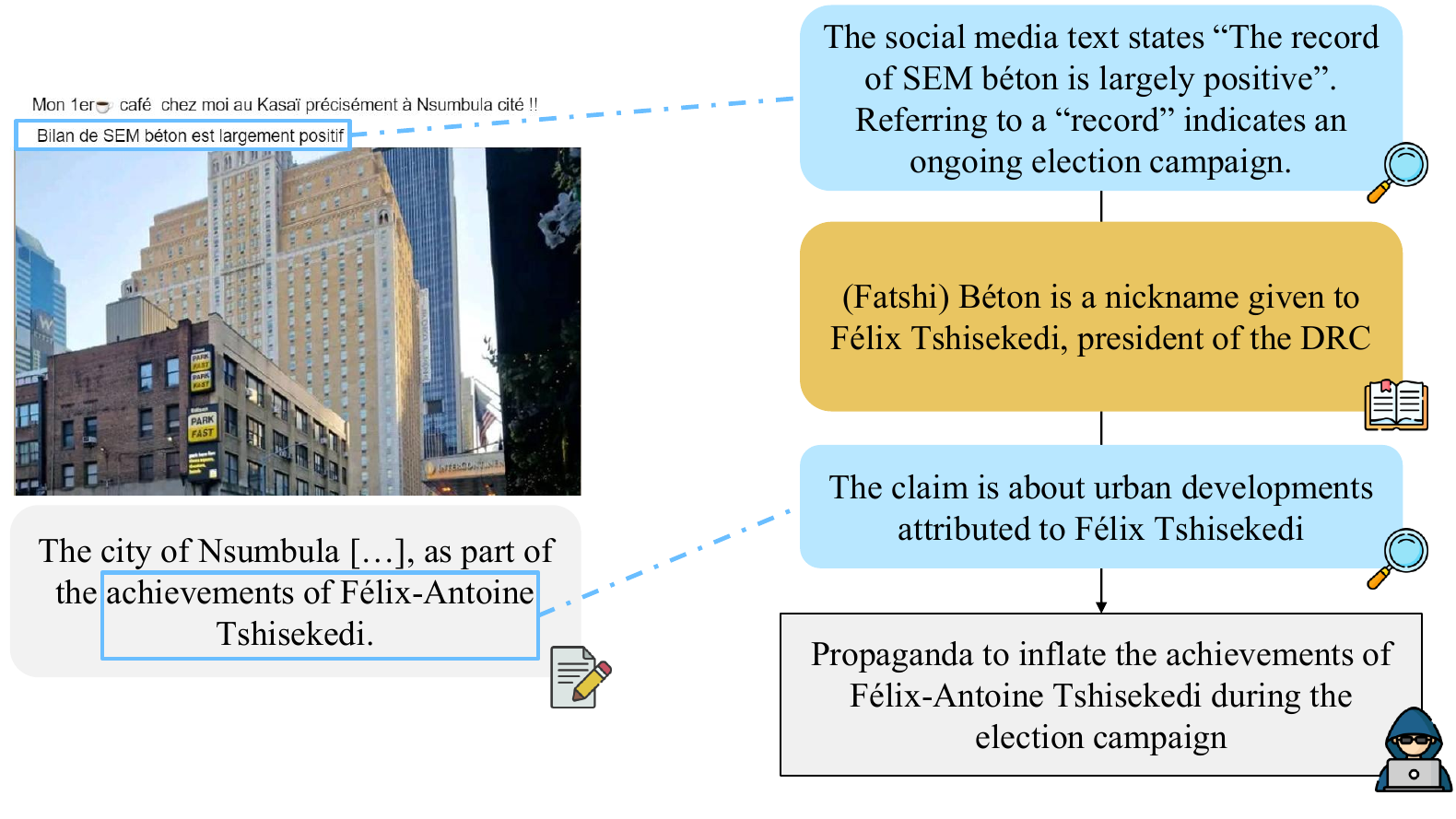}
    \caption{Illustration of a human reasoning for claimant intent prediction.  Blue, yellow, and gray boxes indicate visual cues, external knowledge, and the claimant's intent, respectively.}
    \label{fig:claim-intent}
\end{figure}

\textbf{Claimant intent prediction (CIP)}  seeks to identify the underlying purpose of the claimant, such as expressing sarcasm or expressing support for a political figure. We formulate it as a free-text generation task.  While \citet{da-etal-2021-edited} and \citet{wu2025seeingdeceptionuncoveringmisleading} predicted claimant intents using synthetic data, we extend their work to real-world claims. Figure \ref{fig:claim-intent} shows an example where the claimant expresses support for a political figure. Like VCE, this task involves extracting visual cues and querying external knowledge.

\textbf{Fake image detection (FD)} identifies images that are manipulated or fake. Manipulations include face swaps or background changes, while fake images encompass AI-generated content and human-made digital artwork. This task has two labels: \textit{authentic} and \textit{manipulated/fake}.

\textbf{Image contextualization (IC)} determines the original context of an image \citep{tonglet-etal-2024-image,tonglet2026newsrecon}. This task is framed as predicting a set of items. We cover the following context items \citep{tonglet-etal-2024-image,tonglet-etal-2025-cove}: \emph{provenance} (binary variable indicating whether the image has previously appeared on the web), \emph{source} (original author), \emph{date}, \emph{location}, \emph{motivation} (why the source made or published the image), \emph{people}, \emph{things}, and \emph{event}.

\begin{figure}[!t]
    \centering
    \includegraphics[width=0.95\columnwidth]{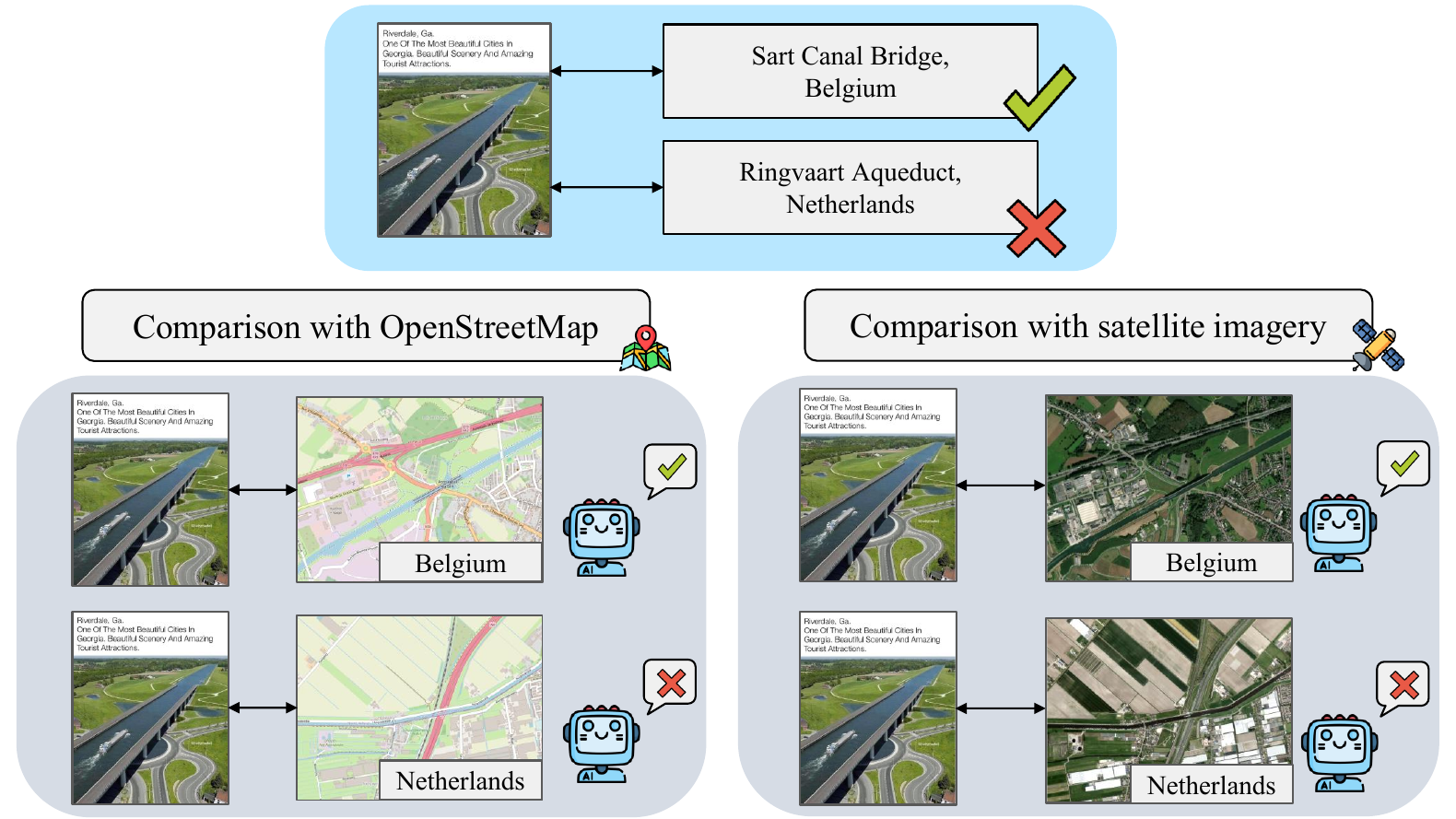}
    \caption{Illustration of the location verification task. Satellite images are sourced from ESRI World Imagery.}
    \label{fig:location-verification}
    
\end{figure}

 Once IC identifies a location context item, human fact-checkers often compare the image with aerial views of that location to validate it \citep{khan2024debunking}. This is the \textbf{location verification (LV)} task: given a map or satellite view of a candidate location, the goal is to determine whether it is consistent with the image. In the example of Figure \ref{fig:location-verification}, the first candidate location, Sart Canal Bridge, is confirmed by the matching road network below the bridge. In contrast, another candidate location, the Ringvaart aqueduct, is ruled out because it spans a four-lane highway, which does not match the image content. This task relates to cross-view geolocalization \citep{6618964}, in which the objective is to retrieve a matching satellite image from a large database given a ground-level image.

\textbf{Verdict prediction (VP)} classifies a claim as true or false. The task is defined in four settings, based on two factors: whether claims are \textit{English-only} or \textit{multilingual}, and whether the verdicts are \textit{imbalanced} or \textit{balanced}. The \textit{multilingual} setting covers 10 languages. The \textit{imbalanced} setting captures the real-world distribution of fact-checking verdicts, in which most claims are false. The \textit{balanced} setting replaces a portion of false claims with synthetic true claims based on the image's context items. The \textit{English-only} \textit{imbalanced} setting can be used for training, while other settings are only recommended for evaluation, due to the presence of linguistic biases, as discussed in Appendix \ref{sec:biases}.

\subsection{Dataset construction}
\subsubsection{Data collection}

The data in M4FC is labeled by human fact-checking experts. We collected fact-checking articles from 22 organizations based in 17 countries, all members of the International Fact-Checking Network (IFCN). Appendix \ref{sec:fc_org_dist} reports the distribution of articles per  organization.

All task labels are provided in the human-written article text. Following \citet{tonglet-etal-2024-image}, we use GPT4o \citep{openai2023gpt4} to extract task labels from the article, if available. Afterward, GPT4o generates synthetic true claims by combining the context items (date, location, people, and event) of instances labeled as false. For a random half of those instances, the original false claim is replaced by the synthetic true claim under the \textit{balanced} verdict prediction setting. 

Images suitable for location verification are identified semi-automatically. We use keywords to filter articles referencing maps, satellites, street names, town squares, or landmarks. Furthermore, the image needs a location context item that can be mapped to coordinates using Geonames. We then manually verify that the image shows an outdoor scene. We also need incorrect location candidates for each suitable image. We obtain them by prompting GPT4o to generate a list of the most likely locations for the image. The first two that can be mapped to valid coordinates are selected. As a result, each image is paired with one correct and two incorrect locations to verify.

 To mitigate knowledge leakage, we apply a temporal train/dev/test split \citep{glockner-etal-2022-missing}, ensuring that the test claims are the most recent.
 
  All prompts are shown in Appendix \ref{sec:prompts_labeling}. 
   Appendix \ref{sec:data_collection} provides more details about data collection.

\subsubsection{Data validation}

We conduct rigorous data validation in two stages.

\paragraph{Annotators validation} For each organization, we randomly sample up to 20 articles. Each article is reviewed by two Prolific annotators who evaluate the correctness of the data fields extracted by GPT4o. 79 annotators participated in the study, of whom 33 were excluded for failing more than five out of 15 attention checks. A label is considered incorrect if both annotators agree on it. The majority of labels are considered  accurate, with scores ranging from 86.2\% (for claimed date) to 99.31\% (for verdict).  We compute inter-annotator agreement (IAA) using Randolph’s $\kappa$ \citep{randolph2005free}, which ranges from 0.32 to 0.81 across all data fields. Guidelines and IAA scores are provided in Appendices \ref{sec:annotation_guidelines} and \ref{sec:iaa}.

\paragraph{Manual correction} Given the low to moderate IAA of the Prolific annotators, the authors further manually corrected the labels of the entire test set, using the corresponding fact-checking articles as reference. Only 0.7\% of the labels were found to be (partially) incorrect, validating the high-quality of the GPT4o label extraction. This also indicates that Prolific annotators were overestimating the amount of incorrect labels. \ref{sec:disagreement} provides such examples where one or two annotator mistakenly consider a GPT4o label to be incorrect.

\subsubsection{Dataset statistics}

Table \ref{tab:tasks_recap} provides the distribution of instances per task. 1,142 images are unsuitable for visual claim extraction because they do not contain any embedded text. Location verification has the fewest instances because it applies only to the images that show outdoor scenes and have a precise location context item. M4FC images span six continents and 159 countries, resulting in greater cultural diversity and reduced geographic bias compared to prior work. The top three continents represented are \emph{Asia}, \emph{Africa}, and \emph{Europe}.  Appendix \ref{sec:dataset_statistics} provides more statistics. Appendix \ref{sec:subset_annotated} discusses the subset of M4FC with labels validated by all annotators.

\subsubsection{Evidence collection}

For image contextualization and verdict prediction, we collect up to 20 web evidence using the Google RIS engine.\footnote{\href{https://cloud.google.com/vision/docs/detecting-web}{cloud.google.com/vision/docs/detecting-web}} We remove evidence published by fact-checking organizations or after the claim date.   For location verification, we use the coordinates of the candidate locations to collect maps from OpenStreetMap \citep{OpenStreetMap} and satellite images from ESRI World Imagery,\footnote{\href{https://www.arcgis.com/home/item.html?id=10df2279f9684e4a9f6a7f08febac2a9}{arcgis.com/home/item.html}}  accessed with staticmap.\footnote{\href{https://github.com/komoot/staticmap}{github.com/komoot/staticmap}}

\section{Experiments} 

We describe below our experiments on the six tasks of M4FC. The tasks, and their inputs and outputs, are defined in Section \ref{sec:tasks} and in Table \ref{tab:tasks_recap}.

\subsection{Evaluation metrics}

For visual claim extraction and claimant intent prediction, we use Rouge-L \citep{lin-2004-rouge}, METEOR \citep{banerjee-lavie-2005-meteor}, BERTScore \citep{bert-score}, and G-Eval~\cite{liu-etal-2023-g} using GPT4o as the judge. Appendix \ref{sec:g-eval} provides the G-Eval prompts. For fake image detection, location verification, and verdict prediction, we report the macro F1-score (F1). For image contextualization, following \citet{tonglet-etal-2025-cove}, we adopt METEOR for source, motivation, things, and event, F1 for people, $\Delta$ for date, and $CO\Delta$ for location. $\Delta$ and $CO\Delta$ report the inverse proportional distance from the ground-truth in years and in thousand kilometers, respectively. For verdict prediction with intermediate tasks, we report the accuracy (Acc), precision (Pre), recall (Rec), and F1.

\subsection{Baselines}

\begin{table}
  \small
  \centering
  \resizebox{0.7\columnwidth}{!}{ %
  \begin{tabular}{llllll}
    \toprule
    Task   & \# instances & \# train   & \# dev   & \# test  &    Input    \\
    \midrule
     VCE & 3,840 & 2,284 & 414 & 1,142 & I  \\
     CIP & 1,754 & 1,104 & 161 & 489 & I, C  \\
     FD & 4,954 & 2,978 & 498 & 1,478 & I \\
     IC & 4,720 & 2,887 & 480 & 1,353 & I, C, E   \\
     LV & 195 & 96 & 27 & 72  & I, E   \\
     VP & 4,955 & 2,979 & 498 & 1,478 & I, C, E  \\

    \bottomrule
  \end{tabular}}
  \caption{Number of instances and input types per task. I is the image, C is the claim, E is the evidence set.}
  \label{tab:tasks_recap}
\end{table}

\begin{figure*}[htbp]
    \centering
    \begin{subfigure}[b]{0.45\textwidth}
        \centering
        \includegraphics[width=\linewidth]{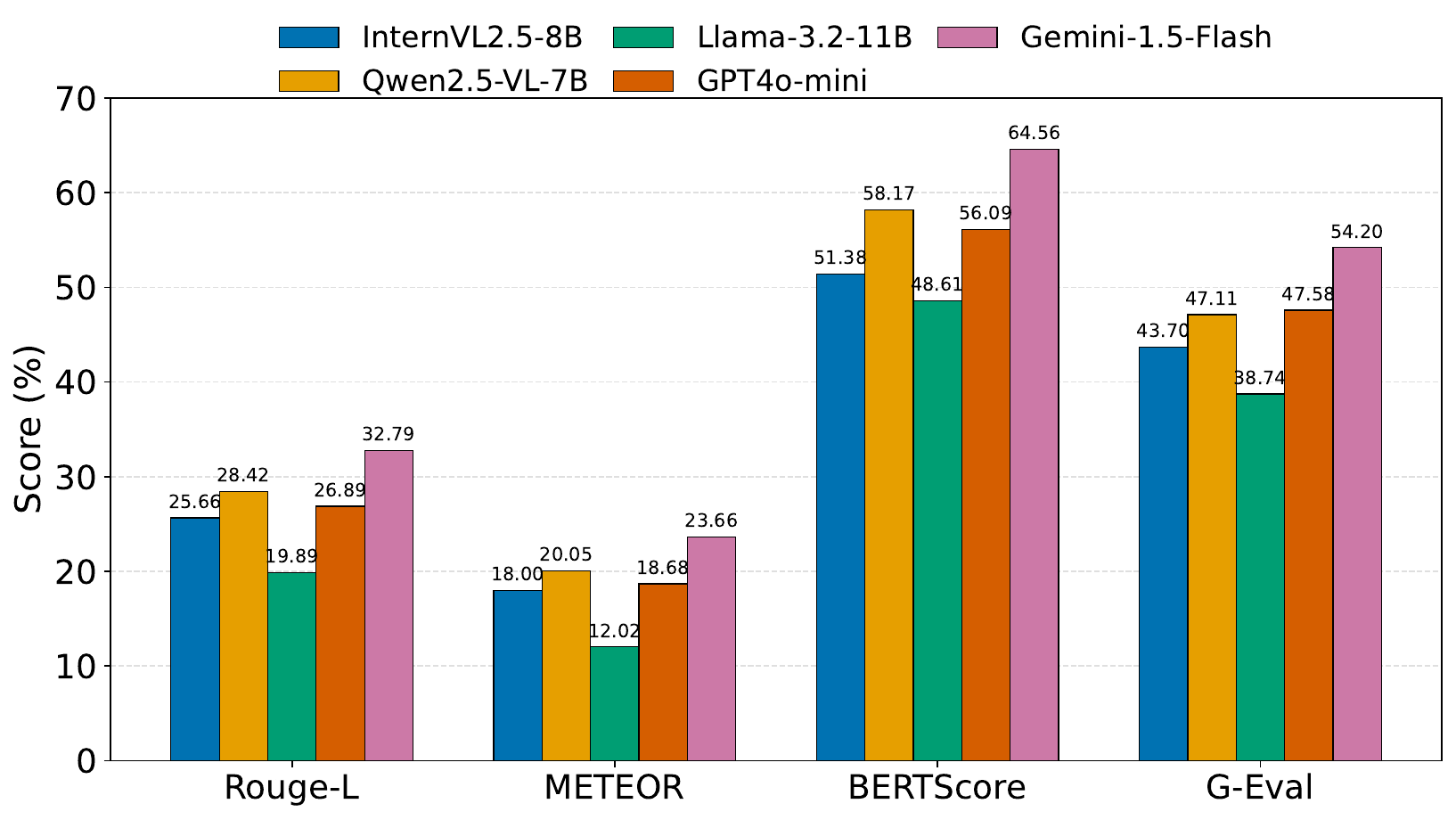}
        \caption{Visual claim extraction results, grouped by metric.}
        \label{fig:vce_vis}
    \end{subfigure}
    \hfill
    \begin{subfigure}[b]{0.45\textwidth}
        \centering
        \includegraphics[width=\linewidth]{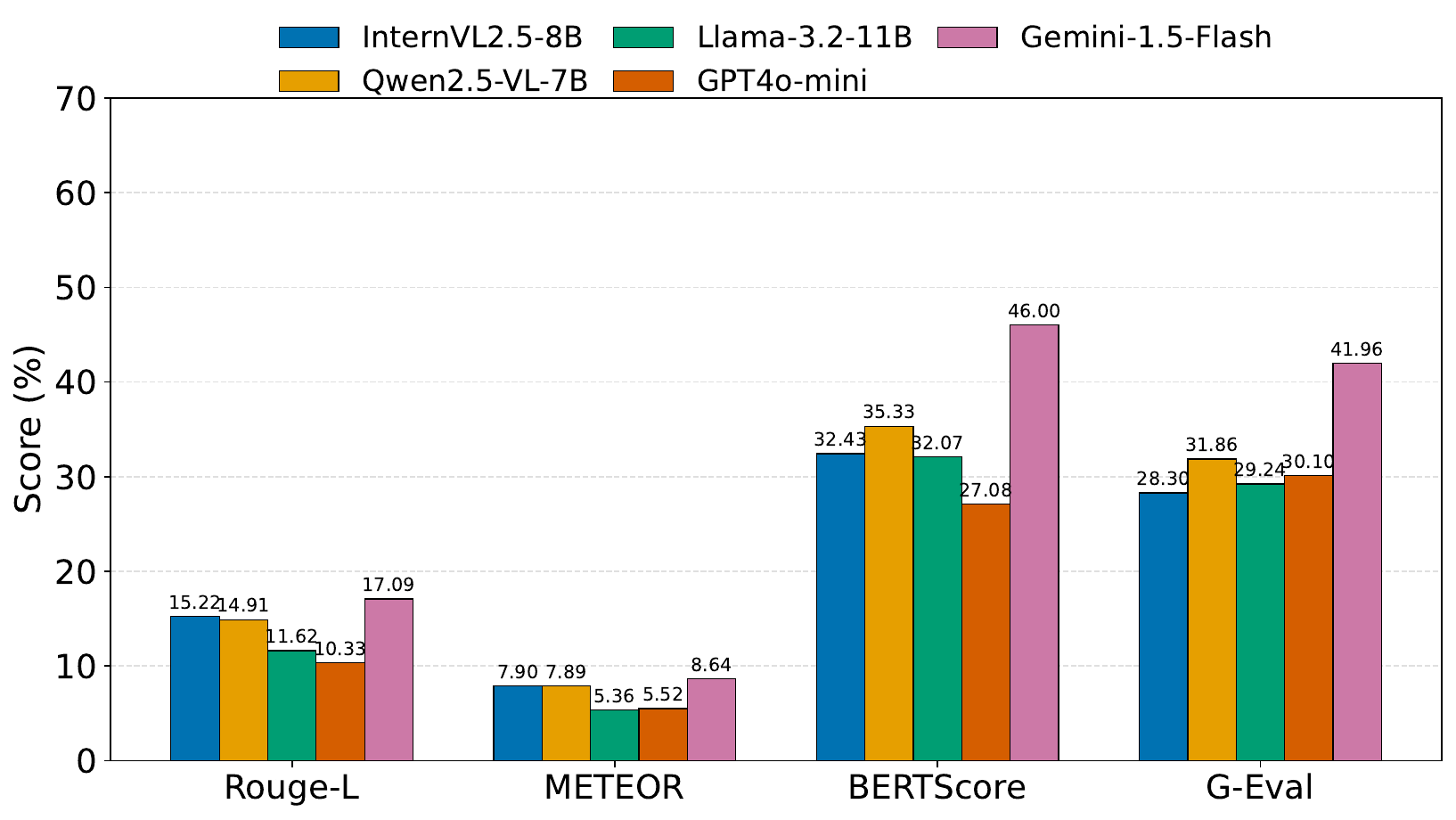}
        \caption{Claimant intent prediction results, grouped by metric.}
        \label{fig:cip_vis}
    \end{subfigure}
    
    \vskip\baselineskip
    \begin{subfigure}[b]{0.45\textwidth}
        \centering
        \includegraphics[width=\linewidth]{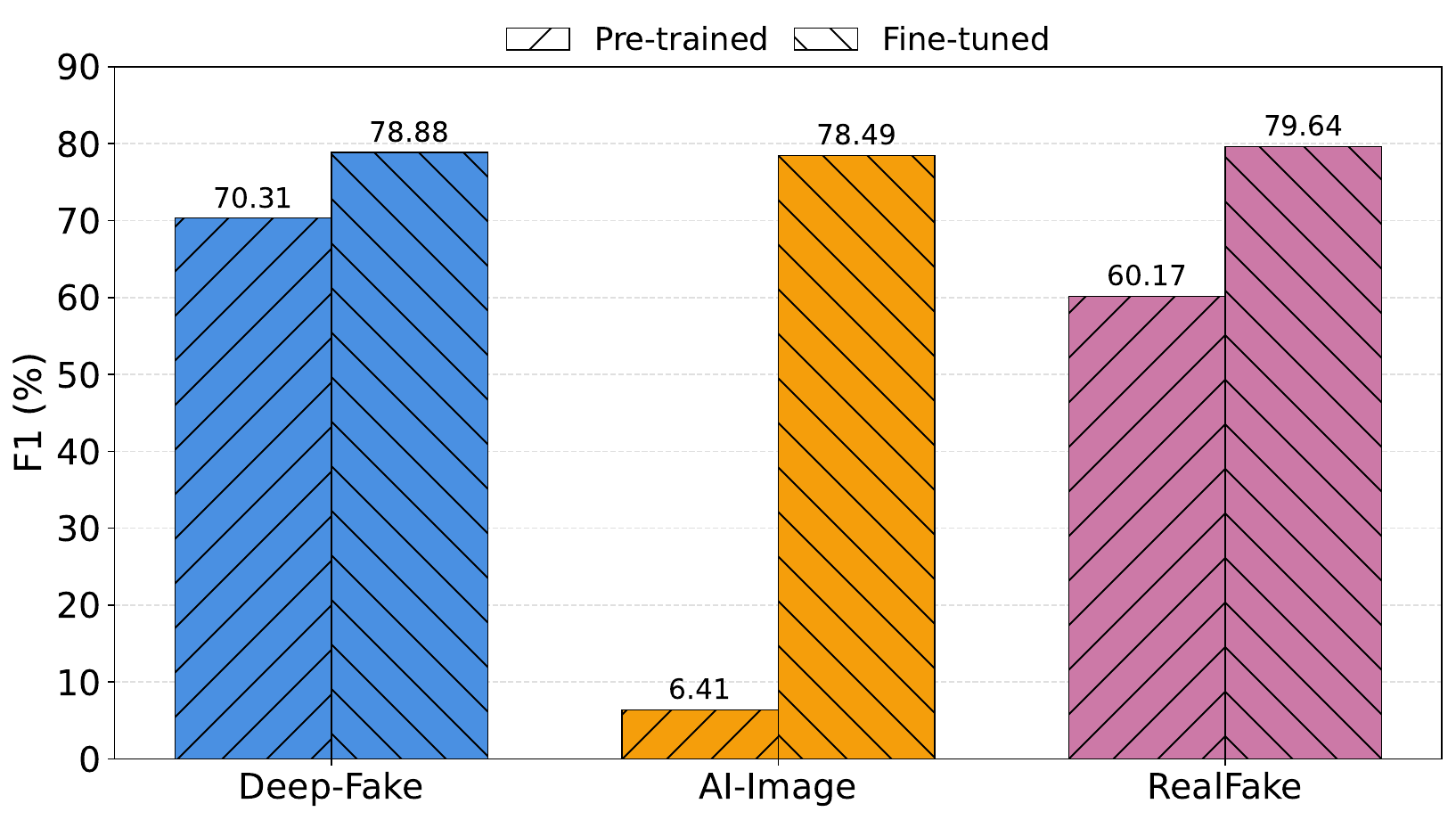}
        \caption{Fake image detection results, grouped by model.}
        \label{fig:fd_vis}
    \end{subfigure}
    \hfill
    \begin{subfigure}[b]{0.45\textwidth}
        \centering
        \includegraphics[width=\linewidth]{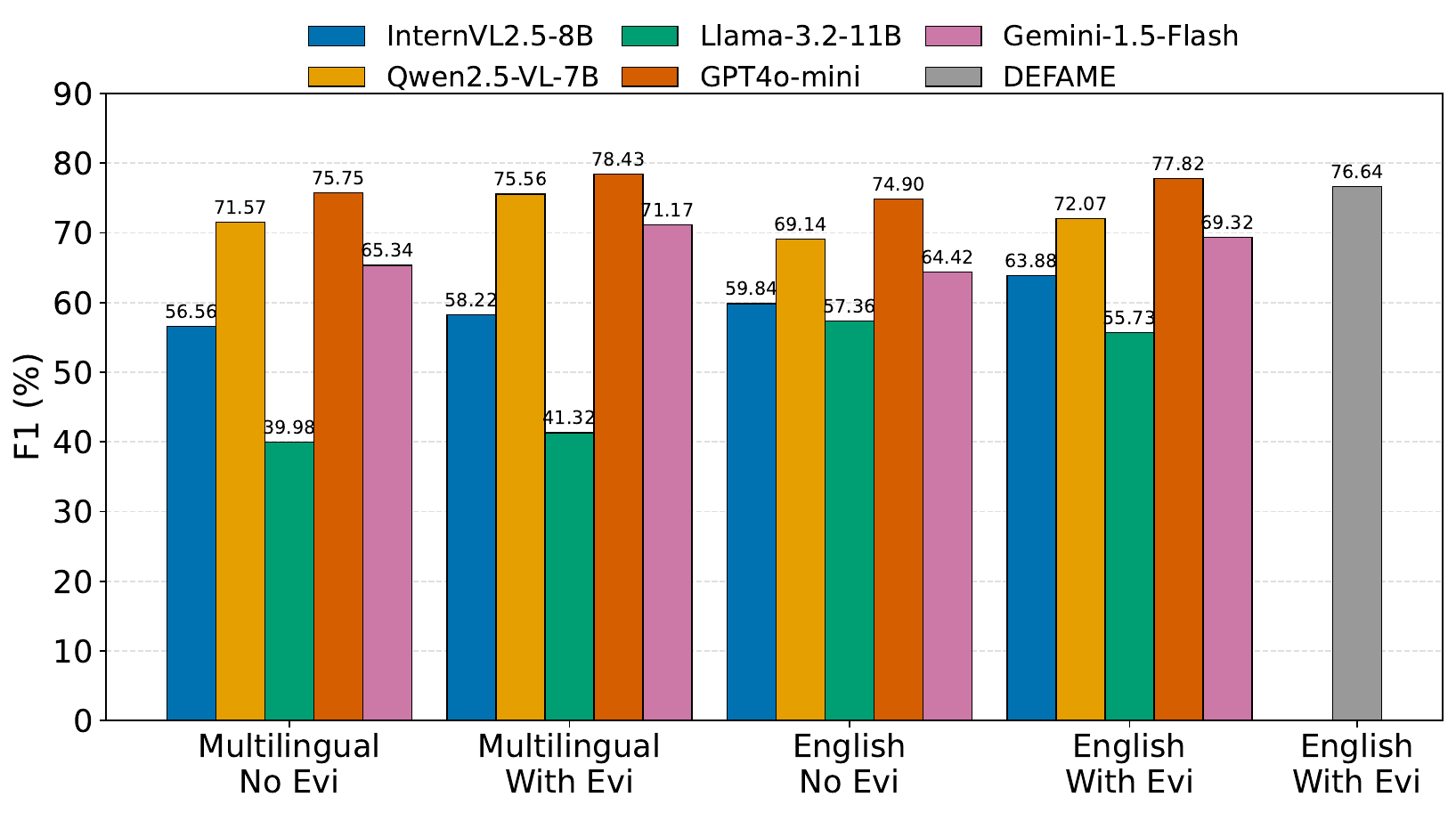}
        \caption{Verdict prediction results, grouped by language setting.}
        \label{fig:vp_vis}
    \end{subfigure}

    \caption{Visual claim extraction (VCE), claimant intent prediction (CIP), fake image detection (FD), and verdict prediction (VP) results (\%). All VP results are shown in the \textit{balanced} setting.}
    \label{fig:2x2-barplots}
\end{figure*}

 \begin{figure}[!t]
    \centering
    \includegraphics[width=\linewidth]{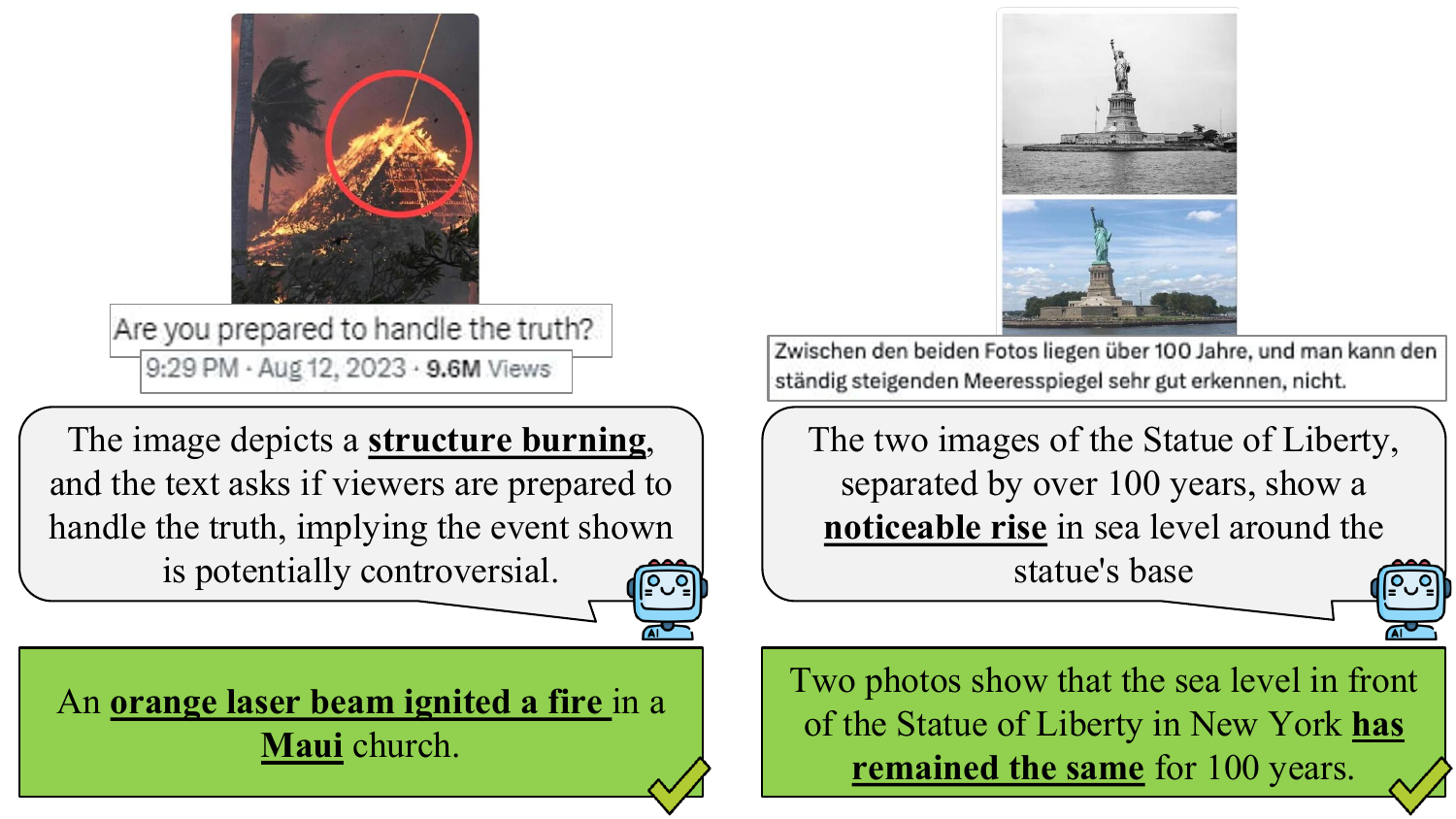}
    \caption{Visual claim extraction error examples with Gemini-1.5-Flash on M4FC test.}
    \label{fig:error-claim-extraction}
\end{figure}

For fake image detection, we evaluate three specialized detectors, Deep-Fake Detector \citep{Prithiv}, AI-Image Detector \citep{Dafilab}, and RealFake Detector \citep{Mikedata}, considering both their pre-trained versions and fine-tuned variants on the MF4C training set. For other tasks, we employ three widely adopted open-weight MLLMs, InternVL2.5-8B \citep{internvl}, Qwen2.5-VL-7B \citep{qwen2.5vl}, and Llama-3.2-11B-Vision \citep{dubey2024llama3}, alongside two proprietary models, GPT4o-mini \citep{openai2023gpt4} and Gemini-1.5-Flash \citep{google2024gemini}, with task-specific prompts provided in Appendix~\ref{sec:prompts_tasks}. In addition, for image contextualization and verdict prediction, we benchmark against the current state-of-the-art (SOTA) methods, COVE \citep{tonglet-etal-2025-cove} and DEFAME \citep{braundefame}, respectively. COVE combines two models: LlavaNext-7B \citep{10.5555/3666122.3667638,Liu_2024_CVPR} and Llama-3-8B \citep{dubey2024llama3}. For DEFAME, we use GPT4o-mini as the LLM. Appendix~\ref{sec:implementation-details} provides implementation details.

\subsection{Baselines results}

We analyze results for all tasks on the test set. Additional results are provided in Appendix \ref{sec:additional_results}.

\paragraph{Visual claim extraction}

 Figure~\ref{fig:vce_vis} shows that most models achieve decent performance, with the exception of Llama-3.2-11B-Vision, while Gemini-1.5-Flash consistently outperforms all others. Rouge-L and METEOR scores show the token-level alignment between generated and reference claims, whereas BERTScore and G-Eval indicate a more moderate level of semantic correspondence. We manually analyzed 50 randomly sampled predictions from Gemini-1.5-Flash. Among them, 38 claims are accurate, nine were partially correct but lacked external world knowledge, and three were incorrect due to the model’s inability to capture sarcasm in social media posts. Figure~\ref{fig:error-claim-extraction} presents two error cases: on the left, the model fails to connect the image with its broader context, although the claim publication date and the landscape hint at a connection with the Maui fires; on the right, the model overlooks the sarcasm and extracts a claim that directly contradicts the ground truth.

\paragraph{Claimant intent prediction}

Figure~\ref{fig:cip_vis} shows that Gemini-1.5-Flash achieves the best performance, with a clear margin on BERTScore and G-Eval compared to other models. GPT4o-mini underperforms because it often refuses to execute the task, flagging claimant intent prediction as a potential policy violation. The overall scores are low,  highlighting that real-world claimant intent prediction remains challenging for current MLLMs.

\paragraph{Fake image detection}

Figure~\ref{fig:fd_vis} reports F1 scores for specialized fake detectors. While the pre-trained Deep-Fake detector achieves a relatively strong F1 of 70.31\%, AI-Image detector is almost ineffective at only 6.41\%. Although the M4FC train set is small and does not overlap in time with the test set, fine-tuning on it yields high performance gains across all models. This means that features can be learned from prior real-world cases and generalized to newer cases.

 \paragraph{Image contextualization}  As shown in Table~\ref{tab:contextualization-baseline}, performance varies substantially across different context items, with no model dominating all categories. Qwen2.5-VL-7B achieves the highest scores on Date (30.3\%), Location (45.0\%), Motivation (16.8\%), and Event (16.2\%). In contrast, Gemini-1.5-Flash demonstrates strength in entity recognition, ranking first in People (45.0\%) and Things (22.7\%).  COVE does not provide better results than most MLLMs. On the contrary, it is outperformed by all MLLMs for Source and Date.  Consistent with prior work, predicting the date of the image is more challenging than predicting its location \citep{tonglet-etal-2024-image,tonglet-etal-2025-cove}.

\begin{table}
\small
    \centering
    \resizebox{0.9\columnwidth}{!}{ %
    \begin{tabular}{cccccccc}
    \toprule 
        & Source & Date & Loc. & Mot. & People & Things & Event  \\
        & (M) & ($\Delta$) & (CO$\Delta$) & (M) & (F1) & (M) & (M) \\
    \midrule
        InternVL-2.5-8B & 4.8 &  21.6 &  26.5 & 9.5 &  24.4 &   13.4 & 8.0  \\
        Qwen2.5-VL-7B & 5.2 & \digitbf{30.3} & \digitbf{45.0} & \digitbf{16.8} & 41.0 & 13.7 & \digitbf{16.2}\\
        Llama-3.2-11B & \digitbf{7.7} & 29.1 &  38.6 & 10.5 &  34.6 & 14.0 & 12.2 \\
        \midrule
        GPT4o-mini & 5.7 & 20.4 & 28.9 & 7.6 & 20.4 & 13.3 & 8.3  \\
        Gemini-1.5-Flash & 7.1 & 23.9 &  42.3 &  16.6 &  \digitbf{45.0} & \digitbf{22.7} & 15.8\\
        \midrule
        COVE (Llama-3-8B) & 1.7 & 17.3  &  33.0 &  16.2 & 34.5  & 12.1 & 14.4  \\
        \bottomrule
    \end{tabular}}
    \caption{Image contextualization (IC) results  (\%). Loc., Mot., and M stand for Location, Motivation, and METEOR, respectively.}
    \label{tab:contextualization-baseline}
\end{table}

\paragraph{Location verification}

\begin{table}
\small
    \centering
    \resizebox{0.8\columnwidth}{!}{ %
    \begin{tabular}{lccccc}
        \toprule
        &
        \includegraphics[height=1em]{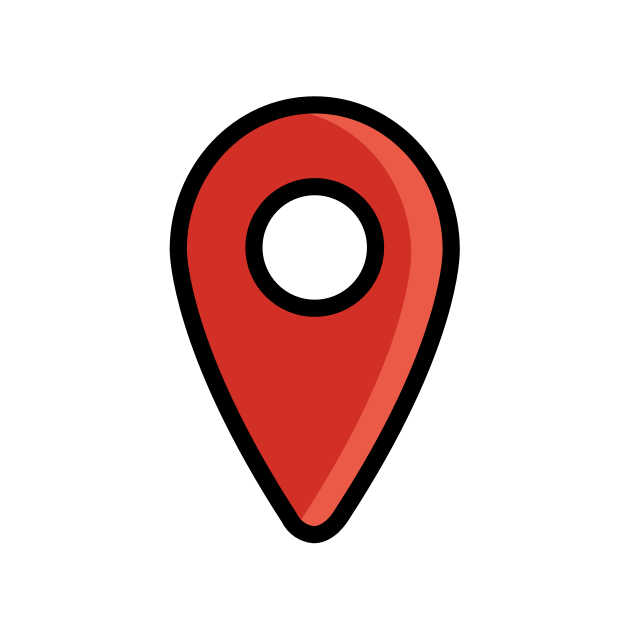}
        & \includegraphics[height=1em]{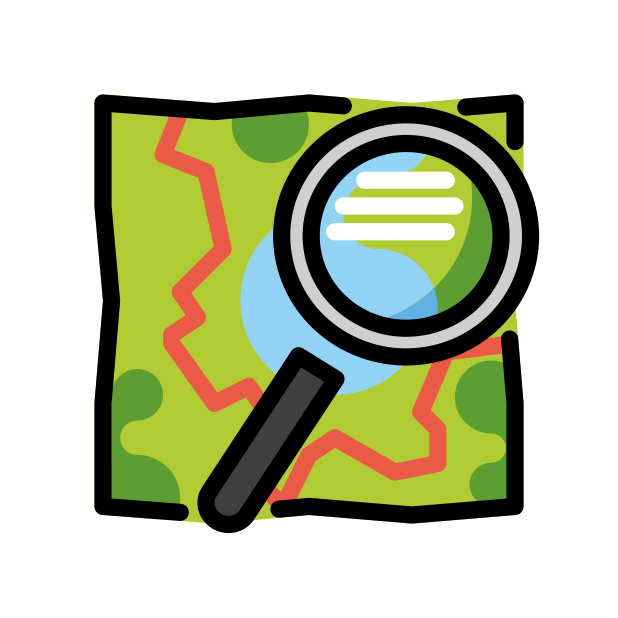} & \includegraphics[height=1em]{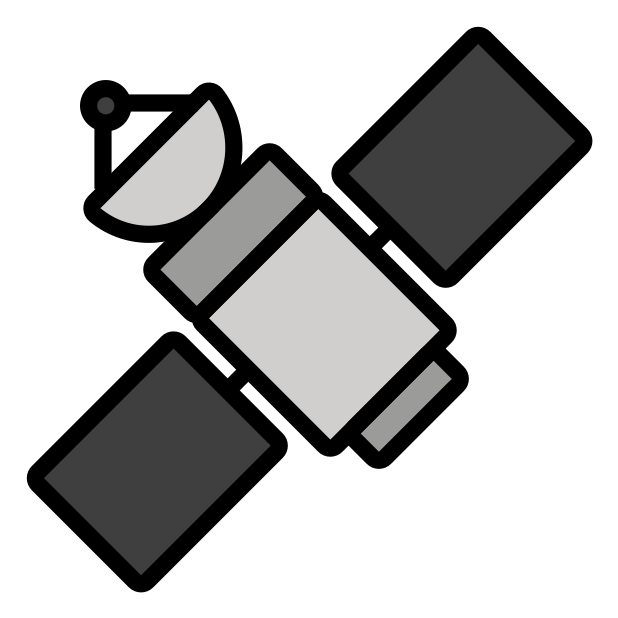} & \includegraphics[height=1em]{figures/OSM.png} \& \includegraphics[height=1em]{figures/satellite.png} \\
        \midrule
        InternVL-2.5-8B & \cellcolor{orange!20}\textbf{62.47} & 25.00 & \cellcolor{lightblue}53.12 &  20.53 \\
        Qwen2.5-VL-7B &  \cellcolor{orange!20}65.19 &58.17  & \cellcolor{lightblue} \textbf{66.59} &  64.20 \\

        Llama-3.2-11B & \cellcolor{orange!20}40.00 & \cellcolor{lightblue} \textbf{56.30} & 44.50&  48.04\\
       \midrule
        GPT4o-mini&  \cellcolor{orange!20} \textbf{67.61}  &  49.95 & 61.50 &  \cellcolor{lightblue} 65.19 \\
                Gemini-1.5-Flash & \cellcolor{orange!20} \textbf{80.83} & \cellcolor{lightblue} 74.43  &  67.52 &   65.62 \\
        \bottomrule
    \end{tabular}}
    \caption{Location verification (LV) F1 scores (\%).  \includegraphics[height=1em]{figures/location.png} is the candidate location as a text, while  \includegraphics[height=1em]{figures/OSM.png} and \includegraphics[height=1em]{figures/satellite.png} are the corresponding map and satellite views.}
    
    \label{tab:location_results}
\end{table}

\begin{figure}[!t]
    \centering
    \includegraphics[width=0.9\linewidth]{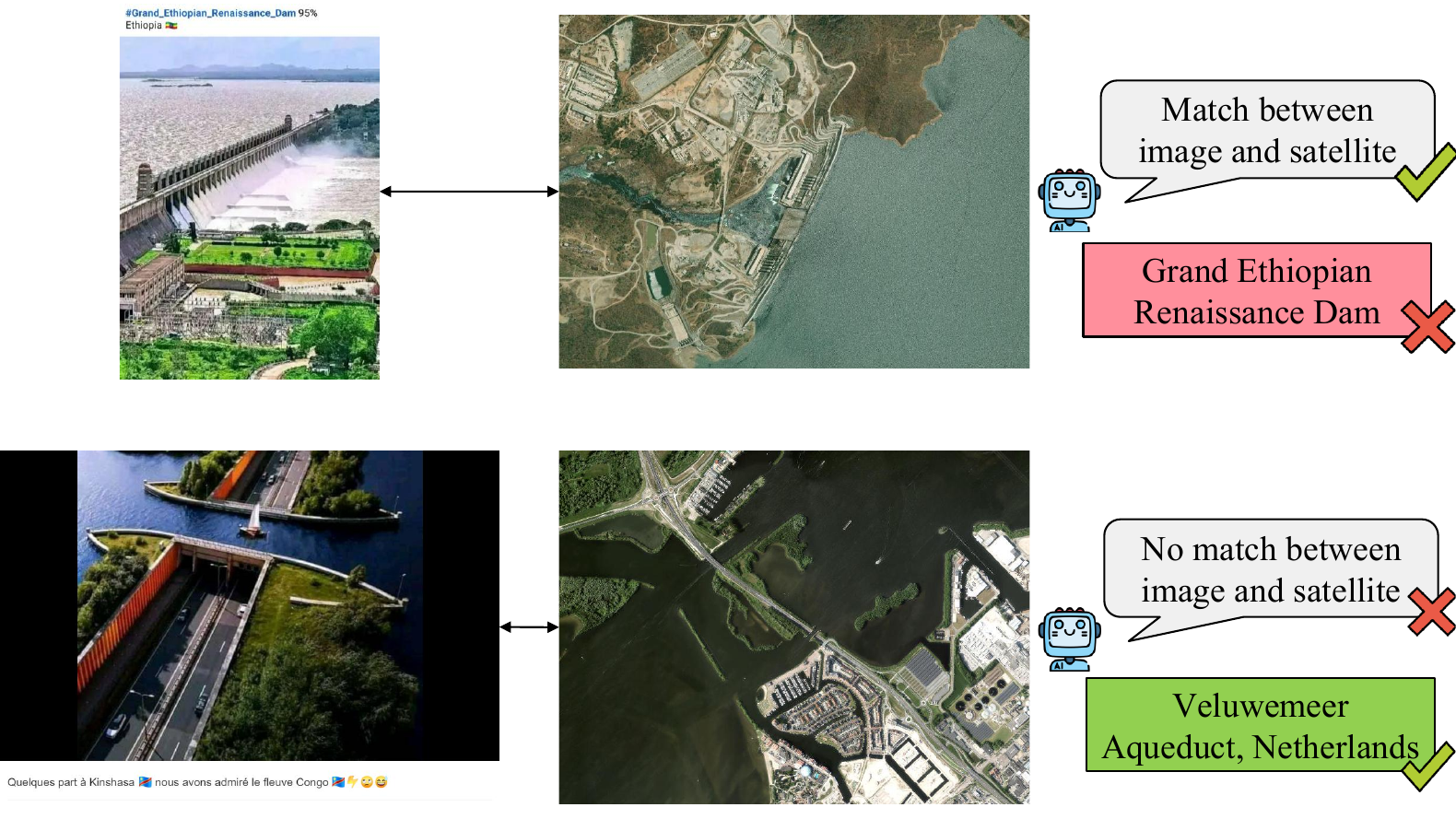}
    \caption{Location verification error examples with Gemini-1.5-Flash on M4FC test. Satellite images are sourced from ESRI World Imagery.}
    \label{fig:error-location-verification}
\end{figure}

We evaluate MLLMs by comparing each claim image with candidate locations in three formats: text, map, and satellite.   As shown in Table~\ref{tab:location_results}, F1 is often lower with aerial views compared to the text baseline.  This suggests that many MLLMs can recognize the location but struggle to match the image to its corresponding aerial view. Model behaviors further diverge by modality: Llama-3.2-11B performs worse than random (49.18\%) when given the candidate location as text, while Qwen2.5-VL-7B achieves its best scores with satellite imagery. Gemini-1.5-Flash consistently leads across all settings, and GPT4o-mini is the only model to benefit from combining map and satellite views. Despite these differences, overall F1 scores remain modest. Error cases in Figure~\ref{fig:error-location-verification} illustrate common failure modes: the dam example shows the model’s difficulty in distinguishing fine-grained structural details, while the aqueduct example highlights limitations of using a fixed zoom level (15), as discussed further in Appendix~\ref{sec:satellite_zoom}.

\begin{table*}[!ht]
  \centering
  \resizebox{0.75\textwidth}{!}{ %
  \begin{tabular}{cccccccccccccc}
    \toprule
    \multicolumn{6}{c}{\textbf{Intermediate Tasks}} &  \multicolumn{4}{c}{\textbf{GPT4o-mini}} &  \multicolumn{4}{c}{\textbf{Qwen2.5-VL-7}B}                 \\
    \cmidrule(lr){1-6}\cmidrule(r){7-10} \cmidrule(r){11-14}
    \textbf{VCE} & \textbf{CIP} & \textbf{FD} & \textbf{IC} & \textbf{LV} & \textbf{ER} & \textbf{Acc} & \textbf{Pre} & \textbf{Rec} & \textbf{F1} & \textbf{Acc} & \textbf{Pre} & \textbf{Rec} & \textbf{F1} \\

    \midrule
    
     \cmark & & & & & & 60.92 & 72.41 & 71.41 & 71.91  & 54.84 & 66.66 & 71.05 & 68.79\\
     \cmarksilver & & & & &  & 53.62 & 73.14 & 53.40 & 61.73 & 49.26 & 64.52 & 58.80 & 61.53 \\
     \cmark & \cmarksilver & & & & & 60.92 & 72.61 & 70.98 & 71.78  & 53.20 & 66.03 & 62.73 & 64.34\\
     \cmark &  & \cmarksilver & &&  &61.31 &73.85 & 70.89 & 72.34 & 51.43 & 65.13 & 60.76 & 62.87 \\
     \cmark & & & \cmarksilver & & & 62.58 & 74.14 & 73.20 & 73.67 & 54.96 & 67.79 & 69.98 & 68.87  \\
     \cmark &  & & \cmarksilver & \cmarksilver & & 62.66 & 74.21 & 73.89 & 74.05 & 55.10 &  68.63 & 69.88 & 69.25 \\
     \cmark &  & &  &  & \cmarksilver & 68.82 & 78.13 & 77.06 & 77.59 & 63.30 & 72.97 & 75.61 & 74.27 \\
     \cmark & \cmarksilver & \cmarksilver & \cmarksilver & \cmarksilver & \cmarksilver & 68.88 & 79.50 & 77.06 & 78.26 & 63.42 & 73.03 & 76.39 & 74.67 \\
    \cmarksilver & \cmarksilver & \cmarksilver & \cmarksilver & \cmarksilver & \cmarksilver & 56.28 & 74.34 & 54.51 & 62.90 & 53.76 & 72.63 & 57.44 & 64.15 \\

    \midrule
    \cmark & \cmark &  &  &  && 61.48 & 72.03 & 73.59& 72.80 & 55.69 & 69.46 & 68.74 & 69.10\\ 
    \cmark &  & \cmark & &  &  & 58.61 & 73.61 & 66.20 & 69.71 & 52.58 & 66.34 & 63.48 & 64.88\\
    \cmark &  &   & \cmark  & & & 84.90 & 91.82 & 86.10 & 88.87 & \digitbf{78.21} & \digitbf{85.79} & \digitbf{82.56} & \digitbf{84.14}\\
    \cmark &  &   & \cmark  & \cmark & & 84.79 & 91.74 & 86.03 & 88.79 & 78.16 & 85.78 & 82.49 & 84.10\\
    \cmark & \cmark &  \cmark & \cmark  &  \cmark &  \cmarksilver & \digitbf{85.98} & 91.38 & \digitbf{87.04} & \digitbf{89.16} & 76.12 & \digitbf{85.79} & 81.42 & 83.55\\
    \midrule
    \midrule
    \multicolumn{6}{c}{DEFAME}  &  81.23  & \digitbf{95.41}  &  79.32 & 86.63 & - & -  & - &  -  \\

    \bottomrule
  \end{tabular}}
  \caption{Verdict prediction (VP) results on M4FC test set (\textit{English-only}, \textit{imbalanced}) by combining intermediate tasks and retrieved evidence (\%). ER: RIS evidence retrieval. \cmarksilver $\,$ and  \cmark $\,$ indicate the inclusion of predicted and ground truth intermediate task outputs and evidence, respectively.}
  \label{tab:pipeline_results}
\end{table*}

\paragraph{Verdict prediction}

Figure~\ref{fig:vp_vis} compares models under both \textit{multilingual} and \textit{English-only} \textit{balanced} settings. Proprietary models exhibit strong linguistic robustness, maintaining comparable performance across multilingual and English claims regardless of RIS evidence availability. By contrast, open-weight models display much larger disparities across languages. InternVL2.5-8B and Llama-3.2-11B, in particular, perform markedly worse on multilingual claims, with Llama-3.2-11B suffering the steepest drop. Interestingly, Qwen2.5-VL-7B shows the opposite trend, achieving slightly better results in multilingual settings. Across all systems, providing RIS evidence consistently improves results, and the effect is most pronounced for models struggling with multilingual inputs. These results highlight substantial variation in cross-lingual generalization: while leading proprietary models achieve near-linguistic parity, many open-weight alternatives remain challenged in multilingual contexts. Overall performance levels also underscore the difficulty of verdict prediction: no model exceeds 80\% F1. The \textit{English-only} performance of GPT4o-mini with DEFAME or with RIS evidence does not differ by much. This suggests that RIS is the most useful source of evidence, and that other evidence search tools introduced by the DEFAME framework, such as web search, image search, and geolocation, have a limited impact on performance.

\begin{figure}
    \centering
    \includegraphics[width=\linewidth]{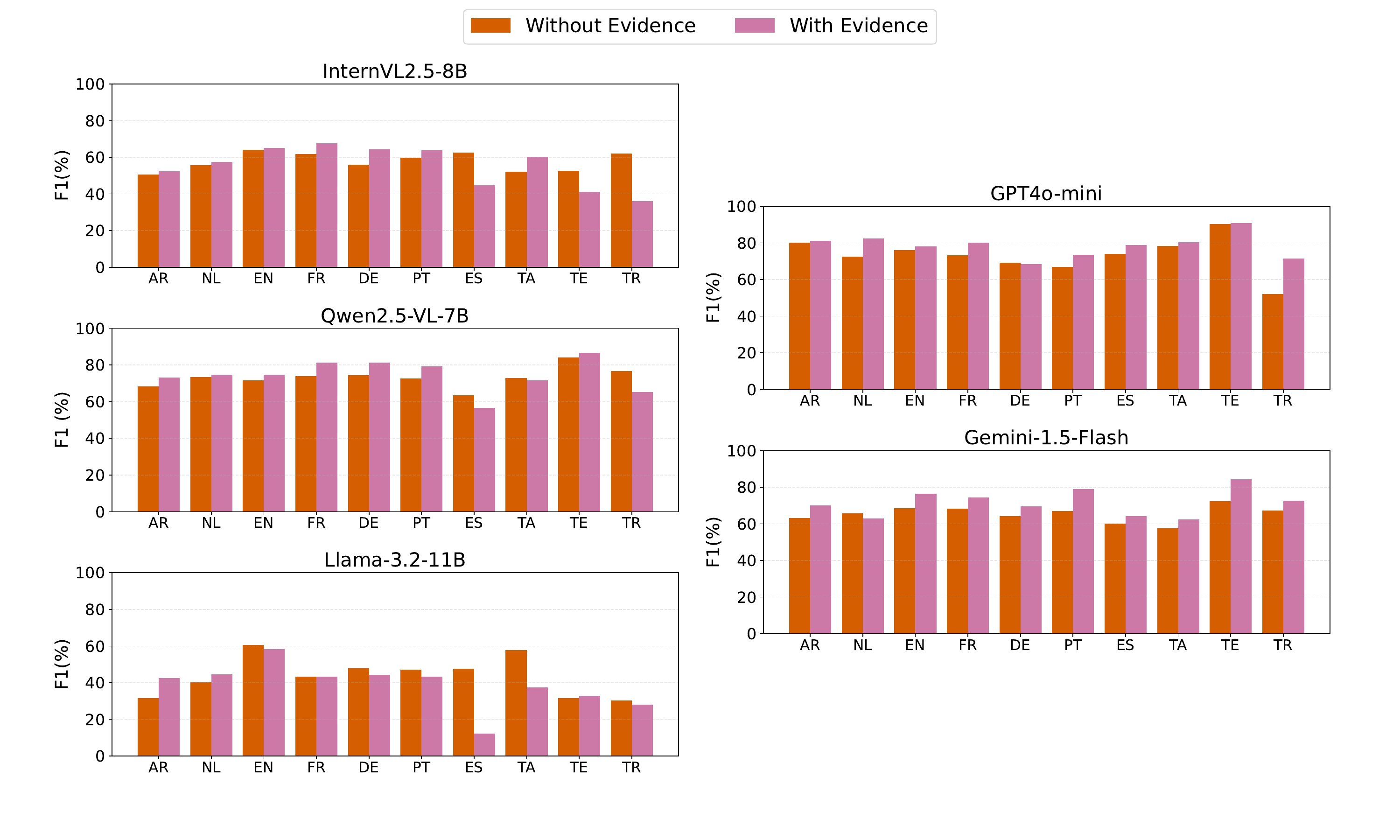}
    \caption{Verdict prediction (VP) F1 scores (\%) across different languages.
    AR: Arabic, 
    NL: Dutch, 
    EN: English, 
    FR: French, 
    DE: German, 
    PT: Portuguese, 
    ES: Spanish, 
    TA: Tamil, 
    TE: Telugu, 
    TR: Turkish.}
    \label{fig:veracity_lang}
\end{figure}

To further analyze \textit{multilingual} performance, We disaggregate results by language in Figure~\ref{fig:veracity_lang}. Proprietary models again lead overall, generally benefiting from RIS evidence. GPT4o-mini performs weakest on Turkish, while Gemini underperforms on Spanish and Tamil. Telugu is the language on which models perform best, surpassing several higher-resource ones. Yet, open-weight models exhibit smaller or inconsistent gains from evidence, and sometimes perform worse when evidence is added, with Telugu ranking among their weakest languages. These findings suggest that 
cross-lingual generalization is highly model-dependent, underscoring the need of multilingual evaluation.

\paragraph{Verdict prediction with intermediate tasks}

Table~\ref{tab:pipeline_results} reports how intermediate task outputs affect verdict prediction. We consider both predicted and ground truth intermediate outputs in the \textit{imbalanced}, \textit{English-only} setting.
We use GPT4o Mini and Qwen2.5-VL-7B, the best performing MLLMs for verdict prediction in the \textit{balanced} setting (see Figures~\ref{fig:vp_vis}). Figure \ref{fig:pipeline} shows an example of pipeline predictions with GPT4o Mini. Replacing ground truth claims with extracted ones substantially decreases F1, showing that inaccuracies in extracted claims harm downstream verdict prediction. Adding all or some of the other intermediate predicted outputs and the web evidence improves F1. Web evidence has the strongest impact. Surprisingly, using only predicted image contextualization, despite it relying on the same web evidence, performs worse than using the web evidence directly.  This suggests that the web evidence is not sufficient to predict accurately most context items during the intermediate image contextualization step. When ground truth intermediate outputs are available, image contextualization yields the largest performance boost, improving F1 by 16.96 percentage points. In that case, adding web evidence provides little to no improvement. The highest F1 is achieved by combining web evidence with all ground truth intermediate tasks, highlighting the complementarity of the intermediate tasks for verdict prediction. These results show the promise of combining AFC tasks and open the door to future work on how best to implement these combinations, in terms of accuracy, interpretability, and computational efficiency.

We compare these results with the end-to-end system DEFAME. DEFAME's F1 remains lower than the pipeline with ground truth intermediate task outputs. However, it largely outperforms the pipeline with predicted intermediate outputs. This shows that, while intermediate tasks are necessary to produce a fact-checking article, SOTA end-to-end systems like DEFAME still have a clear advantage for verdict prediction.

\section{Conclusion}

We introduce M4FC, a real-world dataset for multimodal AFC. The images in M4FC have a broad geographic coverage, representing diverse cultures and events. The dataset supports 10 claim languages and spans six tasks, including two new ones, broadening the scope of multimodal AFC. Our baseline results show that M4FC is a challenging dataset for MLLMs and SOTA AFC methods.

\section*{Limitations}

We identify four limitations to M4FC. 

(1) \textbf{Modality coverage.} M4FC does not cover videos. Videos have become the most prevalent misinformation modality verified by fact-checkers \citep{dufour2024ammeba}. However, obtaining the videos used in real-world multimodal claims is technically and legally challenging. The fact-checking articles usually only include a screenshot, and the videos themselves are often removed from the social media platforms where they appeared.

(2) \textbf{Language distribution.} While M4FC covers substantially more languages than prior datasets, it is limited by the under-representation of low-resource languages in the professional fact-checking landscape. The real-world distribution of fact-checking articles is skewed toward English-speaking organizations, which are more numerous and produce more articles. To partially mitigate this in the evaluation, we augment the test set with synthetic true claims, ensuring that each language is represented by both true and false claims under the \textit{balanced multilingual} setting. 

(3) \textbf{Location verification scale.} The location verification task contains 195 instances, far fewer than the other tasks. Only a subset of instances is suitable for this task: a ground- or aerial-view image of the location. Furthermore, the identification of a precise location is done semi-automatically using a set of keywords with manual validation. While this procedure ensures high precision, it might not achieve perfect recall. Hence, scaling the dataset for this task with real-world cases is not a viable option. Instead, future work should consider creating larger synthetic datasets by systematically collecting outdoor news images paired with their exact coordinates. However, to the best of our knowledge, no such corpus exists.

(4) \textbf{Knowledge leakage risk.} Although we adopt a temporal dataset split and apply evidence filters to mitigate knowledge leakage issues, the risk cannot be fully eliminated. M4FC claims, or the corresponding fact-checking articles, may have leaked into the pretraining data of MLLMs.  Follow-up works, such as VERITAS  \citep{DBLP:journals/corr/abs-2601-08611}, tackle this issue by continuously updating benchmarks with recent claims. In Appendix \ref{sec:knowledge}, we verified that there is no leakage for the five models that we evaluated in our experiments.

\section*{Ethics statement}

\paragraph{Intended use}  This work enables the development of more performant AFC models.  M4FC also contributes to the inclusion of low-resource languages and underrepresented cultural contexts in AFC research. M4FC is intended solely for multimodal AFC research. 

\paragraph{Data access} We collected data from publicly available fact-checking articles. Many images show real-world events, including graphic scenes of violence. Furthermore, the images depict real people, some of whom are identified by name in the annotations. Throughout the annotation process, we did not contact fact-checkers and claimants. While fact-checkers have often hidden account names and profile pictures from images, they did not do it for all of them. We did not take any additional anonymization steps. For these reasons, we do not provide the images directly. Instead, we provide URLs directing to the images and a script to download them. We also provide \textbf{Wayback Machine} URLs to ensure the images remain available in the long term. We follow the policy of the AVeriTeC dataset \citep{10.5555/3666122.3668964}: we will remove instances upon request from the claimants, the people shown in the image, or the authors of the fact-checking article. M4FC is released under a CC BY-SA 4.0 license. The code is released under an Apache 2.0 license.

\paragraph{AI assistants use}  AI assistants were used to correct grammar mistakes and typos.

\section*{Acknowledgments}
This work has been funded by the LOEWE initiative (Hesse, Germany) within the emergenCITY center (Grant Number: LOEWE/1/12/519/03/05.001(0016)/72), the European Union’s Horizon Europe research and innovation programme through the TrustLLM project (Grant Agreement No. 101135671), and the German Federal Ministry of Research, Technology and Space and the Hessian Ministry of Higher Education, Research, Science and the Arts through their joint support of the National Research Center for Applied Cybersecurity ATHENE. The figures have been designed using resources from Flaticon.com. We thank Max Glockner, Shivam Sharma, and Chen Liu for their feedback on an earlier draft of this work.


\bibliography{anthology,custom}

\begin{thebibliography}{55}
\providecommand{\natexlab}[1]{#1}

\bibitem[{Akhtar et~al.(2023)Akhtar, Schlichtkrull, Guo, Cocarascu, Simperl, and Vlachos}]{akhtar-etal-2023-multimodal}
Mubashara Akhtar, Michael Schlichtkrull, Zhijiang Guo, Oana Cocarascu, Elena Simperl, and Andreas Vlachos. 2023.
\newblock \href {https://doi.org/10.18653/v1/2023.findings-emnlp.361} {Multimodal automated fact-checking: A survey}.
\newblock In \emph{Findings of the Association for Computational Linguistics: EMNLP 2023}, pages 5430--5448, Singapore. Association for Computational Linguistics.

\bibitem[{Aneja et~al.(2023)Aneja, Bregler, and Nie{\ss}ner}]{aneja2023cosmos}
Shivangi Aneja, Chris Bregler, and Matthias Nie{\ss}ner. 2023.
\newblock \href {https://doi.org/10.1609/AAAI.V37I12.26648} {{COSMOS:} catching out-of-context image misuse using self-supervised learning}.
\newblock In \emph{Thirty-Seventh {AAAI} Conference on Artificial Intelligence, {AAAI} 2023, Thirty-Fifth Conference on Innovative Applications of Artificial Intelligence, {IAAI} 2023, Thirteenth Symposium on Educational Advances in Artificial Intelligence, {EAAI} 2023, Washington, DC, USA, February 7-14, 2023}, pages 14084--14092. {AAAI} Press.

\bibitem[{Bai et~al.(2025)Bai, Chen, Liu, Wang, Ge, Song, Dang, Wang, Wang, Tang, Zhong, Zhu, Yang, Li, Wan, Wang, Ding, Fu, Xu, Ye, Zhang, Xie, Cheng, Zhang, Yang, Xu, and Lin}]{qwen2.5vl}
Shuai Bai, Keqin Chen, Xuejing Liu, Jialin Wang, Wenbin Ge, Sibo Song, Kai Dang, Peng Wang, Shijie Wang, Jun Tang, Humen Zhong, Yuanzhi Zhu, Ming{-}Hsuan Yang, Zhaohai Li, Jianqiang Wan, Pengfei Wang, Wei Ding, Zheren Fu, Yiheng Xu, and 8 others. 2025.
\newblock \href {https://doi.org/10.48550/ARXIV.2502.13923} {Qwen2.5-vl technical report}.
\newblock \emph{CoRR}, abs/2502.13923.

\bibitem[{Banerjee and Lavie(2005)}]{banerjee-lavie-2005-meteor}
Satanjeev Banerjee and Alon Lavie. 2005.
\newblock \href {https://aclanthology.org/W05-0909/} {{METEOR}: An automatic metric for {MT} evaluation with improved correlation with human judgments}.
\newblock In \emph{Proceedings of the {ACL} Workshop on Intrinsic and Extrinsic Evaluation Measures for Machine Translation and/or Summarization}, pages 65--72, Ann Arbor, Michigan. Association for Computational Linguistics.

\bibitem[{Barbaresi(2021)}]{barbaresi-2021-trafilatura}
Adrien Barbaresi. 2021.
\newblock \href {https://doi.org/10.18653/v1/2021.acl-demo.15} {Trafilatura: {A} web scraping library and command-line tool for text discovery and extraction}.
\newblock In \emph{Proceedings of the 59th Annual Meeting of the Association for Computational Linguistics and the 11th International Joint Conference on Natural Language Processing: System Demonstrations}, pages 122--131, Online. Association for Computational Linguistics.

\bibitem[{Braun et~al.(2025)Braun, Rothermel, Rohrbach, and Rohrbach}]{braundefame}
Tobias Braun, Mark Rothermel, Marcus Rohrbach, and Anna Rohrbach. 2025.
\newblock \href {https://proceedings.mlr.press/v267/braun25b.html} {{DEFAME:} dynamic evidence-based fact-checking with multimodal experts}.
\newblock In \emph{Forty-second International Conference on Machine Learning, {ICML} 2025, Vancouver, BC, Canada, July 13-19, 2025}, Proceedings of Machine Learning Research. {PMLR} / OpenReview.net.

\bibitem[{Cao et~al.(2025)Cao, Ding, Guo, Schlichtkrull, and Vlachos}]{cao2025averimatec}
Rui Cao, Zifeng Ding, Zhijiang Guo, Michael~Sejr Schlichtkrull, and Andreas Vlachos. 2025.
\newblock \href {http://papers.nips.cc/paper\_files/paper/2025/hash/0695cbd1f85f54780e2b351db4252a97-Abstract-Datasets\_and\_Benchmarks\_Track.html} {Averimatec: {A} dataset for automatic verification of image-text claims with evidence from the web}.
\newblock In \emph{Advances in Neural Information Processing Systems 38: Annual Conference on Neural Information Processing Systems 2025, NeurIPS 2025, San Diego, CA, USA, December 2-7, 2025 / Mexico City, Mexico, November 30 - December 5, 2025}.

\bibitem[{Cheema et~al.(2022)Cheema, Hakimov, Sittar, M{\"u}ller-Budack, Otto, and Ewerth}]{cheema-etal-2022-mm}
Gullal~Singh Cheema, Sherzod Hakimov, Abdul Sittar, Eric M{\"u}ller-Budack, Christian Otto, and Ralph Ewerth. 2022.
\newblock \href {https://doi.org/10.18653/v1/2022.findings-naacl.72} {{MM}-claims: A dataset for multimodal claim detection in social media}.
\newblock In \emph{Findings of the Association for Computational Linguistics: NAACL 2022}, pages 962--979, Seattle, United States. Association for Computational Linguistics.

\bibitem[{Chen et~al.(2024)Chen, Wang, Cao, Liu, Gao, Cui, Zhu, Ye, Tian, Liu, Gu, Wang, Li, Ren, Chen, Luo, Wang, Jiang, Wang, He, Shi, Zhang, Lv, Wang, Shao, Chu, Tu, He, Wu, Deng, Ge, Chen, Dou, Lu, Zhu, Lu, Lin, Qiao, Dai, and Wang}]{internvl}
Zhe Chen, Weiyun Wang, Yue Cao, Yangzhou Liu, Zhangwei Gao, Erfei Cui, Jinguo Zhu, Shenglong Ye, Hao Tian, Zhaoyang Liu, Lixin Gu, Xuehui Wang, Qingyun Li, Yimin Ren, Zixuan Chen, Jiapeng Luo, Jiahao Wang, Tan Jiang, Bo~Wang, and 21 others. 2024.
\newblock \href {https://doi.org/10.48550/ARXIV.2412.05271} {Expanding performance boundaries of open-source multimodal models with model, data, and test-time scaling}.
\newblock \emph{CoRR}, abs/2412.05271.

\bibitem[{Da et~al.(2021)Da, Forbes, Zellers, Zheng, Hwang, Bosselut, and Choi}]{da-etal-2021-edited}
Jeff Da, Maxwell Forbes, Rowan Zellers, Anthony Zheng, Jena~D. Hwang, Antoine Bosselut, and Yejin Choi. 2021.
\newblock \href {https://doi.org/10.18653/v1/2021.acl-long.158} {Edited media understanding frames: Reasoning about the intent and implications of visual misinformation}.
\newblock In \emph{Proceedings of the 59th Annual Meeting of the Association for Computational Linguistics and the 11th International Joint Conference on Natural Language Processing (Volume 1: Long Papers)}, pages 2026--2039, Online. Association for Computational Linguistics.

\bibitem[{Dafilab(2025)}]{Dafilab}
Dafilab. 2025.
\newblock Ai image detector.
\newblock \url{https://huggingface.co/Dafilab/ai-image-detector}.
\newblock Accessed: 2025-04-13.

\bibitem[{Dufour et~al.(2024)Dufour, Pathak, Samangouei, Hariri, Deshetti, Dudfield, Guess, Escayola, Tran, Babakar, and Bregler}]{dufour2024ammeba}
Nicholas Dufour, Arkanath Pathak, Pouya Samangouei, Nikki Hariri, Shashi Deshetti, Andrew Dudfield, Christopher Guess, Pablo~Hern{\'{a}}ndez Escayola, Bobby Tran, Mevan Babakar, and Christoph Bregler. 2024.
\newblock \href {https://doi.org/10.48550/ARXIV.2405.11697} {Ammeba: {A} large-scale survey and dataset of media-based misinformation in-the-wild}.
\newblock \emph{CoRR}, abs/2405.11697.

\bibitem[{Geng et~al.(2024)Geng, Kementchedjhieva, Nakov, and Gurevych}]{geng2024multimodallargelanguagemodels}
Jiahui Geng, Yova Kementchedjhieva, Preslav Nakov, and Iryna Gurevych. 2024.
\newblock \href {https://doi.org/10.48550/ARXIV.2403.03627} {Multimodal large language models to support real-world fact-checking}.
\newblock \emph{CoRR}, abs/2403.03627.

\bibitem[{Glockner et~al.(2022)Glockner, Hou, and Gurevych}]{glockner-etal-2022-missing}
Max Glockner, Yufang Hou, and Iryna Gurevych. 2022.
\newblock \href {https://doi.org/10.18653/v1/2022.emnlp-main.397} {Missing counter-evidence renders {NLP} fact-checking unrealistic for misinformation}.
\newblock In \emph{Proceedings of the 2022 Conference on Empirical Methods in Natural Language Processing}, pages 5916--5936, Abu Dhabi, United Arab Emirates. Association for Computational Linguistics.

\bibitem[{He et~al.(2025)He, Li, Huang, Dong, Fei, Zhang, Wu, and Cheng}]{he2025explainablebilingualmultimodalmisinformation}
Yiwei He, Xiangtai Li, Zhenglin Huang, Yi~Dong, Hao Fei, Jiangning Zhang, Baoyuan Wu, and Guangliang Cheng. 2025.
\newblock \href {https://doi.org/10.48550/ARXIV.2506.22930} {Towards explainable bilingual multimodal misinformation detection and localization}.
\newblock \emph{CoRR}, abs/2506.22930.

\bibitem[{Hu et~al.(2023)Hu, Guo, Chen, Wen, and Yu}]{10.1145/3539618.3591896}
Xuming Hu, Zhijiang Guo, Junzhe Chen, Lijie Wen, and Philip~S. Yu. 2023.
\newblock \href {https://doi.org/10.1145/3539618.3591896} {{MR2:} {A} benchmark for multimodal retrieval-augmented rumor detection in social media}.
\newblock In \emph{Proceedings of the 46th International {ACM} {SIGIR} Conference on Research and Development in Information Retrieval, {SIGIR} 2023, Taipei, Taiwan, July 23-27, 2023}, pages 2901--2912. {ACM}.

\bibitem[{Huang et~al.(2024)Huang, Dugan, Yang, and Callison-Burch}]{huang-etal-2024-miragenews}
Runsheng Huang, Liam Dugan, Yue Yang, and Chris Callison-Burch. 2024.
\newblock \href {https://doi.org/10.18653/v1/2024.findings-emnlp.959} {{M}i{RAG}e{N}ews: Multimodal realistic {AI}-generated news detection}.
\newblock In \emph{Findings of the Association for Computational Linguistics: EMNLP 2024}, pages 16436--16448, Miami, Florida, USA. Association for Computational Linguistics.

\bibitem[{Jin et~al.(2017)Jin, Cao, Guo, Zhang, and Luo}]{10.1145/3123266.3123454}
Zhiwei Jin, Juan Cao, Han Guo, Yongdong Zhang, and Jiebo Luo. 2017.
\newblock \href {https://doi.org/10.1145/3123266.3123454} {Multimodal fusion with recurrent neural networks for rumor detection on microblogs}.
\newblock In \emph{Proceedings of the 2017 {ACM} on Multimedia Conference, {MM} 2017, Mountain View, CA, USA, October 23-27, 2017}, pages 795--816. {ACM}.

\bibitem[{Khan et~al.(2025)Khan, Dierickx, Furuly, Vold, Tahseen, Linden, and Dang{-}Nguyen}]{khan2024debunking}
Sohail~Ahmed Khan, Laurence Dierickx, Jan~Gunnar Furuly, Henrik~Brattli Vold, Rano Tahseen, Carl{-}Gustav Linden, and Duc{-}Tien Dang{-}Nguyen. 2025.
\newblock \href {https://doi.org/10.1002/ASI.24970} {Debunking war information disorder: {A} case study in assessing the use of multimedia verification tools}.
\newblock \emph{J. Assoc. Inf. Sci. Technol.}, 76(5):752--769.

\bibitem[{Khan et~al.(2023)Khan, Sheikhi, Opdahl, Rabbi, Stoppel, Trattner, and Dang{-}Nguyen}]{10017287}
Sohail~Ahmed Khan, Ghazaal Sheikhi, Andreas~L. Opdahl, Fazle Rabbi, Sergej Stoppel, Christoph Trattner, and Duc{-}Tien Dang{-}Nguyen. 2023.
\newblock \href {https://doi.org/10.1109/ACCESS.2023.3236993} {Visual user-generated content verification in journalism: An overview}.
\newblock \emph{{IEEE} Access}, 11:6748--6769.

\bibitem[{Lin(2004)}]{lin-2004-rouge}
Chin-Yew Lin. 2004.
\newblock \href {https://aclanthology.org/W04-1013/} {{ROUGE}: A package for automatic evaluation of summaries}.
\newblock In \emph{Text Summarization Branches Out}, pages 74--81, Barcelona, Spain. Association for Computational Linguistics.

\bibitem[{Lin et~al.(2013)Lin, Belongie, and Hays}]{6618964}
Tsung{-}Yi Lin, Serge~J. Belongie, and James Hays. 2013.
\newblock \href {https://doi.org/10.1109/CVPR.2013.120} {Cross-view image geolocalization}.
\newblock In \emph{2013 {IEEE} Conference on Computer Vision and Pattern Recognition, Portland, OR, USA, June 23-28, 2013}, pages 891--898. {IEEE} Computer Society.

\bibitem[{Liu et~al.(2024)Liu, Li, Li, and Lee}]{Liu_2024_CVPR}
Haotian Liu, Chunyuan Li, Yuheng Li, and Yong~Jae Lee. 2024.
\newblock \href {https://doi.org/10.1109/CVPR52733.2024.02484} {Improved baselines with visual instruction tuning}.
\newblock In \emph{{IEEE/CVF} Conference on Computer Vision and Pattern Recognition, {CVPR} 2024, Seattle, WA, USA, June 16-22, 2024}, pages 26286--26296. {IEEE}.

\bibitem[{Liu et~al.(2023{\natexlab{a}})Liu, Li, Wu, and Lee}]{10.5555/3666122.3667638}
Haotian Liu, Chunyuan Li, Qingyang Wu, and Yong~Jae Lee. 2023{\natexlab{a}}.
\newblock \href {http://papers.nips.cc/paper\_files/paper/2023/hash/6dcf277ea32ce3288914faf369fe6de0-Abstract-Conference.html} {Visual instruction tuning}.
\newblock In \emph{Advances in Neural Information Processing Systems 36: Annual Conference on Neural Information Processing Systems 2023, NeurIPS 2023, New Orleans, LA, USA, December 10 - 16, 2023}.

\bibitem[{Liu et~al.(2025)Liu, Li, Li, Huang, Xia, Cui, Huang, Deng, and He}]{liu2025mmfakebenchmixedsourcemultimodalmisinformation}
Xuannan Liu, Zekun Li, Pei{-}Pei Li, Huaibo Huang, Shuhan Xia, Xing Cui, Linzhi Huang, Weihong Deng, and Zhaofeng He. 2025.
\newblock \href {https://openreview.net/forum?id=D6zn6ozJs7} {Mmfakebench: {A} mixed-source multimodal misinformation detection benchmark for lvlms}.
\newblock In \emph{The Thirteenth International Conference on Learning Representations, {ICLR} 2025, Singapore, April 24-28, 2025}. OpenReview.net.

\bibitem[{Liu et~al.(2023{\natexlab{b}})Liu, Iter, Xu, Wang, Xu, and Zhu}]{liu-etal-2023-g}
Yang Liu, Dan Iter, Yichong Xu, Shuohang Wang, Ruochen Xu, and Chenguang Zhu. 2023{\natexlab{b}}.
\newblock \href {https://doi.org/10.18653/v1/2023.emnlp-main.153} {{G}-eval: {NLG} evaluation using gpt-4 with better human alignment}.
\newblock In \emph{Proceedings of the 2023 Conference on Empirical Methods in Natural Language Processing}, pages 2511--2522, Singapore. Association for Computational Linguistics.

\bibitem[{{Llama Team}(2024)}]{dubey2024llama3}
{Llama Team}. 2024.
\newblock \href {https://doi.org/10.48550/ARXIV.2407.21783} {The llama 3 herd of models}.
\newblock \emph{CoRR}, abs/2407.21783.

\bibitem[{Luo et~al.(2021)Luo, Darrell, and Rohrbach}]{luo-etal-2021-newsclippings}
Grace Luo, Trevor Darrell, and Anna Rohrbach. 2021.
\newblock \href {https://doi.org/10.18653/v1/2021.emnlp-main.545} {{N}ews{CLIP}pings: {A}utomatic {G}eneration of {O}ut-of-{C}ontext {M}ultimodal {M}edia}.
\newblock In \emph{Proceedings of the 2021 Conference on Empirical Methods in Natural Language Processing}, pages 6801--6817, Online and Punta Cana, Dominican Republic. Association for Computational Linguistics.

\bibitem[{Mikedata(2025)}]{Mikedata}
Mikedata. 2025.
\newblock Real vs fake image model vit base.
\newblock \url{https://huggingface.co/mikedata/real_vs_fake_image_model_vit_base}.
\newblock Accessed: 2025-04-13.

\bibitem[{Mossou and Higgins(2021)}]{bellingcat2021}
Annique Mossou and Ross Higgins. 2021.
\newblock \href {https://www.bellingcat.com/resources/2021/11/01/a-beginners-guide-to-social-media-verification/} {A beginner's guide to social media verification}.
\newblock Accessed: 2023-09-15.

\bibitem[{M{\"{u}}ller{-}Budack et~al.(2020)M{\"{u}}ller{-}Budack, Theiner, Diering, Idahl, and Ewerth}]{10.1145/3372278.3390670}
Eric M{\"{u}}ller{-}Budack, Jonas Theiner, Sebastian Diering, Maximilian Idahl, and Ralph Ewerth. 2020.
\newblock \href {https://doi.org/10.1145/3372278.3390670} {Multimodal analytics for real-world news using measures of cross-modal entity consistency}.
\newblock In \emph{Proceedings of the 2020 on International Conference on Multimedia Retrieval, {ICMR} 2020, Dublin, Ireland, June 8-11, 2020}, pages 16--25. {ACM}.

\bibitem[{Nielsen and McConville(2022)}]{10.1145/3477495.3531744}
Dan~Saattrup Nielsen and Ryan McConville. 2022.
\newblock \href {https://doi.org/10.1145/3477495.3531744} {Mumin: {A} large-scale multilingual multimodal fact-checked misinformation social network dataset}.
\newblock In \emph{{SIGIR} '22: The 45th International {ACM} {SIGIR} Conference on Research and Development in Information Retrieval, Madrid, Spain, July 11 - 15, 2022}, pages 3141--3153. {ACM}.

\bibitem[{OpenAI(2023)}]{openai2023gpt4}
OpenAI. 2023.
\newblock \href {https://doi.org/10.48550/ARXIV.2303.08774} {{GPT-4} technical report}.
\newblock Technical report.

\bibitem[{{OpenStreetMap contributors}(2017)}]{OpenStreetMap}
{OpenStreetMap contributors}. 2017.
\newblock {Planet dump retrieved from https://planet.osm.org }.
\newblock \url{ https://www.openstreetmap.org }.

\bibitem[{Papadopoulos et~al.(2024)Papadopoulos, Koutlis, Papadopoulos, and Petrantonakis}]{papadopoulos2024verite}
Stefanos{-}Iordanis Papadopoulos, Christos Koutlis, Symeon Papadopoulos, and Panagiotis~C. Petrantonakis. 2024.
\newblock \href {https://doi.org/10.1007/S13735-023-00312-6} {{VERITE:} a robust benchmark for multimodal misinformation detection accounting for unimodal bias}.
\newblock \emph{Int. J. Multim. Inf. Retr.}, 13(1):4.

\bibitem[{Papadopoulos et~al.(2025)Papadopoulos, Koutlis, Papadopoulos, and Petrantonakis}]{10944123}
Stefanos{-}Iordanis Papadopoulos, Christos Koutlis, Symeon Papadopoulos, and Panagiotis~C. Petrantonakis. 2025.
\newblock \href {https://doi.org/10.1109/WACV61041.2025.00544} {Similarity over factuality: Are we making progress on multimodal out-of-context misinformation detection?}
\newblock In \emph{{IEEE/CVF} Winter Conference on Applications of Computer Vision, {WACV} 2025, Tucson, AZ, USA, February 26 - March 6, 2025}, pages 5041--5050. {IEEE}.

\bibitem[{Pei et~al.(2022)Pei, Ananthasubramaniam, Wang, Zhou, Dedeloudis, Sargent, and Jurgens}]{pei-etal-2022-potato}
Jiaxin Pei, Aparna Ananthasubramaniam, Xingyao Wang, Naitian Zhou, Apostolos Dedeloudis, Jackson Sargent, and David Jurgens. 2022.
\newblock \href {https://doi.org/10.18653/v1/2022.emnlp-demos.33} {{POTATO}: The portable text annotation tool}.
\newblock In \emph{Proceedings of the 2022 Conference on Empirical Methods in Natural Language Processing: System Demonstrations}, pages 327--337, Abu Dhabi, UAE. Association for Computational Linguistics.

\bibitem[{Pham et~al.(2024)Pham, Nguyen{-}Nhat, Dinh, Le, Nguyen, Tran, Tran, and Dang{-}Nguyen}]{pham2024ookpik}
Kha{-}Luan Pham, Minh{-}Khoi Nguyen{-}Nhat, Anh{-}Huy Dinh, Quang{-}Tri Le, Manh{-}Thien Nguyen, Anh{-}Duy Tran, Minh{-}Triet Tran, and Duc{-}Tien Dang{-}Nguyen. 2024.
\newblock \href {https://doi.org/10.1007/978-3-031-53302-0\_10} {Ookpik- {A} collection of out-of-context image-caption pairs}.
\newblock In \emph{MultiMedia Modeling - 30th International Conference, {MMM} 2024, Amsterdam, The Netherlands, January 29 - February 2, 2024, Proceedings, Part {IV}}, Lecture Notes in Computer Science, pages 132--144. Springer.

\bibitem[{PrithivMLmods(2024)}]{Prithiv}
PrithivMLmods. 2024.
\newblock Deep-fake-detector-model.
\newblock \url{https://huggingface.co/prithivMLmods/Deep-Fake-Detector-Model}.
\newblock Accessed: 2025-04-13.

\bibitem[{Radford et~al.(2021)Radford, Kim, Hallacy, Ramesh, Goh, Agarwal, Sastry, Askell, Mishkin, Clark, Krueger, and Sutskever}]{pmlr-v139-radford21a}
Alec Radford, Jong~Wook Kim, Chris Hallacy, Aditya Ramesh, Gabriel Goh, Sandhini Agarwal, Girish Sastry, Amanda Askell, Pamela Mishkin, Jack Clark, Gretchen Krueger, and Ilya Sutskever. 2021.
\newblock \href {http://proceedings.mlr.press/v139/radford21a.html} {Learning transferable visual models from natural language supervision}.
\newblock In \emph{Proceedings of the 38th International Conference on Machine Learning, {ICML} 2021, 18-24 July 2021, Virtual Event}, Proceedings of Machine Learning Research, pages 8748--8763. {PMLR}.

\bibitem[{Randolph(2005)}]{randolph2005free}
Justus~J. Randolph. 2005.
\newblock \href {https://eric.ed.gov/?id=ED490661} {Free-marginal multirater kappa (multirater k [free]): An alternative to fleiss' fixed-marginal multirater kappa.}
\newblock In \emph{Joensuu Learning and Instruction Symposium}. ERIC.

\bibitem[{Reid et~al.(2024)Reid, Savinov, Teplyashin, Lepikhin, Lillicrap, Alayrac, Soricut, Lazaridou, Firat, Schrittwieser, Antonoglou, Anil, Borgeaud, Dai, Millican, Dyer, Glaese, Sottiaux, Lee, Viola, Reynolds, Xu, Molloy, Chen, Isard, Barham, Hennigan, McIlroy, Johnson, Schalkwyk, Collins, Rutherford, Moreira, Ayoub, Goel, Meyer, Thornton, Yang, Michalewski, Abbas, Schucher, Anand, Ives, Keeling, Lenc, Haykal, Shakeri, Shyam, Chowdhery, Ring, Spencer, Sezener, and et~al.}]{google2024gemini}
Machel Reid, Nikolay Savinov, Denis Teplyashin, Dmitry Lepikhin, Timothy~P. Lillicrap, Jean{-}Baptiste Alayrac, Radu Soricut, Angeliki Lazaridou, Orhan Firat, Julian Schrittwieser, Ioannis Antonoglou, Rohan Anil, Sebastian Borgeaud, Andrew~M. Dai, Katie Millican, Ethan Dyer, Mia Glaese, Thibault Sottiaux, Benjamin Lee, and 34 others. 2024.
\newblock \href {https://doi.org/10.48550/ARXIV.2403.05530} {Gemini 1.5: Unlocking multimodal understanding across millions of tokens of context}.
\newblock Technical report.

\bibitem[{Rothermel et~al.(2026)Rothermel, Kornmann, Rohrbach, and Rohrbach}]{DBLP:journals/corr/abs-2601-08611}
Mark Rothermel, Marcus Kornmann, Marcus Rohrbach, and Anna Rohrbach. 2026.
\newblock \href {https://aclanthology.org/2026.acl-long.1948/} {Veritas: The first dynamic benchmark for multimodal automated fact-checking}.
\newblock In \emph{Proceedings of the 64th Annual Meeting of the Association for Computational Linguistics (Volume 1: Long Papers), {ACL} 2026, San Diego, California, United States, July 2-7, 2026}, pages 42066--42100. Association for Computational Linguistics.

\bibitem[{Sabir et~al.(2018)Sabir, AbdAlmageed, Wu, and Natarajan}]{10.1145/3240508.3240707}
Ekraam Sabir, Wael AbdAlmageed, Yue Wu, and Prem Natarajan. 2018.
\newblock \href {https://doi.org/10.1145/3240508.3240707} {Deep multimodal image-repurposing detection}.
\newblock In \emph{2018 {ACM} Multimedia Conference on Multimedia Conference, {MM} 2018, Seoul, Republic of Korea, October 22-26, 2018}, pages 1337--1345. {ACM}.

\bibitem[{Schlichtkrull et~al.(2023)Schlichtkrull, Guo, and Vlachos}]{10.5555/3666122.3668964}
Michael~Sejr Schlichtkrull, Zhijiang Guo, and Andreas Vlachos. 2023.
\newblock \href {http://papers.nips.cc/paper\_files/paper/2023/hash/cd86a30526cd1aff61d6f89f107634e4-Abstract-Datasets\_and\_Benchmarks.html} {Averitec: {A} dataset for real-world claim verification with evidence from the web}.
\newblock In \emph{Advances in Neural Information Processing Systems 36: Annual Conference on Neural Information Processing Systems 2023, NeurIPS 2023, New Orleans, LA, USA, December 10 - 16, 2023}.

\bibitem[{Shao et~al.(2024)Shao, Wu, Wu, Nie, and Liu}]{10440475}
Rui Shao, Tianxing Wu, Jianlong Wu, Liqiang Nie, and Ziwei Liu. 2024.
\newblock \href {https://doi.org/10.1109/TPAMI.2024.3367749} {Detecting and grounding multi-modal media manipulation and beyond}.
\newblock \emph{{IEEE} Trans. Pattern Anal. Mach. Intell.}, 46(8):5556--5574.

\bibitem[{Silverman(2013)}]{silverman2013verification}
Craig Silverman. 2013.
\newblock \href {https://verificationhandbook.com/} {Verification handbook: A definitive guide to verifying digital content for emergency coverage}.
\newblock Accessed: 2023-09-15.

\bibitem[{Tonglet et~al.(2026)Tonglet, Gurevych, Tuytelaars, and Moens}]{tonglet2026newsrecon}
Jonathan Tonglet, Iryna Gurevych, Tinne Tuytelaars, and Marie{-}Francine Moens. 2026.
\newblock \href {https://doi.org/10.48550/ARXIV.2601.14121} {Newsrecon: News article retrieval for image contextualization}.
\newblock \emph{CoRR}, abs/2601.14121.

\bibitem[{Tonglet et~al.(2024)Tonglet, Moens, and Gurevych}]{tonglet-etal-2024-image}
Jonathan Tonglet, Marie-Francine Moens, and Iryna Gurevych. 2024.
\newblock \href {https://doi.org/10.18653/v1/2024.emnlp-main.448} {{\textquotedblleft}image, tell me your story!{\textquotedblright} predicting the original meta-context of visual misinformation}.
\newblock In \emph{Proceedings of the 2024 Conference on Empirical Methods in Natural Language Processing}, pages 7845--7864, Miami, Florida, USA. Association for Computational Linguistics.

\bibitem[{Tonglet et~al.(2025)Tonglet, Thiem, and Gurevych}]{tonglet-etal-2025-cove}
Jonathan Tonglet, Gabriel Thiem, and Iryna Gurevych. 2025.
\newblock \href {https://aclanthology.org/2025.naacl-long.102/} {{COVE}: {CO}ntext and {VE}racity prediction for out-of-context images}.
\newblock In \emph{Proceedings of the 2025 Conference of the Nations of the Americas Chapter of the Association for Computational Linguistics: Human Language Technologies (Volume 1: Long Papers)}, pages 2029--2049, Albuquerque, New Mexico. Association for Computational Linguistics.

\bibitem[{Wolf et~al.(2020)Wolf, Debut, Sanh, Chaumond, Delangue, Moi, Cistac, Rault, Louf, Funtowicz, Davison, Shleifer, von Platen, Ma, Jernite, Plu, Xu, Le~Scao, Gugger, Drame, Lhoest, and Rush}]{wolf-etal-2020-transformers}
Thomas Wolf, Lysandre Debut, Victor Sanh, Julien Chaumond, Clement Delangue, Anthony Moi, Pierric Cistac, Tim Rault, Remi Louf, Morgan Funtowicz, Joe Davison, Sam Shleifer, Patrick von Platen, Clara Ma, Yacine Jernite, Julien Plu, Canwen Xu, Teven Le~Scao, Sylvain Gugger, and 3 others. 2020.
\newblock \href {https://doi.org/10.18653/v1/2020.emnlp-demos.6} {Transformers: State-of-the-art natural language processing}.
\newblock In \emph{Proceedings of the 2020 Conference on Empirical Methods in Natural Language Processing: System Demonstrations}, pages 38--45, Online. Association for Computational Linguistics.

\bibitem[{Wu et~al.(2026)Wu, Li, Fu, Kan, and Hooi}]{wu2025seeingdeceptionuncoveringmisleading}
Jiaying Wu, Fanxiao Li, Zihang Fu, Min-Yen Kan, and Bryan Hooi. 2026.
\newblock \href {https://openreview.net/forum?id=02NbD16OnA} {Seeing through deception: Uncovering misleading creator intent in multimodal news with vision-language models}.
\newblock In \emph{The Fourteenth International Conference on Learning Representations, {ICLR} 2026, Rio de Janeiro, April 23-27, 2026}.

\bibitem[{Xiao et~al.(2025)Xiao, Han, Wang, and Jiang}]{xiao2025xfactacontemporaryrealworlddataset}
Yuzhuo Xiao, Zeyu Han, Yuhan Wang, and Huaizu Jiang. 2025.
\newblock \href {https://doi.org/10.48550/ARXIV.2508.09999} {Xfacta: Contemporary, real-world dataset and evaluation for multimodal misinformation detection with multimodal llms}.
\newblock \emph{CoRR}, abs/2508.09999.

\bibitem[{Zhang et~al.(2020)Zhang, Kishore, Wu, Weinberger, and Artzi}]{bert-score}
Tianyi Zhang, Varsha Kishore, Felix Wu, Kilian~Q. Weinberger, and Yoav Artzi. 2020.
\newblock \href {https://openreview.net/forum?id=SkeHuCVFDr} {Bertscore: Evaluating text generation with {BERT}}.
\newblock In \emph{8th International Conference on Learning Representations, {ICLR} 2020, Addis Ababa, Ethiopia, April 26-30, 2020}. OpenReview.net.

\bibitem[{Zlatkova et~al.(2019)Zlatkova, Nakov, and Koychev}]{zlatkova-etal-2019-fact}
Dimitrina Zlatkova, Preslav Nakov, and Ivan Koychev. 2019.
\newblock \href {https://doi.org/10.18653/v1/D19-1216} {Fact-checking meets fauxtography: Verifying claims about images}.
\newblock In \emph{Proceedings of the 2019 Conference on Empirical Methods in Natural Language Processing and the 9th International Joint Conference on Natural Language Processing (EMNLP-IJCNLP)}, pages 2099--2108, Hong Kong, China. Association for Computational Linguistics.

\end{thebibliography}

\appendix

\section{Verdict prediction settings biases}
\label{sec:biases}

\paragraph{Imbalanced English-only setting}

In this default setting, all claims appear as in the fact-checking articles. This setting reflects the real-world distribution of claims verified by fact-checkers and can be used for both model training and evaluation. 

\paragraph{Balanced English-only setting}

The balanced English-only setting replaces several false claims with automatically generated true claims based on the true contextual attributes of the image, namely its date, location, event, and depicted people. As a result, the generated true claims typically contain more named entities than real-world claims verified by fact-checkers. In addition, the generated true claims have an average length of 109 characters, whereas real claims in the dataset have an average length of 77 characters. These systematic linguistic differences introduce biases that trained classifiers could exploit as superficial heuristics. For this reason, we do not recommend using this setting for model training.

However, we verify that these biases do not affect model evaluation. Table \ref{tab:balanced-biases} compares the verdict prediction performance of two MLLMs under the balanced and imbalanced English-only settings. The resulting differences remain within 3 percentage points. This suggests that the balanced setting is not less challenging than the imbalanced one as an evaluation setting.

\begin{table}
\small
    \centering
    \resizebox{\columnwidth}{!}{ %
    \begin{tabular}{lcccc}
        \toprule
        & Web evidence &  Balanced & Imbalanced & Delta \\ 
        \midrule
Qwen2.5VL-7B &   & \digitbf{69.1} &  68.8 & 0.3\\
           & \cmarksilver  &  72.1 &  \digitbf{74.3} & 2.2 \\
GPT4o-mini &   & \digitbf{74.9} &   71.9 & 3.0 \\
           & \cmarksilver  &  \digitbf{77.8} & 77.6 & 0.2 \\
        
        \bottomrule
    \end{tabular}}
    \caption{\textit{English-only} verdict prediction performance comparison between the \textit{balanced} and \textit{imbalanced} settings on M4FC test set, measured with F1 (\%). Delta is the absolute difference in F1 score.}
    
    \label{tab:balanced-biases}
\end{table}

\paragraph{Imbalanced multilingual setting}

In the \textit{multilingual} \textit{imbalanced} setting, most true claims are contained exclusively within two languages, English and Portuguese, while other languages only contain false claims, like Arabic. This is because many non-English speaking fact-checking organizations never write articles about true claims. As a result, a trained model could use the claim's language as a heuristic for verdict prediction. For this reason, this setting should not be used for model training.

\paragraph{Balanced multilingual setting}

In the \textit{multilingual} \textit{balanced} setting, both true and false claims are represented for each of the 10 languages. However, the stylistic biases of the synthetic true claims still occur. For this reason, this setting should not be used for model training.

\section{Distribution of fact-checking organizations}
\label{sec:fc_org_dist}

Table \ref{tab:fc_org} shows the distribution of the fact-checking articles by organizations of the IFCN.

\begin{table}
    \small
    \centering
    \resizebox{\columnwidth}{!}{ %
    \begin{tabular}{llll}
    \toprule
     Organization    &   Country & Language & \# articles \\
    \midrule
      factly.in   & India & English, Kannada, Telugu & 1027 (20.6\%)\\
      fatabyyano.net & Jordan & Arabic & 658 (13.2\%) \\
      pesacheck.org & Kenya & English, French & 561 (11.3\%) \\
      snopes.com & USA & English & 454 (9.1\%)\\
      indiatoday.in & India  & English & 333 (6.7\%)\\
      leadstories.com & USA  & English & 328 (6.6\%)\\
      poligrafo.sapo.pt & Portugal  & Portuguese & 284 (5.7\%)\\
      congocheck.net & DRC & French & 260 (5.2\%)\\
      larepublica.pe & Peru  & Spanish & 253 (5.1\%)\\
      correctiv.org & Germany & German & 220 (4.4\%)\\
      dfrac.org & India & English & 104 (2.1\%)\\
      knack.be  & Belgium & Dutch & 95 (1.9\%)\\
      mythdetector.ge &  Georgia & English & 67 (1.3\%) \\
      dubawa.org & Nigeria  & English & 65 (1.3\%)\\
      youturn.in & India & Tamil & 59 (1.2\%)\\
      dogrula.org & Turkey & Turkish & 52 (1.0\%)\\
      211check.org & South Sudan  & English & 51 (1.0\%)\\
      verify-sy.com  & Turkey & Arabic & 42 (0.8\%)\\
      tjekdet.dk & Denmark  & Danish & 29 (0.6\%)\\
      factcheckcenter.jp & Japan & Japanese & 18 (0.4\%)\\
      factchecker.gr & Greece &  Greek & 17 (0.3\%)\\
      factchecklab.org  & Hong Kong & Chinese & 5 (0.1\%)\\
      
    \bottomrule
    \end{tabular}}
    \caption{M4FC data source distribution.} 
    \label{tab:fc_org}
\end{table}

\section{Labeling prompts}
\label{sec:prompts_labeling}

Table~\ref{tab:prompt_annotation} presents the prompt instructions provided to GPT4o to extract task labels and metadata from the fact-checking articles.
Table \ref{tab:prompt_postprocess} shows the prompt instructions provided to GPT4o for the post-processing steps.

\begin{table*}
\centering
\resizebox{0.95\textwidth}{!}{
\begin{tabular}{p{0.97\textwidth}}
\toprule
Please extract the following information from the debunking article and provide the response in the exact format below. Each item should be in one row, and if any information is unavailable, return "None". Do not output anything else. \\[0.5em]

\quad Verification strategy: [What verification strategies were used to check the misinformation? (e.g., reverse image search, keyword search, geolocation, deepfake detection, satellite imagery, street view, etc.)] \\[0.5em]
\quad Verification tools: [What tools were utilized for verification? (e.g., Google, Yandex, Google Earth, etc.)] \\[0.5em]
\quad Claim: [What is the claim accompanying the image?] \\[0.5em]
\quad Claimant: [Who made the claim?] \\[0.5em]
\quad Claimant’s Intent: [What is the claimant's intent or agenda behind making the claim?] \\[0.5em]
\quad Language of the FC article: [In which language is the fact-checking article written?] \\[0.5em]
\quad Language of the claim: [In which language is the claim accompanying the image written?] \\[0.5em]
\quad Gold evidence URLs: [List all URLs that link to external webpage evidence.] \\[0.5em]
\quad Gold image evidence URLs: [List all URLs linking to image evidence.] \\[0.5em]
\quad Claimed date: [According to the claim, when was the image taken?] \\[0.5em]
\quad Claimed location: [According to the claim, where was the image taken?] \\[0.5em]
\quad Claimed people: [Who are the people shown in the image, according to the claim?] \\[0.5em]
\quad  Claimed things: [What objects, buildings, plants, or animals are depicted in the image, according to the claim?] \\[0.5em]
 \quad  Claimed event: [What event is being shown in the image, according to the claim?] \\[0.5em]
\quad Was the image used before? (Provenance): [Did the fact-checkers retrieve a previous version of the same image? (Answer with Yes, No, or Unknown)] \\[0.5em]
\quad Source: [Which person or organization first took or published the image, according to the fact-check? This is different from the claimant who published the claim with the image.] \\[0.5em]
\quad Date: [When was the image taken?] \\[0.5em]
\quad Location: [Where was the image actually taken?] \\[0.5em]
\quad Motivation: [Why was the image taken?] \\[0.5em]
\quad People: [Who is shown in the actual image?] \\[0.5em]
\quad Things: [What objects, buildings, plants, or animals are shown in the actual image?] \\[0.5em]
\quad Event: [What event is depicted in the actual image?] \\[0.5em]

Here is the debunking article: \texttt{\{Article\}} \\

\bottomrule
\end{tabular}
}
\caption{GPT4o prompt template to extract task labels and metadata from fact-checking articles.}
\label{tab:prompt_annotation}
\end{table*}

\begin{table*}
\centering
\resizebox{0.95\textwidth}{!}{
\begin{tabular}{p{0.97\textwidth}}
\toprule
\textbf{Post-processing step:} Normalize dates to YYYY-MM-DD format \\
\textbf{Prompt:} Normalize the following dates to YYYY if it is a year, to YYYY-MM if it is a year and month, or YYYY-MM-DD if it is a specific day. \\ 
If there is more than one date in the text, repeat the operation for each date, separating them with ``;'' . \\
Answer only with the normalized date. \\
Date: \texttt{\{DATE\}} \\
Normalized date:  \\
\midrule
\textbf{Post-processing step:} Normalizing claims \\
\textbf{Prompt:} Normalize the following claim about an image by keeping only the actual content. \\
Remove information about the claimant, unnecessary information such as ``the image shows'', and any words indicating the veracity of the claim, such as ``viral'', ``accurately'', ``purported''. \\
Keep all information that is part of the claim itself.\\
Answer only with the normalized claim text. \\
Claim: \texttt{\{CLAIM\}} \\
Normalized claim:  \\

\midrule
\textbf{Post-processing step:} Extracting claim in the original language \\
\textbf{Prompt:}  You are given a fact-checking article debunking a claim, and the translation of this claim in English. Provide the claim as it appears in its original language in the article text. Remove information about the claimant, unnecessary information such as ``the image shows'', and any words indicating the veracity of the claim, such as ``viral'', ``accurately'', ``purported'' (in the original language).\\
Answer only with the claim in the original language. \\
Fact-checking article:  \texttt{\{ARTICLE\}} \\
Claim: \texttt{\{CLAIM\}} \\
Claim in the original language:  \\
\midrule
\textbf{Post-processing step:} Identify original language \\
\textbf{Prompt:}  In which language is the following claim written. Answer only in one word with the corresponding language.\\
Claim: \texttt{\{CLAIM\}} \\
Language:  \\
\midrule
\textbf{Post-processing step:} Generate true claims for out-of-context images \\
\textbf{Prompt:}  You are given context information describing an image. Combine this information in one caption of maximum 30 words. \\
Write the facts only, avoid journalistic style and adjectives, avoid introducing new information.\\
Date: \texttt{\{DATE\}} \\
Location: \texttt{\{LOCATION\}} \\
People: \texttt{\{PEOPLE\}} \\
Event: \texttt{\{EVENT\}} \\
Caption:  \\
\midrule
\textbf{Post-processing step:} Generate incorrect locations for location verification \\
\textbf{Prompt:} Predict the 5 most distinct, specific and probable locations for this news image.\\
Distinct: locations should be geographically distant.\\
Specific: locations should be, by order of preference: buildings or nature landmarks, streets, villages, small towns or neighborhoods.\\
Locations should not be located within the same city.\\
Ignore the social media text shown around the picture.\\
Answer only with the 5 locations as a | separated list.\\
List: \\
\bottomrule
\end{tabular}
}
\caption{GPT4o prompt templates to post-process M4FC data.}
\label{tab:prompt_postprocess}
\end{table*}

\section{Data collection and labeling details}
\label{sec:data_collection}

\paragraph{Data scraping}

\begin{figure}
    \centering
    \includegraphics[width=\linewidth]{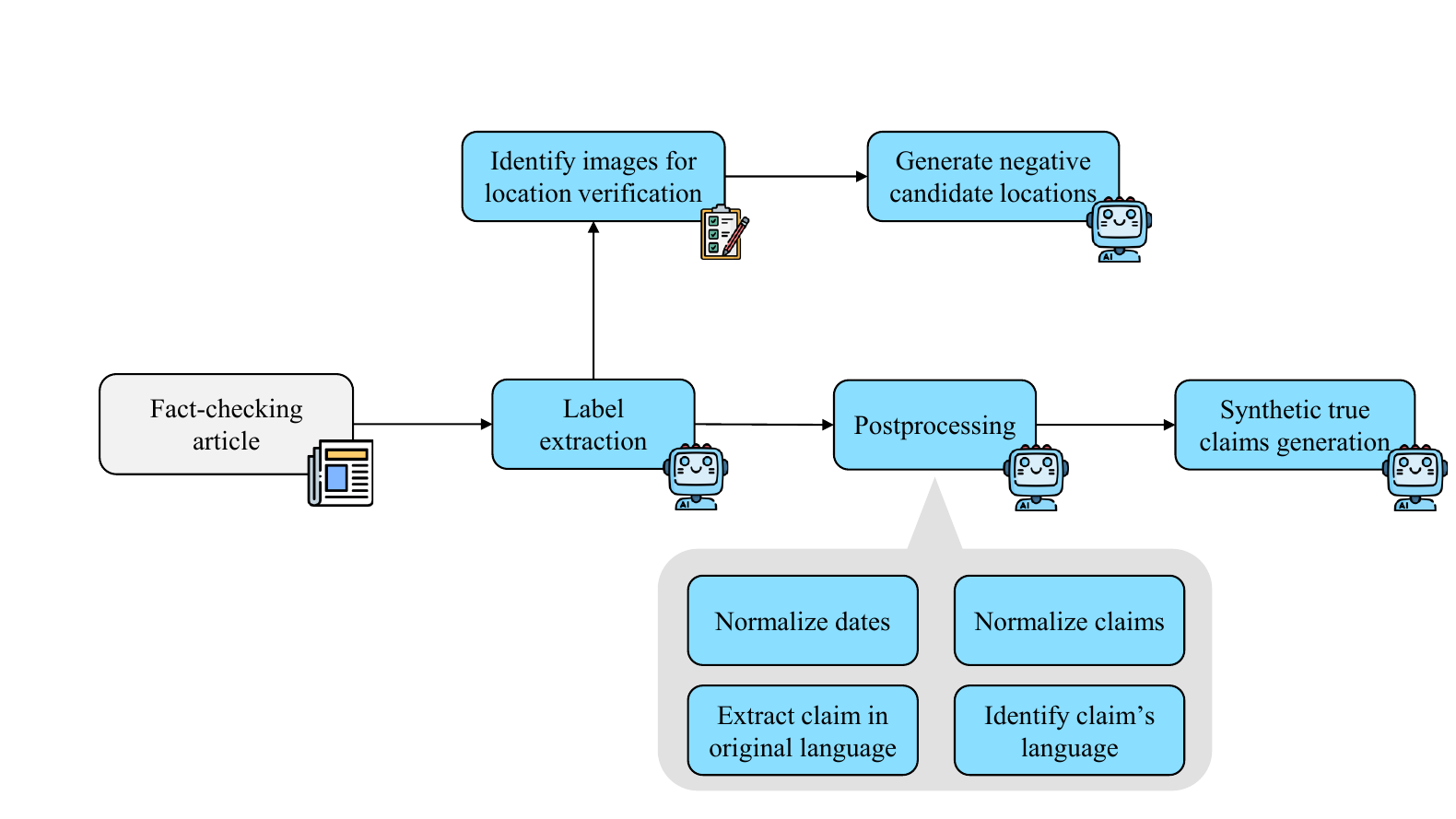}
    \caption{Data extraction and post-processing workflow.}
    \label{fig:data_collection}
\end{figure}

\begin{figure}
    \centering
    \includegraphics[width=\linewidth]{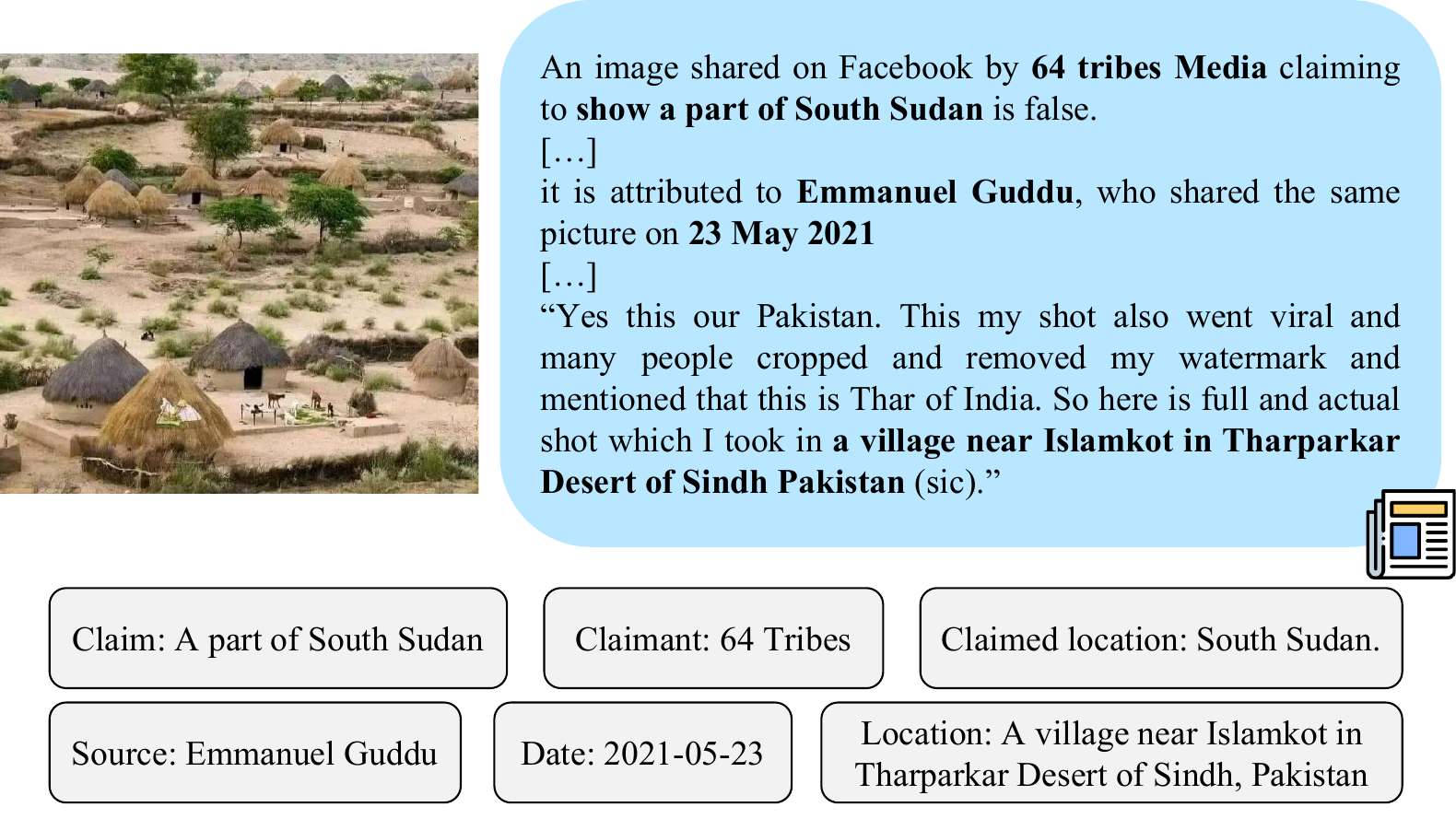}
    \caption{Example of label extraction with GPT4o from a fact-checking article.}
    \label{fig:example_extraction}
\end{figure}

Following the approach of \citet{tonglet-etal-2024-image}, we retrieved all URLs archived on the Wayback Machine\footnote{\href{https://archive.org/help/wayback_api.php}{archive.org/help/wayback\_api.php}} under each organization's web domain. At the time of data collection, all included organizations were current members of the IFCN or under review for membership renewal. We retained URLs containing keywords such as \textit{photo}, \textit{image}, or \textit{picture}, translated into the relevant language. From the raw HTML files, we extracted the article title, publication date, text content, and the URL of the fact-checked image using custom Python scripts. All image URLs were manually verified and corrected when necessary. We manually removed false positives (e.g., non-multimodal claims or fact-checking tutorials), articles containing fact-checking annotations on the image (e.g., an image overlaid with a red cross by the fact-checkers to indicate that it is fake), and duplicate image-claim pairs across organizations. The final dataset includes 4,982 articles.

\paragraph{GPT4o extraction and post-processing}

The data extraction and post-processing workflow is illustrated in Figure \ref{fig:data_collection}. All GPT4o extracted labels are generated in English and follow a predefined dictionary format. Task labels and metadata that are not available in the article are annotated as ``not enough information''. One example is provided in Figure \ref{fig:example_extraction}. Additional post-processing steps with GPT4o include: (1) normalizing all dates to the YYYY-MM-DD format, (2) rewriting claims to remove any words revealing the verdict, and (3) extracting the claim in its original language in the article and identifying that language. For the \textit{multilingual} verdict prediction setting, we consider languages represented with at least 40 claims; otherwise, we default to the English translation of the claim.
 All verdicts are mapped to the following categories: true, false, mixture, and unverified.  The latter two, which account for only 27 instances, are excluded from verdict prediction. False claims are further categorized into: false (out-of-context), false (manipulated/fake),\footnote{Excluding AI-generated images.} and false (AI-generated).

\paragraph{Generating synthetic true claims}

For each authentic image labeled as false (out-of-context), we generate a synthetic true claim using GPT4o, combining the following context items, if at least one is available: date, location, people, and event.  In the \textit{balanced} verdict setting, we replace the original false claim with the corresponding synthetic true claim in half of the instances where a true claim was successfully generated. The prompt is provided in Appendix \ref{sec:prompts_labeling}.

\paragraph{Temporal dataset split}

The dataset is split temporally into train (60\%), dev (10\%), and test (30\%) sets, using cutoff dates of 2022-09-22, 2023-01-18, and 2024-09-01, respectively, to avoid temporal leakage \citep{glockner-etal-2022-missing}.

\begin{figure*}
    \centering
    \includegraphics[width=\linewidth]{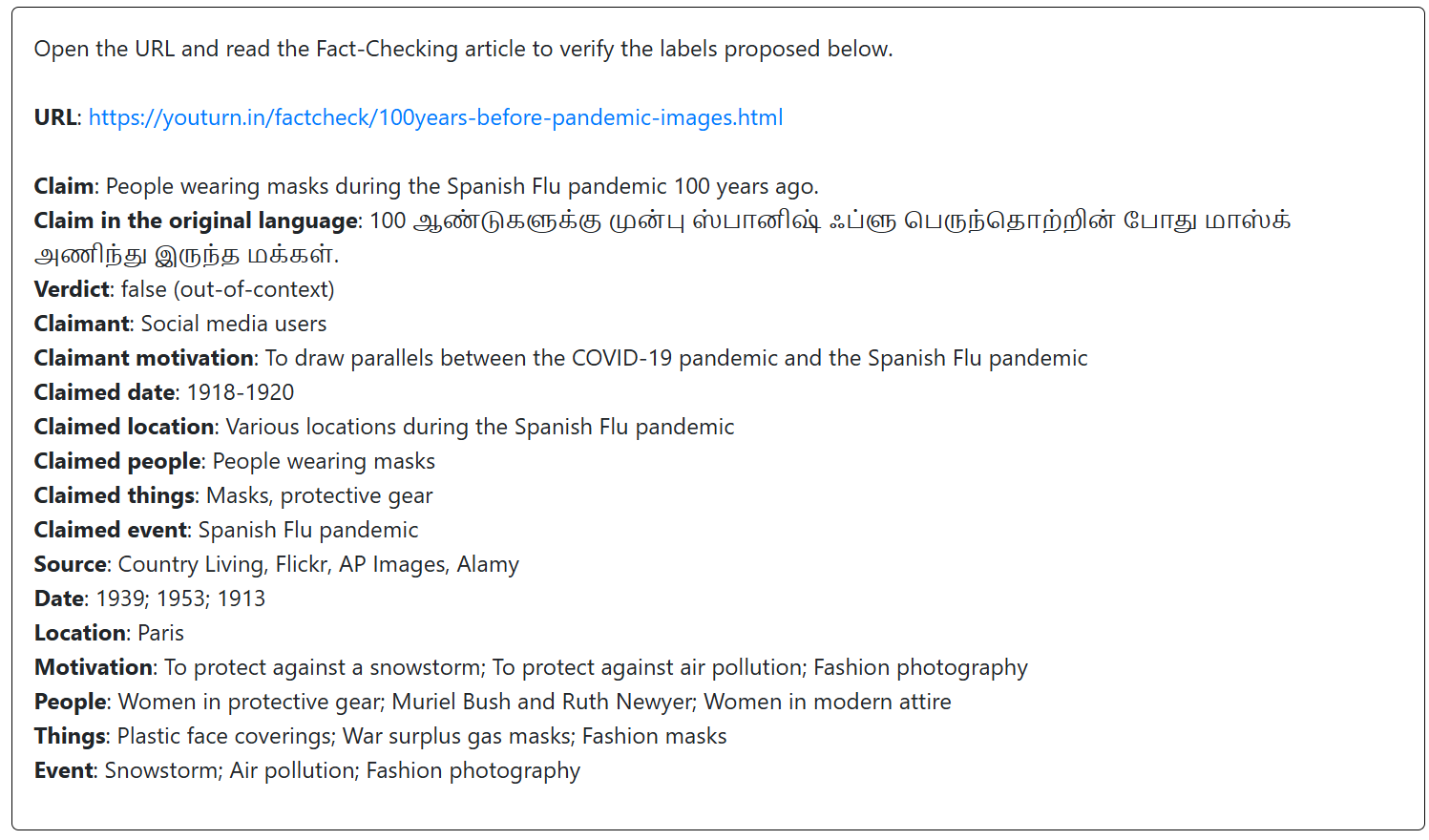}
    \caption{An example instance provided to the Prolific annotators.}
    \label{fig:prolific-1}
\end{figure*}
\begin{figure}[!ht]
    \centering
    \includegraphics[width=\columnwidth]{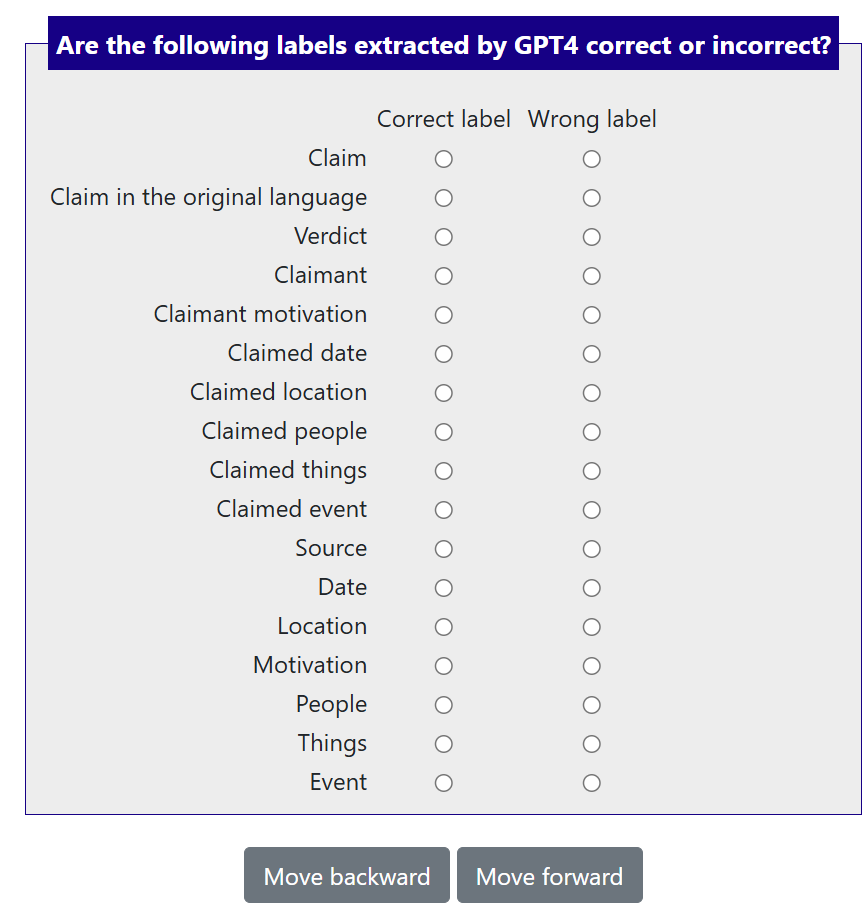}
    \caption{The annotators' interface to validate the labels extracted by GPT4o from fact-checking articles.}
    \label{fig:prolific-2}
\end{figure}

\section{Data validation - Prolific study}

We recruited annotators on the Prolific platform to validate the quality of the data extracted by GPT4o from fact-checking articles. Annotators were paid at the default rate of £9 per hour. We applied the following criteria to select two annotators per study: they need to (1) be fluent speaker of English and their primary language needs to be the one used by the fact-checking organization, (2) have no language-related disorders, (3) have an approval rate of 99\% and  (4) have already participated in 50 annotation tasks in the past.

\subsection{Annotation guidelines}
\label{sec:annotation_guidelines}

We used the Potato annotation tool \citep{pei-etal-2022-potato} to design the data validation interface. Each instance consists of the URL of the fact-checking article and the labels extracted by GPT4o, as shown in Figure \ref{fig:prolific-1}. For each label, annotators have to select either ``correct label'' or ``wrong label'', as shown in Figure \ref{fig:prolific-2}. Cases where the label is ``not enough information'' and the annotator selects ``wrong label'' are automatically mapped to ``missing label'', while others are mapped to ``incorrect label''.

\subsection{Percentage agreement and IAA scores}
\label{sec:iaa}

Table \ref{tab:iaa} provides the detailed percentages of correct, incorrect, and missing labels, and the  IAA scores computed with Randolph's $\kappa$. We distinguish between labels marked as incorrect or missing. A label is marked as missing when GPT4o outputs ``not enough information'', even though relevant information is available.  Results are provided for each label type and averaged across all 23 Prolific studies. These results exclude the attention tests. For the percentage of correct, incorrect, and missing labels, we report both the lower bound, where a label is considered correct only if both annotators agree on it, and the upper bound, where an answer is considered correct if at least one annotator agrees on it.  IAA is reported both with and without accounting for cases where the label is ``not enough information'' (NEI). The scores indicate a clear moderate to high IAA. The highest IAA with NEI is 0.81 for the verdict label. For comparison, the IAA between human annotators for verdict in the AVerImaTeC dataset \citep{cao2025averimatec} is 0.74.

Later, we manually verified the entire test set and identified 0.7\% of incorrect labels, which aligns with the upper bound for percentage agreement. This also means that the lower bound substantially overestimates the number of incorrect labels and that annotators often mistakenly flagged labels as incorrect. We discuss some case studies of errors made by Prolific annotators in the next section.

\begin{table*}
    \centering
    \begin{tabular}{ccccccccc}
    \toprule
    Label   & \multicolumn{2}{c}{Correct (\%)} & \multicolumn{2}{c}{Incorrect (\%)} & \multicolumn{2}{c}{Missing (\%)}  &  \multicolumn{2}{c}{Randolph's $\kappa$} \\
    \cline{2-9}
    &  Low & Up & Low & Up & Low & Up & w/o NEI & w/ NEI \\
    \midrule
    \cellcolor{lightblue}  Claim  (English)  &  83.2 & 98.8 & 16.8 & 1.2 & 0.0& 0.0 & 0.70 & 0.70  \\
      \cellcolor{lightblue} Claim (original language) & 66.7 & 90.6 &  21.9 & 1.1 & 11.5 & 8.3 &  0.52 & 0.53 \\
      \cellcolor{lightblue} Verdict & 89.6 & 99.3 & 10.4 & 0.7 & 0.0 & 0.0 & 0.81 & 0.81  \\
      \cellcolor{orange!20} Claimant & 82.0 & 98.4 & 12.2 & 0.5 & 5.8 & 1.2 & 0.74 & 0.67 \\
      \cellcolor{lightblue} Claimant motivation & 73.3 & 98.4 & 6.5 & 0.9 & 20.3 & 1.2 & 0.78 & 0.51 \\
      \cellcolor{orange!20} Claimed date & 49.6 & 86.2 & 7.4 & 1.5 & 43.0 &  12.3 & 0.62  & 0.32  \\
      \cellcolor{orange!20} Claimed location & 79.6 & 97.6 & 9.0 & 0.5 & 11.4 & 1.9 & 0.75 & 0.65 \\
      \cellcolor{orange!20} Claimed people & 85.0 & 98.3 & 4.8 & 0.0 & 10.1 & 1.7 & 0.83 & 0.72 \\
      \cellcolor{orange!20} Claimed things & 77.2 & 94.7 & 5.3 & 0.9 & 17.5 & 4.4 & 0.85 & 0.65 \\
      \cellcolor{orange!20} Claimed event & 81.2 & 98.1 & 8.9 & 1.0 & 9.9 & 1.0 & 0.80 & 0.67 \\
      \cellcolor{lightblue} Source & 63.7 & 93.9 & 20.0 & 2.9 & 16.4 & 3.2 & 0.45 & 0.37 \\
      \cellcolor{lightblue} Date & 64.0 & 97.4 &  20.9 & 0.8  & 15.1 & 1.8 & 0.51 & 0.41 \\
      \cellcolor{lightblue} Location & 76.4 & 97.3 & 11.3 & 0.2 &  12.3 & 2.4 & 0.67 & 0.57 \\
      \cellcolor{lightblue} Motivation & 74.8 & 99.3  & 12.6 & 0.0 & 12.6 & 0.7  & 0.66 & 0.56\\ 
      \cellcolor{lightblue} People & 79.1 & 99.3 & 8.5& 0.0 &  12.4 & 0.7 & 0.75 & 0.61 \\
      \cellcolor{lightblue} Things & 80.6 & 97.1 & 12.1 & 0.7 & 13.6 & 2.2 & 0.82 & 0.64 \\
      \cellcolor{lightblue} Event & 74.3 & 96.6 & 11.0 & 0.9 & 13.3 & 2.5 & 0.67 & 0.56\\
      \midrule 
      Total & 75.8 & 96.6 & 11.0 & 0.9 & 13.3 & 2.5 & - &  - \\
    \bottomrule
    \end{tabular}
    \caption{Detailed results of the Prolific data validation study. \colorbox{lightblue}{Task labels} and \colorbox{orange!20}{metadata labels} are highlighted in blue and orange, respectively. Low and Up indicate the lower and upper bound scores, respectively.}
    \label{tab:iaa}
\end{table*}

\subsection{Data validation case studies}
\label{sec:disagreement}

\begin{figure}
    \centering
    \includegraphics[width=\columnwidth]{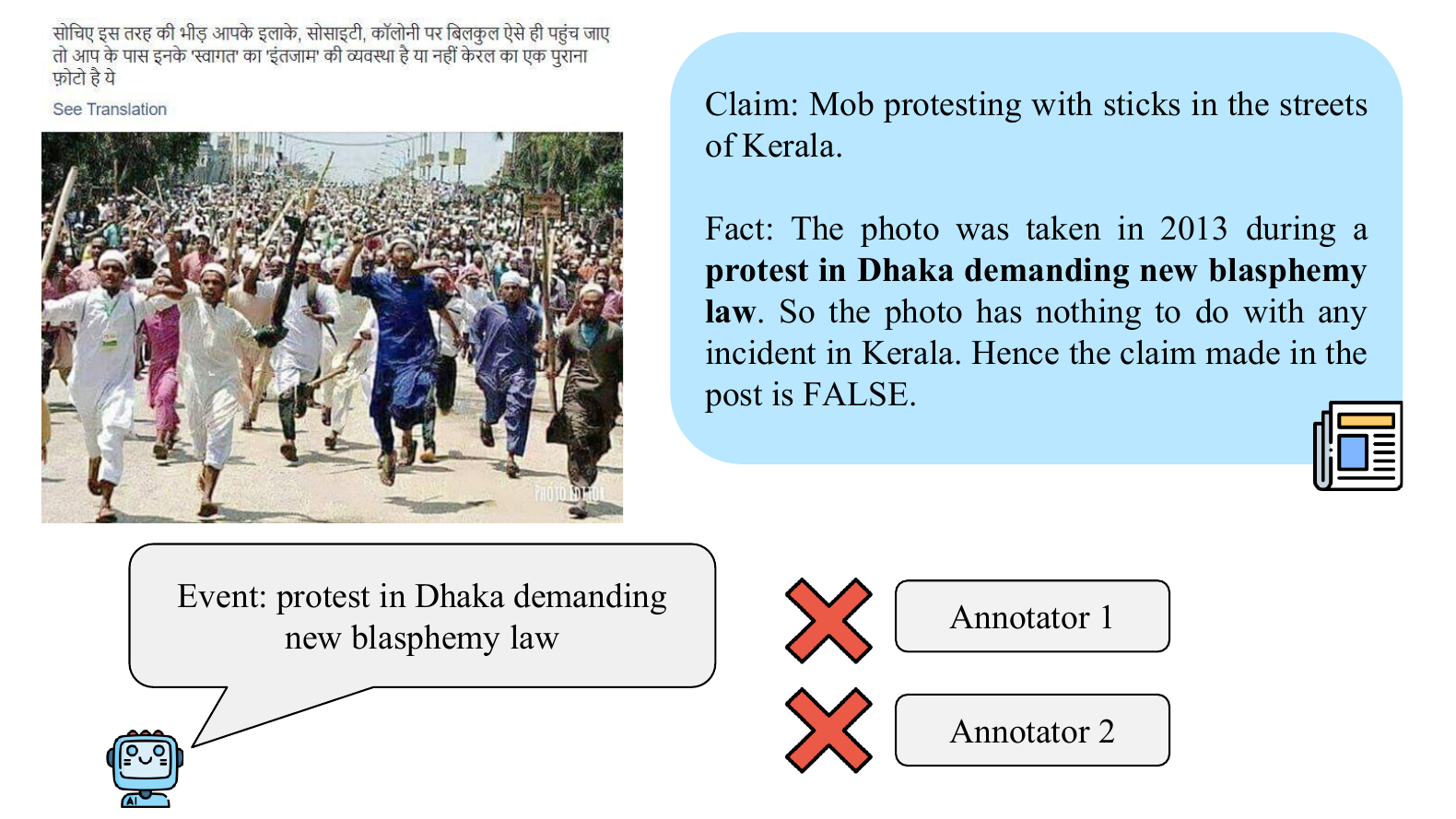}
    \caption{The label extracted by GPT4o is correct, but both annotators flag it as incorrect.}
    \label{fig:prolific1}
\end{figure}

\begin{figure}
    \centering
    \includegraphics[width=\columnwidth]{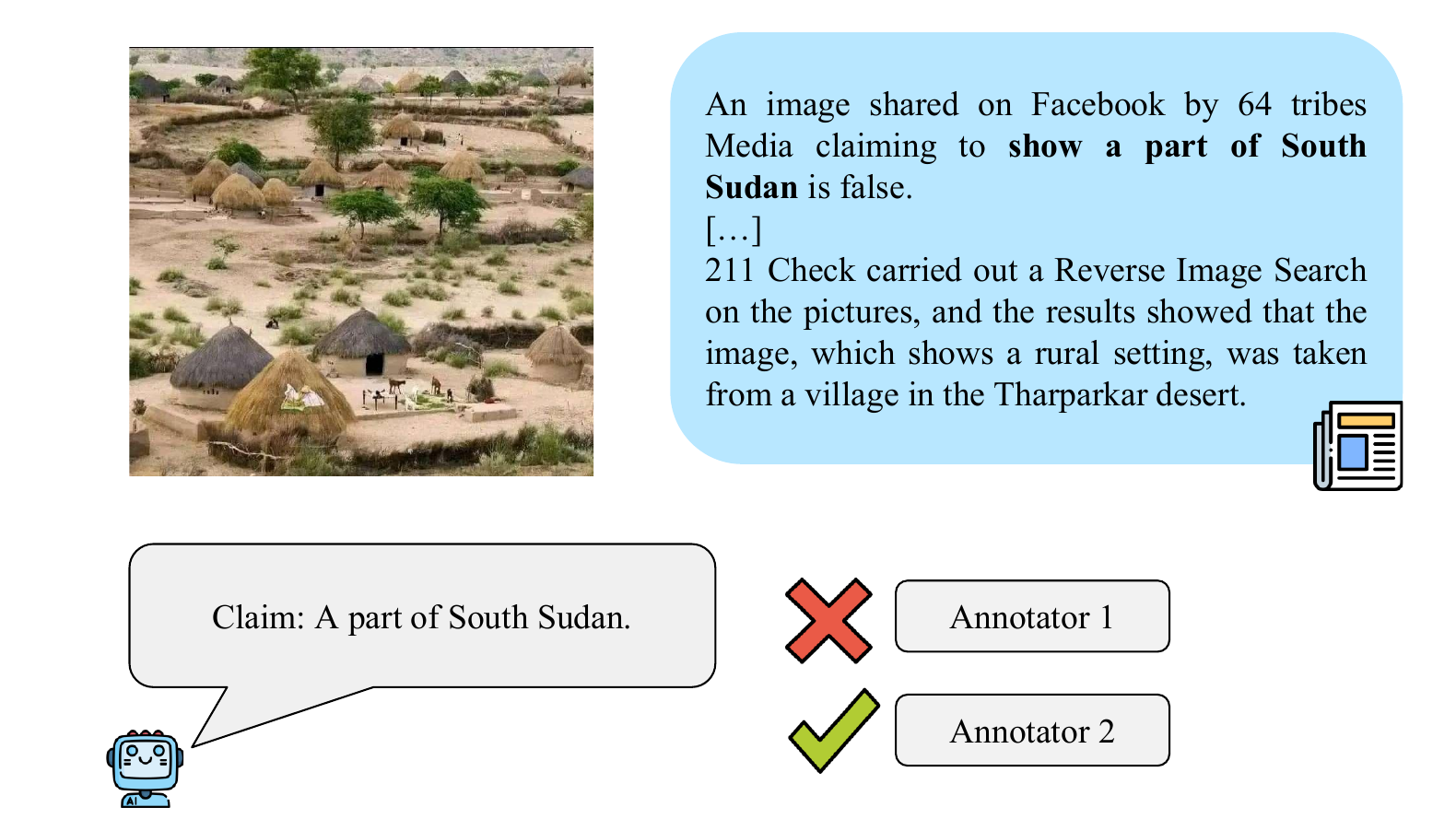}
    \caption{The label extracted by GPT4o is correct, but one annotator flags it as incorrect.}
    \label{fig:prolific3}
\end{figure}

Figures \ref{fig:prolific1} and \ref{fig:prolific3} show examples of correct labels extracted by GPT4o and reviewed by Prolific annotators. In Figure \ref{fig:prolific1}, both annotators flag the label as incorrect. We assume that annotators often confuse the claimed event in the image, in this case a protest in Kerala, with the actual event in the image, a protest in Dhaka. In Figure \ref{fig:prolific3}, one annotator flags the extracted claim as incorrect, while it appears clearly from the fact-checking article that GPT4o extracted the claim accurately.

To be considered for the validation study, all annotators had to pass challenging attention tests; hence, we do not attribute these errors to sloppiness. Instead, we assume these errors are due to the difficulty of the annotation task for a human annotator. This further motivates the use of GPT4o for label extraction, as it appears to better understand the task than human annotators.

\section{Additional dataset statistics}
\label{sec:dataset_statistics}

\begin{figure}
    \centering
    \includegraphics[width=\columnwidth]{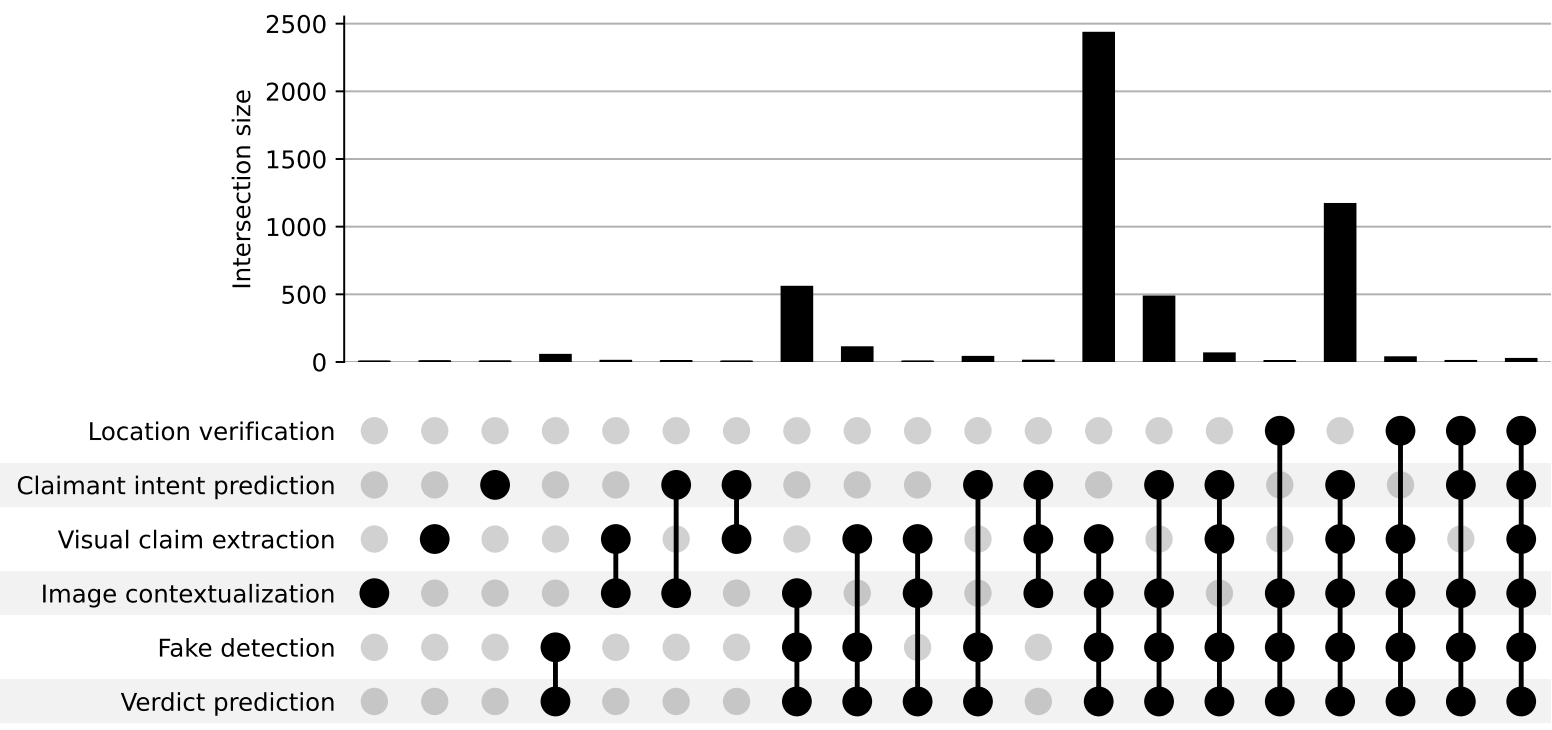}
    \caption{Number of M4FC instances with labels for different task combinations.}
    \label{fig:upset_plot}
\end{figure}

\begin{figure}
    \centering
    \includegraphics[width=\columnwidth]{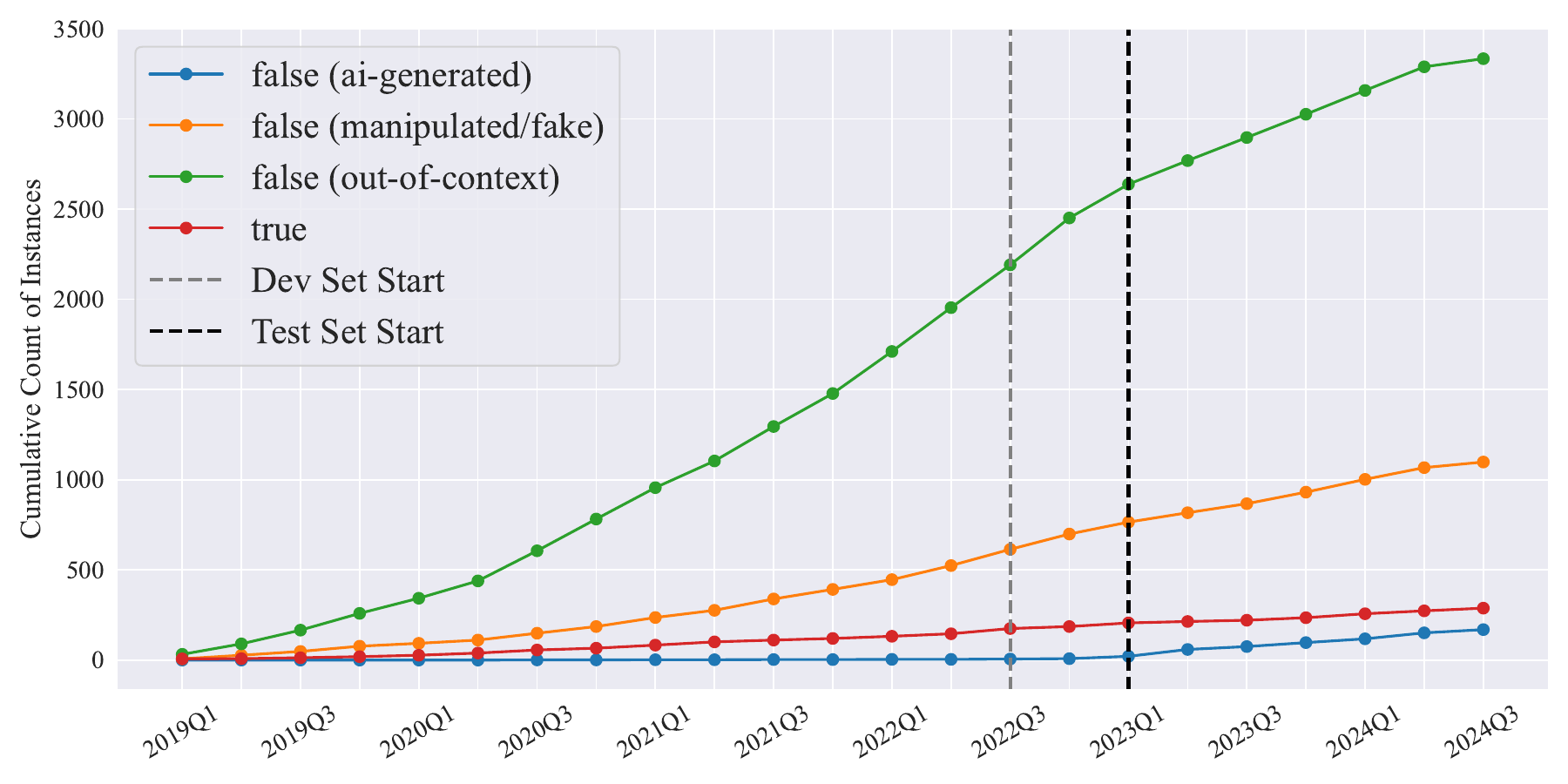}
    \caption{Cumulative distribution of the verdicts over time. The categories with few instances, \textit{unverified}, \textit{mixture},  \textit{false (authentic)}, are not displayed.}
    \label{fig:timeline}
\end{figure}

\paragraph{Task combination distribution}

Figure~\ref{fig:upset_plot} presents the distribution of task combinations in the dataset. Each row corresponds to a task, and each column represents a unique combination of tasks. The bar chart above shows the number of instances for each combination. The majority of instances are annotated with labels for at least four tasks, while combinations involving only one or two tasks are rare.

\paragraph{Verdict distribution}

In the \textit{imbalanced} setting of verdict prediction, true claims represent only 5.9\% of the entire data. This proportion increases to 26.0\% in the \textit{balanced} setting. False claims are dominated by out-of-context misinformation (72\%), aligning with prior findings \citep{dufour2024ammeba}. Manipulated and man-made fake images account for 24.0\%, while only 3.6\% of the false claims involve AI-generated images. Furthermore, AI-generated images mostly appeared from 2023, and are therefore concentrated in the test set. Figure \ref{fig:timeline} shows the cumulative evolution over time of the verdict labels, with the cutoff dates for the dev and test set indicated by vertical dashed lines.

\paragraph{Claim language distribution}

The language distribution under the \textit{multilingual} setting is as follows: English (63.8\%), Arabic (12.4\%), Portuguese (5.0\%), French (4.6\%), Spanish (4.0\%), German (3.6\%), Telugu (3.0\%), Dutch (1.6\%), Tamil (1.0\%), Turkish (1.0\%).

\paragraph{Geographic distribution}

\begin{figure}
    \centering
    \includegraphics[width=\columnwidth]{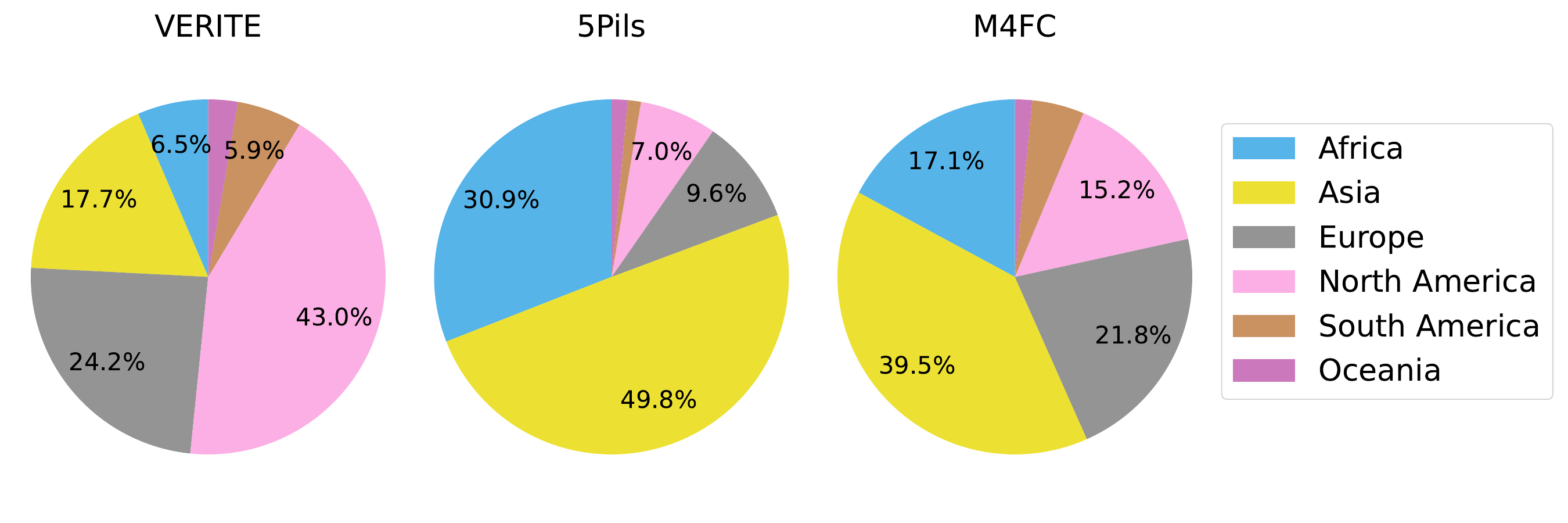}
    \caption{Distribution by continent of the images with known location of VERITE, 5Pils, and M4FC.}
    \label{fig:geographic_dist}
\end{figure}

\begin{figure}
    \centering
    \includegraphics[width=\columnwidth]{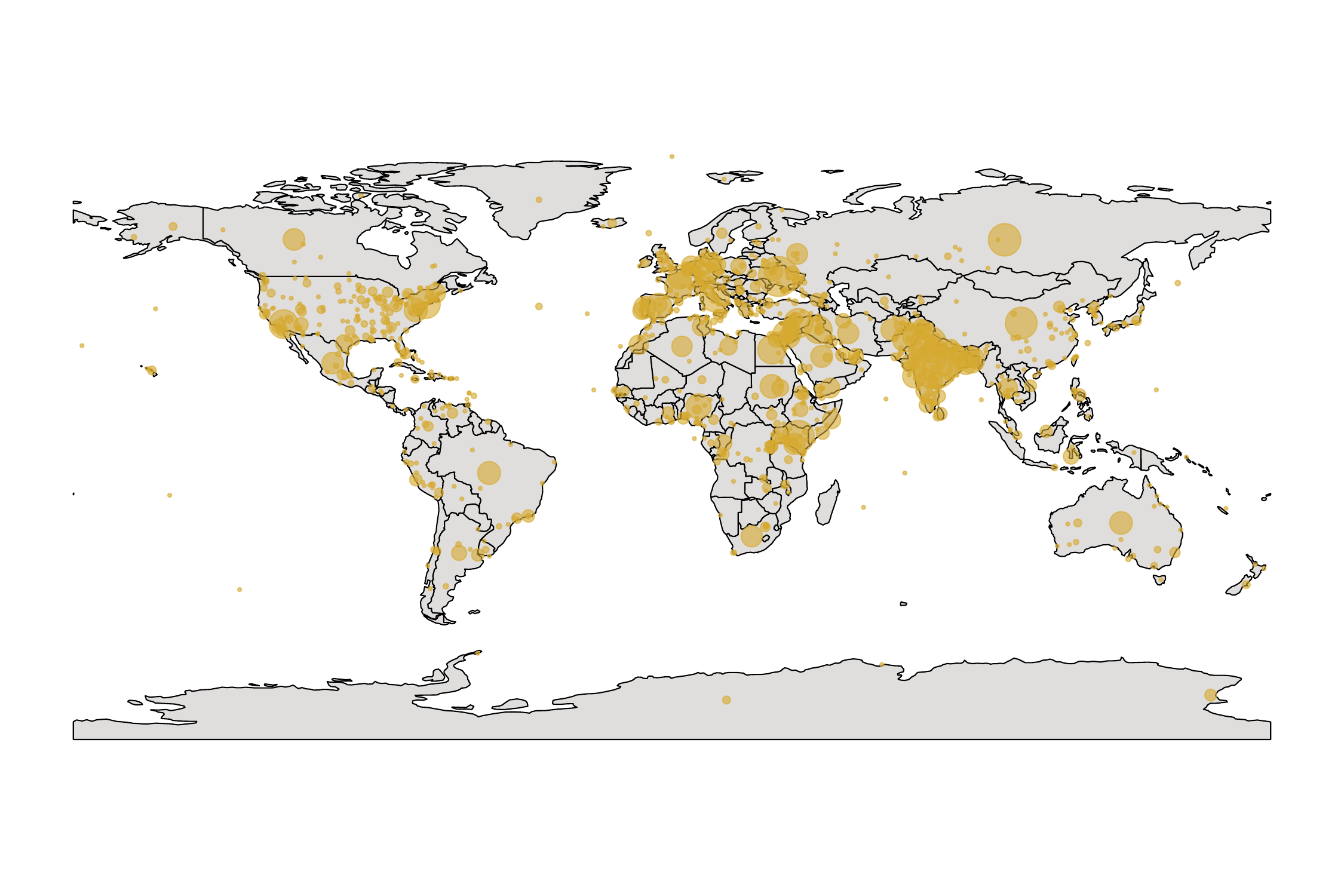}
    \caption{Geographic distribution of images in M4FC with known location.}
    \label{fig:geographic_map}
\end{figure}

\begin{figure}
    \centering
    \includegraphics[width=\linewidth]{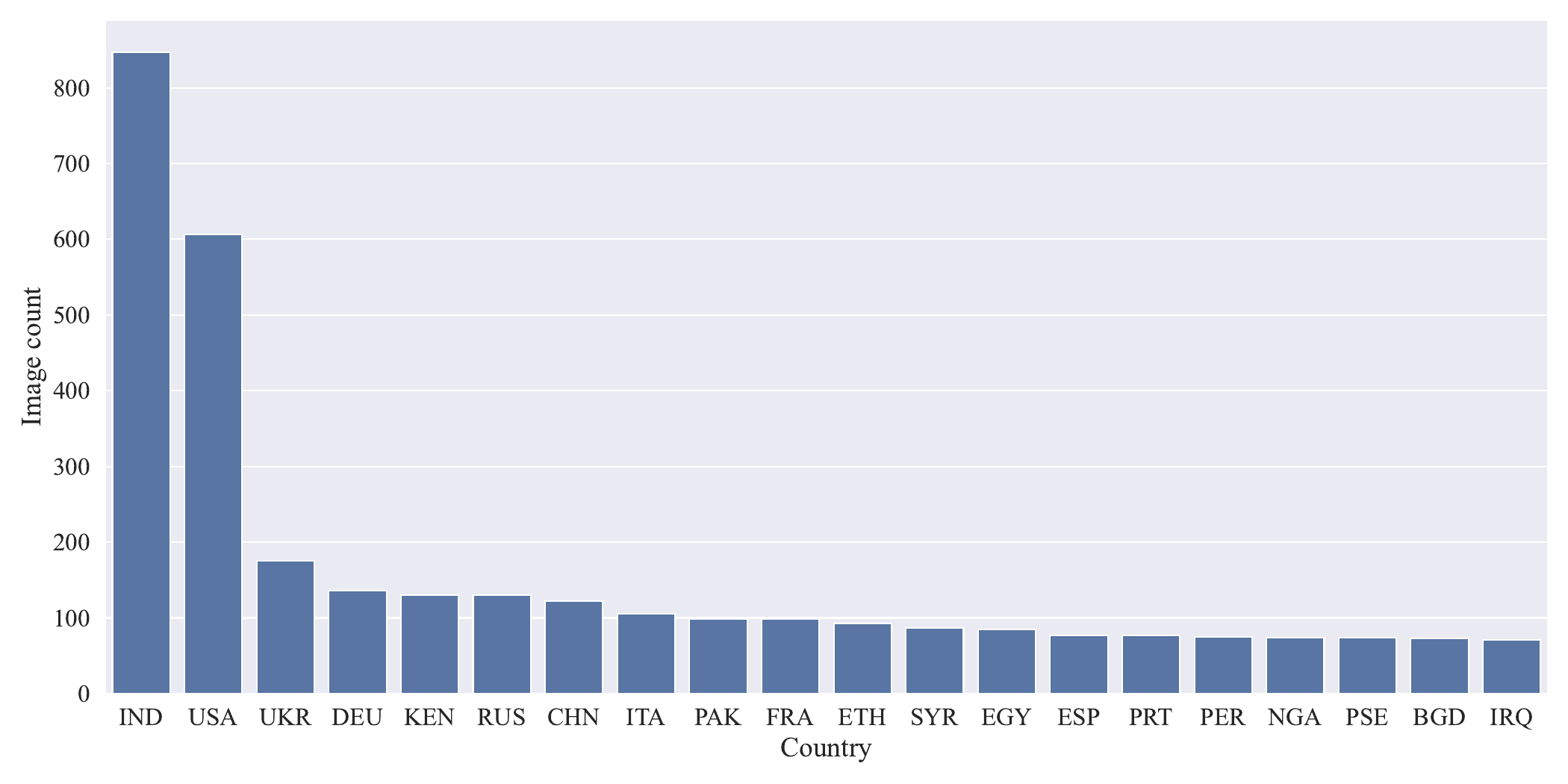}
    \caption{Top-20 country distribution of images in M4FC with known location. Countries are represented by their ISO 3166-1 alpha-3 codes.}
    \label{fig:image_location_distribution}
\end{figure}

Figure \ref{fig:geographic_dist} compares the geographic distribution of images with location context items in M4FC and two previous datasets with known location labels, VERITE and 5Pils. VERITE is skewed toward Western contexts, while 5Pils is largely concentrated in Asia and Africa. In contrast, M4FC shows the highest geographic diversity, featuring images from a broad range of cultural contexts. Figure \ref{fig:geographic_map} further highlights the geographic diversity of M4FC. However, some regions remain underrepresented, including East Asia, South America, and Oceania. Figure \ref{fig:image_location_distribution} shows the distribution by country of images with known locations. India and the USA are the most represented, which aligns with their large populations and the strong presence of fact-checking organizations.

 \begin{table}
  \small
  \centering
  \resizebox{\columnwidth}{!}{ %
  \begin{tabular}{llllll}
    \toprule
    Task   & \# instances & \# train   & \# dev   & \# test   \\
    \midrule
     VCE & 308 & 160 & 27 & 121 \\
     CIP & 212 & 131 & 19 & 62  \\
     FD & 432 & 234 & 38 & 160 \\
     IC & 403 & 221 & 35 & 147   \\
     LV & 21 & 3 & 6 & 12    \\
     VP & 432 & 234 & 38 & 160 \\

    \bottomrule
  \end{tabular}}
  \caption{Number of instances per task in the human-validated subset.}
  \label{tab:subset_table}
\end{table}

\begin{figure*}
    \centering
    \includegraphics[width=\textwidth]{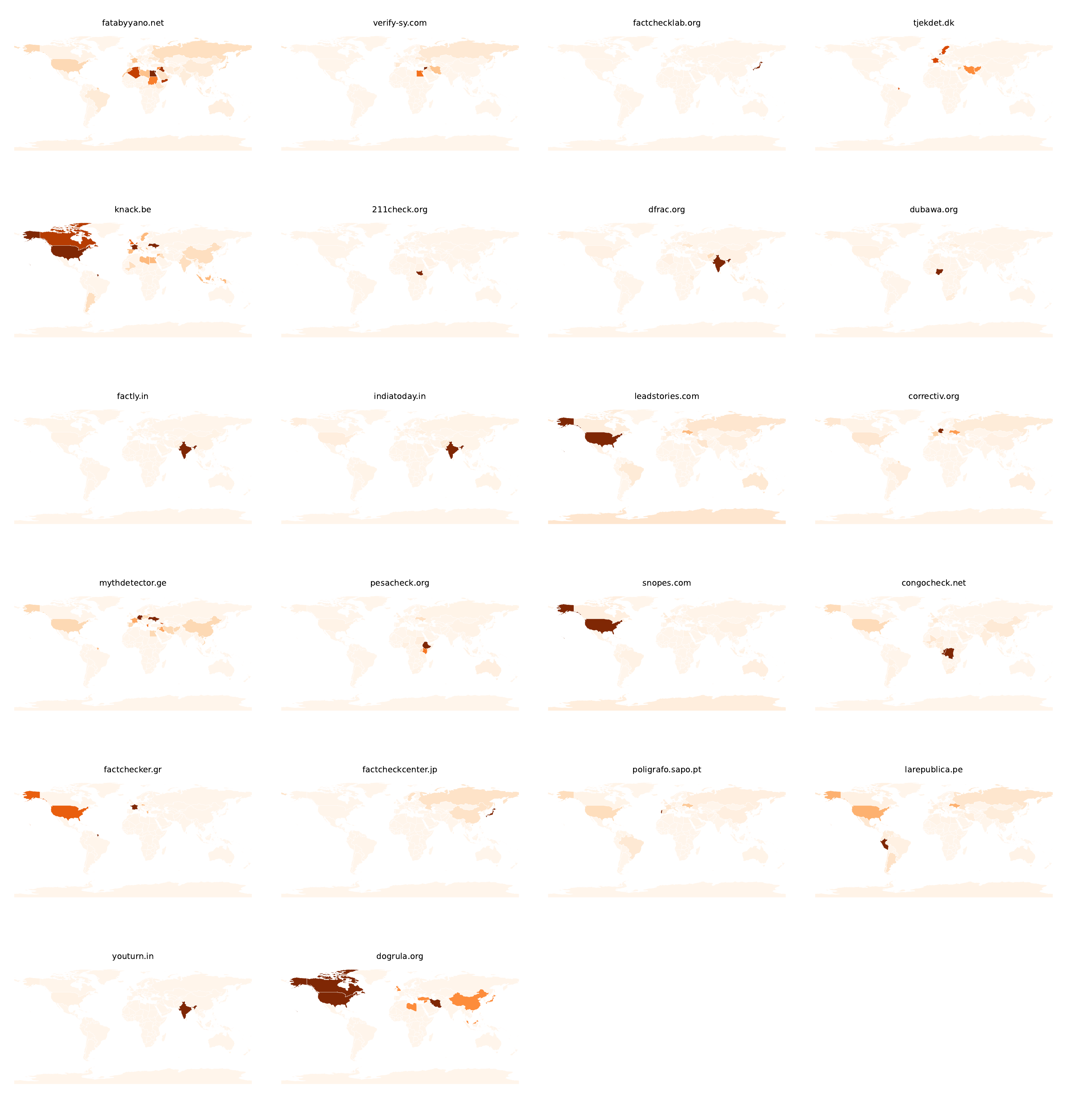}
    \caption{Distribution by country and by fact-checking organization of the claimed locations of images in M4FC.}
    \label{fig:claimed_loc_geo}
\end{figure*}

\begin{figure*}
    \centering
    \includegraphics[width=\textwidth]{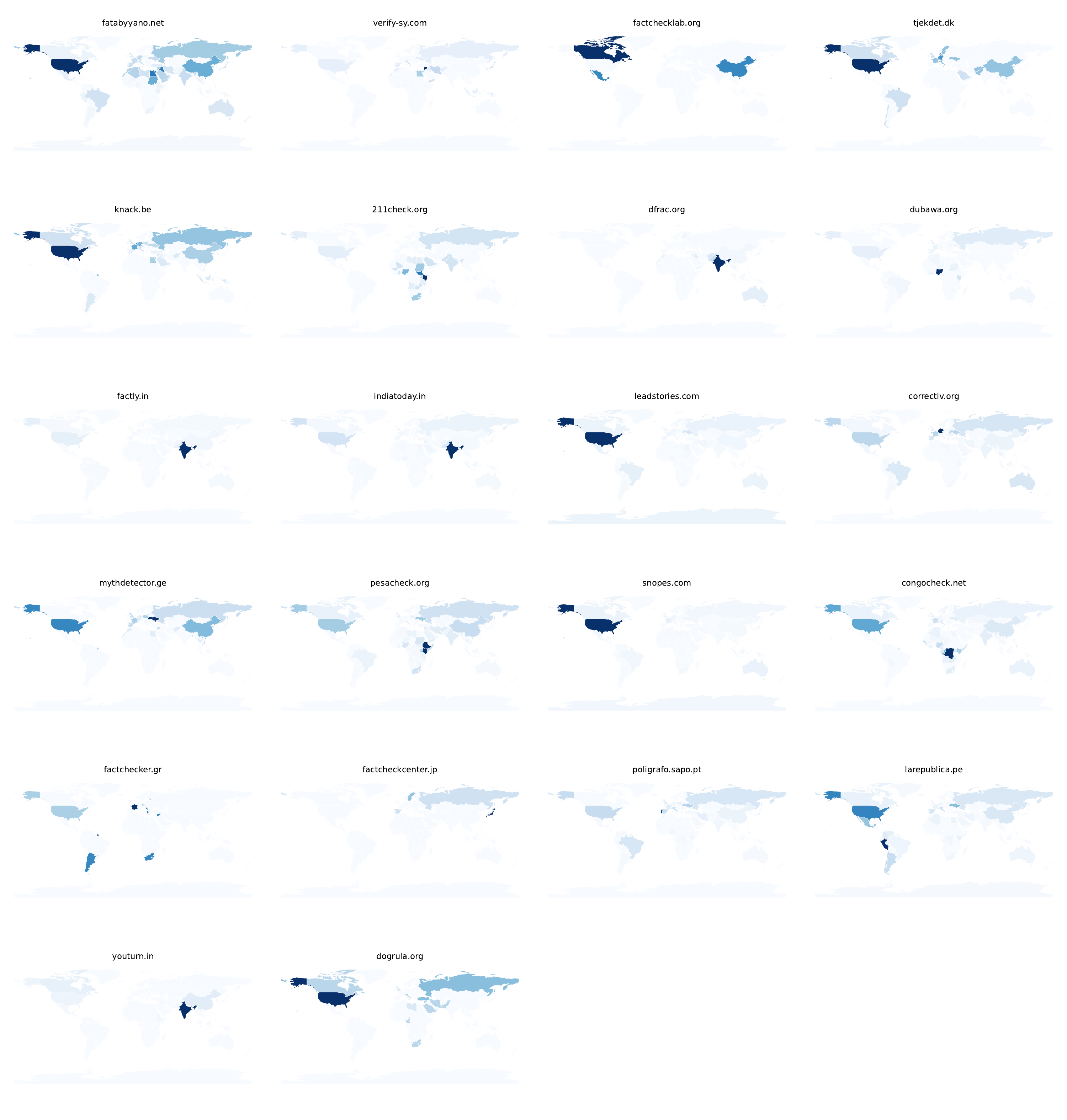}
    \caption{Distribution by country and by fact-checking organization of the locations of images in M4FC.}
    \label{fig:loc_geo}
\end{figure*}

\begin{table*}
\centering
\resizebox{\textwidth}{!}{
\begin{tabular}{p{0.97\textwidth}}
\toprule

\textbf{Task:}  Visual Claim Extraction \\
\textbf{Prompt:} You are an assistant designed to support fact-checking. Your task is to help identify the primary claim conveyed by the shared image that requires verification. 
Clearly articulate this claim in a single, concise, and neutral sentence or short paragraph, beginning with ``Claim:'' \\

\midrule

\textbf{Task:}  Claimant Intent Prediction \\
\textbf{Prompt:}  You are a helpful assistant supporting fact-checking. Given the image and the related claim: \texttt{\{claim\}}, your task is to help initialize the prediction of the user's motivation/intention for sharing this image for further analysis. 
Clearly articulate it in a single, concise, and neutral sentence or short paragraph, starting with ``Motivation: To'' \\

\midrule

\textbf{Task: } Image Contextualization\\
\textbf{Prompt:}  You are an AI assistant designed to help analyze information. Given the image and related article \texttt{\{evidence\}}, your task is to identify the following important information:

\quad 1. People: Who is shown in the image? \\
\quad 2. Things: Which animal, plant, building, or object are shown in the image? \\
\quad 3. Event: Which event is depicted in the image? \\
\quad 4. Date: When was the image taken? \\
\quad 5. Location: Where was the image taken? \\
\quad 6. Motivation: Why was the image taken? \\
\quad 7. Source: Who is the source of the image?

Please provide your answers in the following format in English:

\quad People: [Answer], \\
\quad Things: [Answer], \\
\quad Event: [Answer], \\
\quad Date: [Answer], \\
\quad Location: [Answer], \\
\quad Motivation: [Answer], \\
\quad Source: [Answer]

If there is no answer, just leave it blank, like an empty string. Do not output anything else. The output must be in English. \\

\midrule 
\textbf{Task:}  Location verification \\
\textbf{Prompt:} You are a helpful assistant designed to support fact-checking. Your task is to verify whether a candidate location is accurate for an image. You are given the image, and [a map | a satellite | a map and a satellite] view of the candidate location as input. 
Answer only with true or false. \\
\midrule

\textbf{Task:} Verdict prediction Prediction (optionally with text evidence)\\
\textbf{Prompt:} You are an AI assistant designed to evaluate the veracity of multimodal claims. Given the claim, the related image, and retrieved evidence, your task is to analyze the provided information and predict the more likely veracity: True or False. Please do not refuse or respond with ``not enough information.'' Your prediction will assist professional fact-checkers in further analysis.

Please provide your prediction as ``Veracity: True'' or ``Veracity: False''.

In one or more paragraphs, output your reasoning steps. In the separate final line, output your prediction. (Please don't output anything else except the Veracity Prediction)

Claim: \texttt{\{CLAIM\}}

Retrieved Evidences: \texttt{\{EVIDENCE\}}

Let's think step by step. \\

\bottomrule
\end{tabular}}
\caption{Prompt templates for the baseline experiments.}
\label{tab:prompt_templates_tasks}
\end{table*}

\paragraph{Geographic distribution per organization}

Figures \ref{fig:claimed_loc_geo} and \ref{fig:loc_geo} show the geographic distribution by country of the claimed and real location of M4FC images, respectively. One map is drawn for each fact-checking organization in M4FC. The maps show that fact-checking organizations tend to focus on images claimed to be associated with their own country. For example, Factly.in and Dfrac.org, both based in India, focus on images with claimed location India. Notable exceptions include Fatabyyano.net, Knack.be, and Dogrula.org.
In addition to their own country, many organizations verify images with Ukraine as the claimed location, reflecting the ongoing conflict that began in 2022. For some organizations, the real locations at the country level do not differ significantly from the claimed locations. This is the case for organizations like Factly.in and Indiatoday.in, where all claimed and real locations are within India. This could either mean that location is less frequently misrepresented in out-of-context claims in India, or that the difference in location is at a more fine-grained level, e.g., different cities.

\paragraph{Web evidence statistics}

After filtering, we obtain web pages with the Google RIS engine for only 48.62\% of the images in M4FC. In total, 9,693 web pages are retrieved, averaging 4 per image, excluding images with 0 pages. The web pages are written in 84 different languages. English is the most frequent (65\%), followed by Spanish (5\%), French (3\%), and Arabic (2\%).

 \section{Human-validated subset}
 \label{sec:subset_annotated}

We release a separate subset of M4FC, called M4FC-validated, which includes only the labels approved by both Prolific annotators during the data validation phase. This subset contains 434 instances. Of these, 235 come from the train set, while 38 and 161 come from the dev and test sets, respectively. The task distribution of this subset is reported in Table \ref{tab:subset_table}.

 \section{G-Eval prompts}
 \label{sec:g-eval}

Tables \ref{tab:geval_vce} and \ref{tab:geval_cip} provide the G-Eval prompts.

 \begin{table*}
\centering
\resizebox{\textwidth}{!}{%
\begin{tabular}{p{0.97\textwidth}}
\toprule
\textbf{Task:} Prompt for accessing visual claim extraction results.\\
\textbf{Prompt:} You will be given: \\
 1. A ground-truth normalized claim written by a human. \\
 2. An AI-generated normalized claim. \\
 Both claims are produced from the same underlying image content. \\
Your task is to evaluate how consistent and faithful the AI-generated normalized claim is compared with the ground-truth version. \\
\textbf{Definition}: \\
Claim Normalization refers to the process of transforming the information contained in an image into a concise textual claim that captures the core semantic meaning. A good normalized claim should accurately reflect the image's actual content, entities, relationships, and actions. \\
\textbf{Evaluation Criteria}: \\
Consistency (1-5): \\
5 -- Fully faithful to the ground-truth claim; preserves all core semantic elements (entities, relations, actions) with no distortions or additions. \\
4 -- Mostly faithful; minor omissions or rephrasing but no incorrect or contradictory content. \\
3 -- Partially faithful; preserves some main elements but misses key relationships or adds ambiguous wording. \\
2 -- Low faithfulness; multiple incorrect or missing semantic elements; contains hallucinations or distorts relationships. \\
1 -- Poor; contradicts the main meaning of the ground-truth claim or fails to represent the image content accurately. \\
\textbf{Evaluation Steps:}\\
1. Read the ground-truth normalized claim and identify its core semantic units: \\
   - entities (who/what) \\
   - attributes (descriptions) \\
   - actions/events \\
   - relationships \\
   - spatial or contextual cues \\
2. Read the AI-generated normalized claim. \\
3. Compare the semantic overlap and accuracy: \\
   - Does the AI version preserve the core entities? \\
   - Does it distort actions or relationships? \\
   - Does it add hallucinated content not in the ground-truth? \\
   - Does it omit essential elements? \\
4. Assign a Consistency score (1-5) strictly based on the Evaluation Criteria. \\
Return STRICT JSON with keys: \\
\{"verdict": one of [consistent, partially\_consistent, inconsistent], "score": integer 1-5, "explanation": brief reason\}. No extra text. \\
\textbf{Evaluation Form (scores ONLY): }\\
- Consistency: \\
Ground Truth Normalized Claim: \\
<ground-truth-claim-here> \\
AI-Generated Normalized Claim: \\
<ai-generated-claim-here> \\
Respond with JSON only.\\
\bottomrule
\end{tabular}%
}
\caption{G-Eval prompt for VCE task.}
\label{tab:geval_vce}
\end{table*}

\begin{table*}
\centering
\resizebox{\textwidth}{!}{%
\begin{tabular}{p{0.97\textwidth}}
\toprule
 \textbf{Task:} Prompt for accessing claimant intent prediction results.\\
\textbf{Prompt:} You will be given: \\
 1. A ground-truth motivation written by a human. \\
 2. An AI-generated motivation. \\
Your task is to evaluate how well the AI-generated motivation is consistent and faithful to the meaning and reasoning of the ground-truth version. \\
\textbf{Evaluation Criteria: }\\
Consistency (1-5): \\
5 -- Fully faithful, no contradictions, and captures all essential reasoning points. \\
4 -- Mostly faithful; minor omissions but no distortions. \\
3 -- Partially faithful; captures some main points but misses or alters key logic. \\
2 -- Low faithfulness; contains distortions, irrelevant reasoning, or major omissions. \\
1 -- Poor; contradicts or misrepresents the ground-truth motivation. \\
\textbf{Evaluation Steps:} \\
1. Read the ground-truth motivation and identify its key reasoning points. \\
2. Read the AI-generated motivation. \\
3. Compare meaning, reasoning steps, and any contradictions or hallucinations. \\
4. Assign a Consistency score from 1 to 5 based on the Evaluation Criteria. \\
Return STRICT JSON with keys: \\
\{"verdict": one of [consistent, partially\_consistent, inconsistent], "score": integer 1-5, "explanation": brief reason\}. No extra text. \\
\textbf{Evaluation Form (scores ONLY): }\\
- Consistency: \\
Ground Truth Motivation: \\
<ground-truth-text-here> \\
AI-Generated Motivation: \\
<ai-generated-text-here> \\
Respond with JSON only.\\
\bottomrule
\end{tabular}%
}
\caption{G-Eval prompt for CIP task.}
\label{tab:geval_cip}
\end{table*}

\section{Prompt templates for the M4FC tasks}
\label{sec:prompts_tasks}

Table \ref{tab:prompt_templates_tasks} reports the prompt templates used for the baseline experiments.

\section{Implementation details}
\label{sec:implementation-details}

\paragraph{MLLMs}

All models are evaluated in one zero-shot run on one A100 GPU with temperature set to 0. We use the HuggingFace \citep{wolf-etal-2020-transformers} implementation of the open-weight models. GPT4o-mini and Gemini-1.5-Flash are accessed via the OpenAI API and Google Cloud API, respectively.

\paragraph{Fake image detectors fine-tuning}

We fine-tune all models with the Adam optimizer and a learning rate of $2e-5$ for 10 epochs.

\paragraph{DEFAME} We employ the Google Custom Search API for both text and image retrieval. We allow up to three iterations, with each retrieval step returning the top-3 evidence items. The maximum output length of a single LLM generation is limited to 2,000 tokens (\texttt{max\_tokens}), while the overall fact-checking process, including the final report and all intermediate steps, is capped at 64,000 tokens (\texttt{max\_result\_len}).

\paragraph{COVE}  We reproduce all steps of the COVE pipeline, except the collection of evidence via direct search, which has limited impact on real-world misinformation detection \citep{tonglet-etal-2025-cove}.

\paragraph{RIS evidence}

For each image, we retrieve up to 20 web pages with the Google RIS engine. We extract their textual content using Trafilatura \citep{barbaresi-2021-trafilatura}.\footnote{\href{https://trafilatura.readthedocs.io/en/latest/}{trafilatura.readthedocs.io}} To prevent evidence leakage, we exclude any web pages published by a curated list of fact-checking organizations, as well as those published after the date of the fact-checking article from which the claim was sourced.
For tasks that require text evidence, we treat each web page retrieved with RIS as a single evidence. All 20 text evidence are provided in the prompt, sorted by cosine similarity with the image using CLIP embeddings \citep{pmlr-v139-radford21a}.

\section{Additional Results}
\label{sec:additional_results}

Given the smaller size of the test sets for CIP and LV, we evaluate the significance of the results using a bootstrap sample (n=10,000) with FDR correction.

For CIP, the significant tests showed that Gemini-1.5-Flash, the best-performing model, is significantly (p<0.05) more performant than the next-best MLLM across each metric. 

For LV, we analyze whether the F1 score achieved by a MLLM given a visual representation of a candidate location (either map, satellite or both) outperforms the F1 obtained given the text-based candidate location baseline. The following configurations perform significantly worse (p<0.05) than the text-based baseline: InternVL2.5-8B with map, InternVL2.5-8B with map and satellite, and GPT4o-mini with map.

\begin{table}
    \centering
    \resizebox{0.55\linewidth}{!}{ %
    \begin{tabular}{cccc}
    \toprule
      Zoom   &  \includegraphics[height=1em]{figures/OSM.png} & \includegraphics[height=1em]{figures/satellite.png} & F1 \\
      \midrule
       10  &  \cmarksilver &     & 68.85  \\
       10  &   &   \cmarksilver  & 66.59 \\
       10  &  \cmarksilver &  \cmarksilver   &  64.14 \\
       15 &  \cmarksilver &     & \digitbf{74.43} \\
       15  &   &   \cmarksilver  & 67.52 \\
       15  &  \cmarksilver &  \cmarksilver   & 65.62 \\
       17  &  \cmarksilver &     & 73.63 \\
       17  &   &   \cmarksilver  &  70.55 \\
       17  &  \cmarksilver &  \cmarksilver   & 69.75 \\
       \bottomrule
    \end{tabular}}
    \caption{Location verification F1 at different zoom level on M4FC test set with Gemini-1.5-Flash (\%).}
    \label{tab:zoom-level}
\end{table}

\begin{figure}
    \centering
    \includegraphics[width=\linewidth]{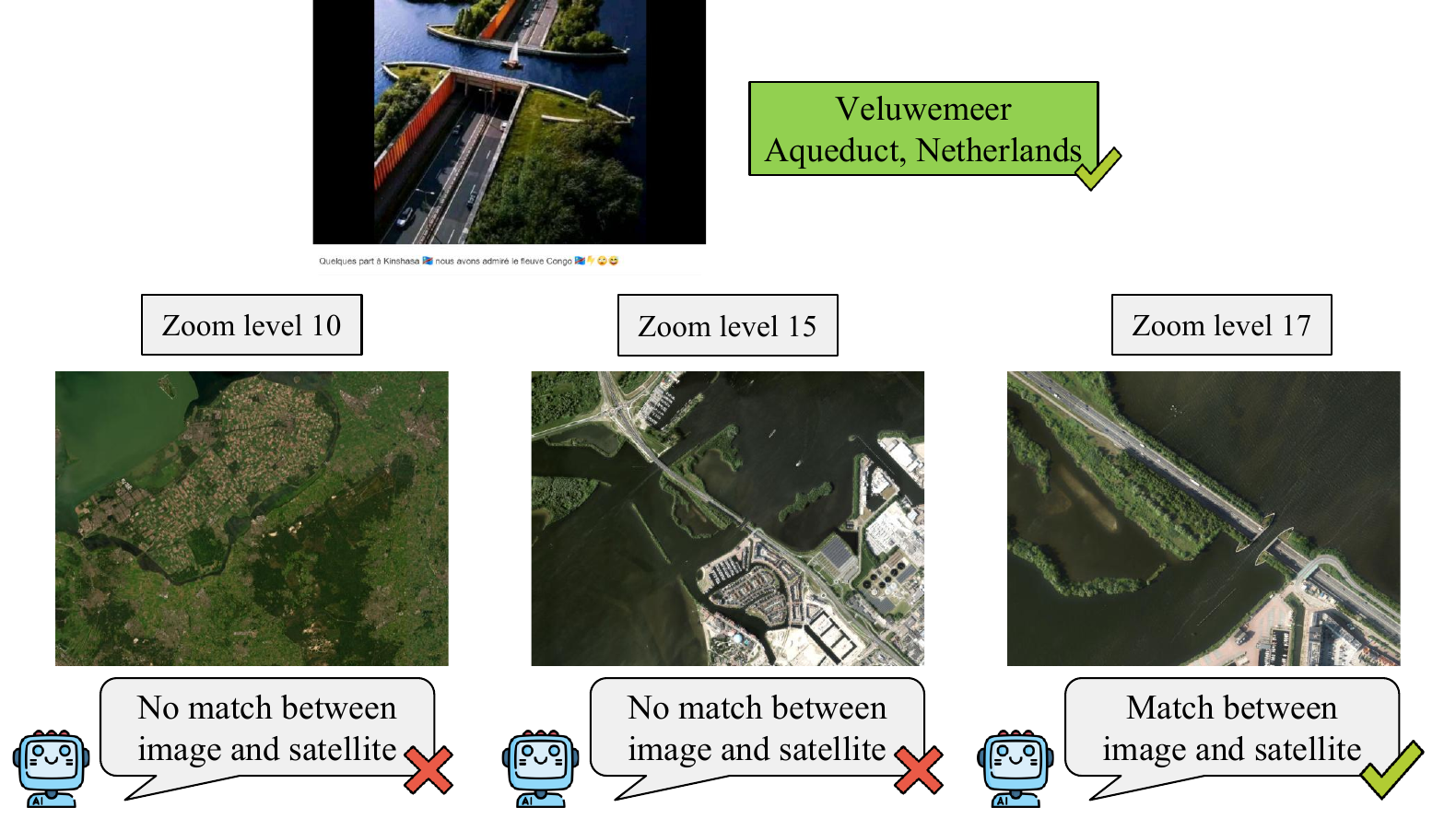}
    \caption{Illustration of location verification with satellite views at different zoom level. Satellite images are sourced from ESRI World Imagery.}
    \label{fig:zoom_level}
\end{figure}

\begin{table}[ht]
    \centering
    \resizebox{\linewidth}{!}{ %
    \begin{tabular}{cccccccc}
    \toprule
      & 2023 Q1 & 2023 Q2 & 2023 Q3 & 2023 Q4 & 2024 Q1 & 2024 Q2 & 2024 Q3 \\
      \midrule
       
InternVL-2.5-8B	&44	& 42	&46	&46	&39	&46	&43 \\
Qwen2.5-VL-7B	& 51	&45	&48	&48&	44	&47&	45\\
Llama-3.2-11B	&38	&39&	38	&39	&37	&40	& 42 \\
GPT4o-mini	&49&	46	&50	&48	&44	&49&	47 \\
Gemini-1.5-Flash&	57	&55&	56&	55&	50&	53&	54\\

       \bottomrule
    \end{tabular}}
    \caption{G-Eval scores by quarter for Visual Claim Extraction (\%).}
    \label{tab:quarter_vce}
 \end{table}

\begin{table}[ht]
    \centering
    \resizebox{\linewidth}{!}{ %
    \begin{tabular}{cccccccc}
    \toprule
      & 2023 Q1 & 2023 Q2 & 2023 Q3 & 2023 Q4 & 2024 Q1 & 2024 Q2 & 2024 Q3 \\
      \midrule
       
InternVL-2.5-8B	& 27	&27	&29&	33&	28&	29&	25 \\
Qwen2.5-VL-7B	& 31	&27	&36	&36	&37	&29	&26\\
Llama-3.2-11B	& 29	&28	&31	&30&	32	&28&	25\\
GPT4o-mini	& 28	&28	&35&	32	&34&	27&	27 \\
Gemini-1.5-Flash&	41&	38&	42&	45&	49&	38&	41\\

       \bottomrule
    \end{tabular}}
    \caption{G-Eval scores by quarter for Claimant Intent Prediction (\%).}
    \label{tab:quarter_cip}
\end{table}

\begin{table}[ht]
    \centering
    \resizebox{\linewidth}{!}{ %
    \begin{tabular}{cccccccc}
    \toprule
      & 2023 Q1 & 2023 Q2 & 2023 Q3 & 2023 Q4 & 2024 Q1 & 2024 Q2 & 2024 Q3 \\
      \midrule
Qwen2.5-VL-7B & 70 & 69 & 68 &  71 & 69 & 67 & 70 \\
GPT4o-mini	& 76 & 72 & 73 & 71 & 78 & 79 & 76 \\

       \bottomrule
    \end{tabular}}
    \caption{F1 scores by quarter fo Verdict prediction (\textit{English-only, imbalanced}) (\%).}
    \label{tab:quarter_vp}
\end{table}

\section{Location verification at different zoom levels}
\label{sec:satellite_zoom}

As shown in the second error example of Figure \ref{fig:error-location-verification}, better results could be obtained at different zoom levels, as illustrated in Figure \ref{fig:zoom_level}. However, the appropriate zoom level may vary a lot. Future work should consider directly incorporating the desired zoom level for the aerial views as a parameter of their modeling approach.  We provide preliminary results in Table \ref{tab:zoom-level}, showing the F1 of Gemini-1.5-Flash on M4FC test set at three different zoom levels: 10, 15, and 17. The best overall performance is obtained at a zoom level of 15.

\section{Test set results by quarter}
\label{sec:knowledge}

Tables \ref{tab:quarter_vce}, \ref{tab:quarter_cip}, and \ref{tab:quarter_vp} report the results by quarter of the year for the test set for Visual Claim Extraction and Claimant Intent Prediction, and Verdict Prediction (\textit{English-only, imbalanced}). In total, there are seven quarters, from 2023 Q1 to 2024 Q3. The MLLMs used in our experiments have their knowledge cutoff in 2023, meaning that events in 2024 are outside of their parametric knowledge. If parts of the RIS evidence set, or the fact-checking articles themselves, had leaked in the pre-training data of the MLLMs, we would expect a higher performance on the 2023 quarters than on the 2024 ones.
On the contrary, our results show that scores oscillate within the same range across all quarters and all MLLMs. This clearly shows that the MLLMs are not advantaged by their parametric knowledge for claims published in 2023.

\end{document}